\documentclass[lettersize,journal]{IEEEtran}
\usepackage{amsmath,amsfonts}
\usepackage{algorithm}
\usepackage{algpseudocode}
\usepackage{amsmath}
\usepackage{array}
\usepackage[caption=false,font=normalsize,labelfont=sf,textfont=sf]{subfig}
\usepackage{textcomp}
\usepackage{stfloats}
\usepackage{url}
\usepackage{verbatim}
\usepackage{graphicx}
\usepackage{cite}
\usepackage{svg}
\usepackage{booktabs}
\usepackage{multirow}
\usepackage{rotating}
\usepackage{tabularx}
\usepackage{bm}
\usepackage{ulem}
\usepackage{orcidlink}
\usepackage{float}
\usepackage{xcolor}
\usepackage{tikz}
\usetikzlibrary{shapes.geometric, arrows}
\newcolumntype{Y}{>{\centering\arraybackslash}X}

\begin{document}
	
	\title{JDPNet: A Network Based on Joint Degradation Processing for Underwater Image Enhancement}
	
	\author{Tao Ye$^{\orcidlink{0000-0002-1814-530X}}$, Hongbin Ren$^{\orcidlink{0009-0009-6964-6604}}$, Chongbing Zhang$^{\orcidlink{0009-0007-6027-0159}}$, Haoran Chen$^{\orcidlink{0009-0001-6071-174X}}$, Xiaosong Li$^{\orcidlink{0000-0003-4672-1527}}$
		\thanks{This work was supported in part by the Fundamental Research Funds for the Central Universities under Grant (2024ZKPYJD04), the National Natural Science Foundation of China (No. 52374166), the Beijing Natural Science Foundation (L221018) and the National Natural Science Foundation of China (No. 62201149). (Corresponding author: Tao Ye.) 
			
			\par Tao Ye, Hongbin Ren and Haoran Chen are with the School of Mechanical and Electrical Engineering, China University of Mining and Technology, Beijing 100083, China (E-mail: yetao@cumtb.edu.cn, rhb\_2020@163.com, a1430743928@gmail.com).
			\par Chongbing Zhang is with the China Shipbuilding Science Research Center, Wuxi 214062, China (E-mail: zhangchongbing@cssrc.com.cn).
			\par Xiaosong Li is with the School of Physics and Optoelectronic Engineering, Foshan University, Foshan 528225, China (E-mail: lixiaosong@buaa.edu.cn).}}

	
	
	\maketitle
	
	\begin{abstract}
		Given the complexity of underwater environments and the variability of water as a medium, underwater images are inevitably subject to various types of degradation. The degradations present nonlinear coupling rather than simple superposition, which renders the effective processing of such coupled degradations particularly challenging. Most existing methods focus on designing specific branches, modules, or strategies for specific degradations, with little attention paid to the potential information embedded in their coupling. Consequently, they struggle to effectively capture and process the nonlinear interactions of multiple degradations from a bottom-up perspective. To address this issue, we propose JDPNet, a joint degradation processing network, that mines and unifies the potential information inherent in coupled degradations within a unified framework. Specifically, we introduce a joint feature-mining module, along with a probabilistic bootstrap distribution strategy, to facilitate effective mining and unified adjustment of coupled degradation features. Furthermore, to balance color, clarity, and contrast, we design a novel AquaBalanceLoss to guide the network in learning from multiple coupled degradation losses. Experiments on six publicly available underwater datasets, as well as two new datasets constructed in this study, show that JDPNet exhibits state-of-the-art performance while offering a better tradeoff between performance, parameter size, and computational cost.
	\end{abstract}


	\section{Introduction}
	\IEEEPARstart{W}{ith} the development of underwater detection technology, underwater robots have found applications in numerous fields such as marine resource exploration \cite{ref1}, underwater environment monitoring \cite{ref2}, and marine archaeology \cite{ref3}. However, underwater imaging is affected by the coupling of various optical and physical factors, such as the absorption and scattering of light by water, the concentration of particulate matter, currents, and plankton, leading to significant degradation in the quality of underwater images. This degradation manifests as color shifts and low brightness, clarity, and contrast \cite{ref4}, which severely affect the imaging performance of underwater robots. Additionally, the complexity and variability of the water medium and the disturbances caused by underwater robot propellers stirring up sediment further deteriorate underwater imaging conditions. Increased uncertainty is not present in isolation and cannot be treated as a simply linear superposition; rather, issues such as light attenuation, color distortion, reduced contrast, and turbidity show strong coupling and nonlinear mixing. This significantly affects the perception of underwater vision and the subsequent tasks of underwater image analysis and understanding, such as detection, identification, and tracking of underwater objects  \cite{ref5,ref6,ref7}. Consequently, effectively restoring the complex coupled degradations in underwater images to  achieve a balanced improvement in contrast, clarity, and realism is a critical and challenging issue in underwater image processing.

	Existing enhancement methods can be broadly classified into deep and non-deep learning approaches. Non-deep learning methods \cite{ref8,ref9,ref10,ref11,ref12,ref13,ref15,MMLE,HLRP} are typically grounded in physical models or image priors and adjust pixel values to enhance visual quality. Although these methods can enhance image quality, they are not designed to address multiple coupled degradations. This renders modeling complex, nonlinear, and degenerate couplings difficult. Its adaptability and transferability are severely limited by predefined assumptions about the water medium and environmental conditions, rendering it ineffective in handling multiple coupled degradations.
	
	Contrarily, neural networks, with their powerful nonlinear representation capabilities, have shown outstanding performance in image processing \cite{ref35,ref36,ref37}. Deep-learning methods leverage vast amounts of data to learn underwater image characteristics and patterns, enabling automated and intelligent enhancement.
	Although many CNN-based restoration models adopt a unified backbone, they typically do not explicitly model the interactions among different degradation types. In practice, models are often trained for specific degradations or rely on the network to implicitly fit each degradation independently, which limits their ability to handle complex coupled degradations. Transformer-based methods excel at global context modeling but are fundamentally limited by the quadratic cost of self-attention; to mitigate this, UIE-oriented designs often introduce hierarchical/windowed or multi-branch modules, whose increased architectural intricacy may weaken generalization under strongly coupled degradations, particularly when tuned for individual degradations instead of their joint interactions.
	Recently, state-space model-based methods have been introduced for underwater image enhancement, leveraging linear-complexity global modeling and capturing long-range dependencies. Although these methods improve the modeling of globally coupled degradations, they still suffer from significant shortcomings in addressing spatially heterogeneous distortions and the restoration of local textures.

	In summary, most existing underwater image enhancement methods either design dedicated branches or modules for different types of degradations, or adopt purely data-driven end-to-end learning approaches. They appear to be designed based on the linear superposition of various degradations without explicitly exploring the underlying mechanisms or coupling of degradations. Several studies have attempted to construct a unified framework that models the nonlinear coupling effects of multiple degradations from a bottom-up perspective. Therefore, although methods based on deep learning show some adaptability to different scenarios, they cannot thoroughly handle coupling degradation scenarios because they do not consider coupling degradation information.

\begin{figure*}[h]
	\centering
	\includegraphics[width=5 in]{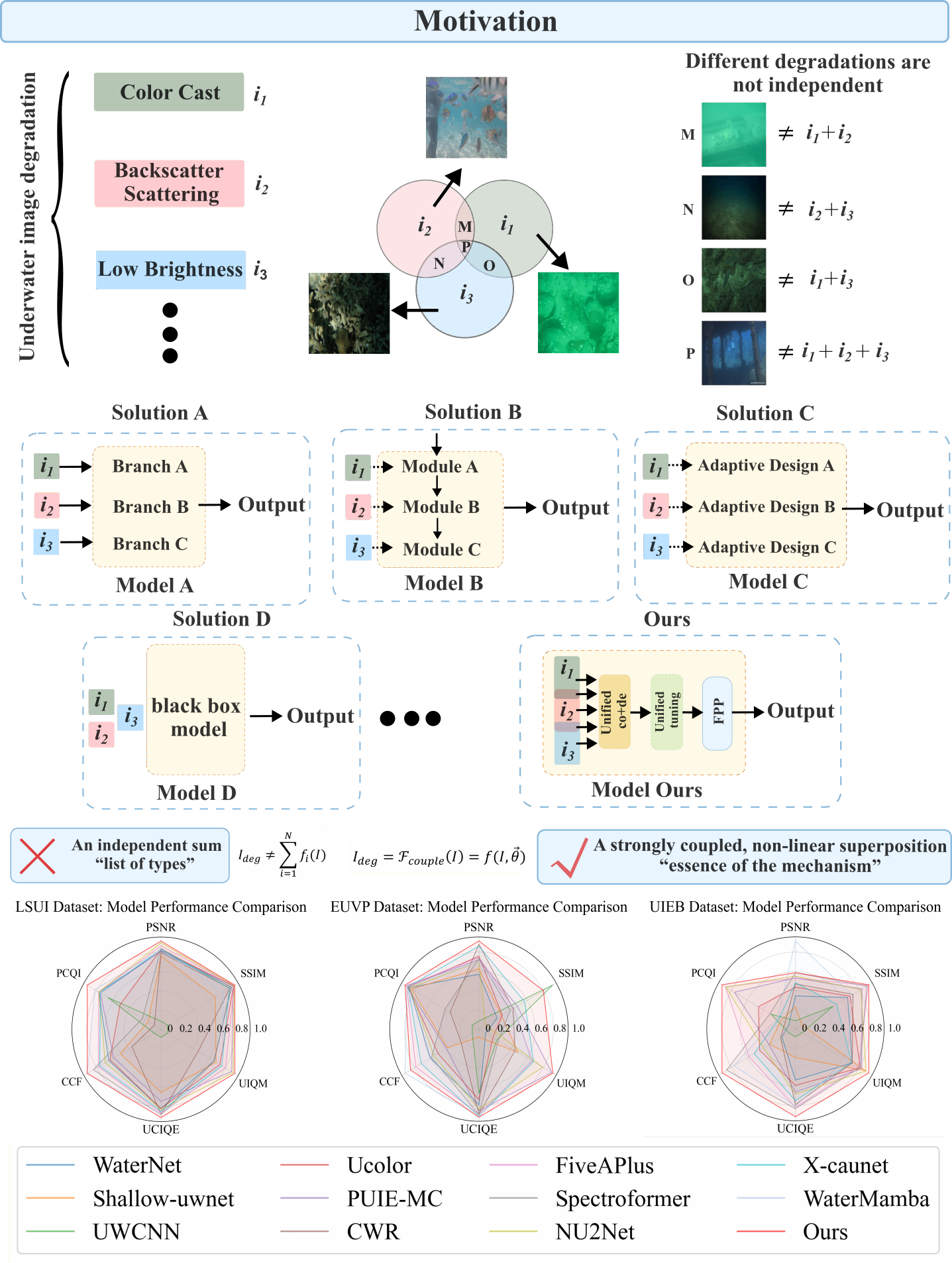}
	\caption{Motivation behind our proposed method; several radar charts comparing the performance of JDPNet with numerous advanced methods across multiple datasets.}
	\label{Motivation}
\end{figure*}

	Therefore, we propose JDPNet, a novel network that implements the mining and unified adjustment of coupling degradation information and corrects coupling degradation scenarios from the perspective of the underlying mechanism for underwater image enhancement (UIE). In contrast to previous works that treat underwater degradations as a simple linear additive combination ($\sum_{i=1}^N f_i(I)$), in this study, we argue that these degradations are strongly coupled and nonlinear, which can be formulated as a complex nonlinear function ($I_{\text{deg}} = \mathcal{F}(I; \vec{\theta})$), where multiple degradation processes interact with each other (Fig. \ref{Motivation}).
	
	In the proposed JDPNet, we first introduce a multi-stage cascaded design in the joint feature extraction (JFE) module to mine the latent information from various coupled degradations. This encoding module considers the effects of color, illumination, scattering, and other factors to capture the complexity of underwater images. Subsequently, these feature maps are passed through a probabilistic bootstrap (PB) module, which adjusts their statistical properties to match those of the training features. We also design a feature post-processing (FPP) module to further refine and enhance the output features of the framework. The FPP module employs grayscale correction and a border enhancement mask (BEM) strategy to reduce information loss in edge areas, address boundary bias during inference, and enhance boundary clarity and detail while maintaining the overall image quality. Finally, to balance color, clarity, and contrast, and ensure that the network effectively learns the features of various coupled degradations, we design a novel AquaBalanceLoss function. This function supervises the learning process of the network and enhances its feature inference capabilities.

	To fully verify the effectiveness of our proposed method in actual coupling degradation application scenarios, we create two new underwater datasets, ROV (7066 images in total) and QDH datasets (6798 images in total).	Using these two datasets, we conduct comprehensive experiments on the open underwater datasets UIEB\cite{UIEB}, LSUI\cite{New5}, EUVP\cite{EUVP}, U45\cite{U45}, C60, and UCCS\cite{UCCS}, and comprehensively evaluate the model performance using PSNR\cite{PSNR}, SSIM\cite{SSIM}, UIQM\cite{UIQM}, UCIQE\cite{UCIQE}, PCQI, and CCF indicators. Comprehensive quantitative and qualitative comparisons indicate that our proposed JDPNet achieved the best performance in addressing coupled degradation scenarios, excellent generalization ability and scene robustness, and a better balance among color, sharpness, and contrast than the UIE method of SOTA.

	The main contributions of this study are as follows.

		\begin{itemize}			
			
			\item To address the challenge of multiply coupled and interactively aggravated degradations in underwater images, we proposed JDPNet, a unified deep-learning framework that enables robust and effective enhancement across diverse and challenging underwater environments, achieving state-of-the-art results on eight underwater datasets, including the newly constructed ROV and QDH datasets.
			
			\item To resolve the problem of complex feature interactions caused by multiple coupled degradations, we designed a series of advanced modules, including JFE for adaptive feature mining, probabilistic bootstrap (PB) for unified adjustment of coupling degradation potential information, and FPP with BEM, which collaboratively mine and unify latent representations for effective global enhancement and artifact mitigation.
			
			\item To guide the model toward balanced optimization in color fidelity, sharpness, and contrast, we design the AquaBalanceLoss, which constrains the network to simultaneously account for perceptual and physical attributes characteristic of diverse underwater degradations.

		\end{itemize}

	The remainder of this article is organized as follows. The first section introduces the challenges and motivations underlying our work. The second section reviews related studies on underwater image enhancement. The third section presents the architecture of JDPNet, detailing its modules and the AquaBalanceLoss function. The fourth section describes the experiments conducted to evaluate our method, and the fifth section describes an ablation study to analyze the contributions of the individual components. Finally, the sixth section concludes the paper and discusses potential future directions.

	\section{Related works}
	In this section, we review the existing studies in the field of underwater image enhancement, emphasizing the advancements and limitations of current methodologies for addressing jointly coupled degradation problems.

	\subsection{Non-Deep Learning Methods for UIE}

	Non-deep learning methods for underwater image enhancement can be primarily categorized into nonphysical model approaches and physics-based methods. Non-model-based methods, such as histogram equalization, white balance, wavelet transforms, and color constancy~\cite{ref8,ref12,ref15,ref9,MMLE,HLRP}, directly manipulate pixel values to enhance underwater images but often fail because of color casts, low contrast, and noise. Some approaches are tailored for underwater conditions by adapting color constancy or designing multistage pipelines; however, they struggle to balance detail enhancement and naturalness under coupling degradation. Conversely, physics-based methods model the image formation process and leverage priors such as dark channels or attenuation curves\cite{ref40,ref41,ref42,ref45,ref48}; however, their strong predefined assumptions limit their robustness when facing real-world coupled degradations; thus, they do not have stability or generalization in coupled degradation underwater environments.

	\subsection{CNN-based Method for UIE}
	
	In contrast to traditional methods, convolutional neural networks (CNNs) based on deep learning can learn the latent features of underwater images using end-to-end training. These approaches leverage the powerful feature extraction and nonlinear fitting capabilities of CNNs to learn the nonlinear mapping between degraded and clear images, resulting in increased attention to different types of degradation and superior image enhancement performance.
	
	The Water-Net model \cite{ref63} employs adaptive filters and deep CNNs to enhance image quality by performing multiplication and addition operations on images and using convolutional kernels of various scales to extract features. The results of the three traditional image-processing methods were fused with the predicted confidence map to obtain enhanced results. Li \cite{ref64} proposed a new convolutional neural network (UWCNN) that designs a unique enhancement branch for each water type, trained with specific data to adapt to different underwater scenes. Subsequently, Li \cite{ref65} introduced a CNN-based UIE method (Ucolor) that adaptively integrates and emphasizes key features across multiple color spaces through guided intermediary transmission. Sun et al. \cite{ref66} introduced a CNN-based UIE that uses convolutional layers for noise filtering and deconvolutional layers for detail restoration and image optimization. Naik et al. and Xue \cite{ref67} proposed a neural network architecture for UIE, termed shallow-UWNet. Xue \cite{ref68} developed a multi-branch aggregation network to jointly predict coarse, veil, and compensatory images to achieve color correction and contrast enhancement. Jiang \cite{fiveA+} designed various modules to address complex degradation challenges.
	However, most CNN-based methods predominantly adopt multi-branch or modular architectures to separately address different types of degradations, focusing mainly on local feature processing. This local-centric design limits their ability to perform effective global modeling and hinders their performance in scenarios involving multiple strongly-coupled degradations.

	\subsection{Transformer-based Methods for UIE}
	
	Transformer architectures, leveraging global self-attention, have demonstrated superior capacity in modeling long-range dependencies compared to CNNs, which are limited by local receptive fields \cite{New1}. Recent studies have adapted transformer variants for underwater image enhancement (UIE), aiming to better capture global degradation patterns; Ren et al. \cite{New2} proposed a Swin transformer-based method with convolutional fusion (RSCTB) and a U-Net structure to jointly enhance and super-resolve underwater images. However, task coupling introduces interference, and the complex architecture struggles to balance detail preservation and computational efficiency, often leading to unnatural textures or edge artifacts. Zamir et al. \cite{New3} designed Restormer with multi-depth transposed attention (MDTA) and gated feedforward networks (GDFN), coupled with progressive learning. Although effective on natural scenes, it lacks underwater-specific design, resulting in inadequate adaptation to coupled degradations and diminished color fidelity. Gu et al. \cite{New4} introduced enhanced DeTrans, combining CNNs and transformers to exploit both local and global features. However, the method does not perform explicit modeling of degradation coupling, and insufficient feature fusion leads to unnatural transitions and edge discontinuities. Peng et al. \cite{New5} developed a U-shape transformer with CMSFFT and SGFMT modules, integrating multi-color-space loss to enhance chromatic consistency. Despite notable improvements, its global modeling lacks adaptability in extreme visibility conditions, often causing color distortion or unnatural textures. Dong et al. \cite{New6} proposed Feaformer within a multi-scale U-Net, fusing CNN-based local extraction and transformer-based global modeling through joint fusion and refinement modules. However, heavy reliance on physical priors can lead to local overenhancement and artificial artifacts.
	
	In summary, transformer-based methods improve global degradation modeling and demonstrate enhanced performance in UIE tasks. Nonetheless, they rely on complex designs for detailed recovery and exhibit poor generalizability in highly coupled degraded scenarios. Neglecting the potential information embedded in coupled degradation renders black-box models sensitive to data distribution, which limits their robustness in underwater coupled degradation environments.

	\subsection{Mamba-based Methods for UIE}
	
	Recent methods based on state space models (SSMs), such as Mamba \cite{New7}, leverage linear-complexity global modeling and long-range dependency capture for underwater image enhancement. These models aim to combine the efficiency of CNNs with the global capacity of transformers. Guan \cite{New8} proposed SCOSS and MSFFN to fuse global and local features; however, the adaptability of these models to non-uniform degradation is limited, often leading to detail blurring. Lin et al. \cite{New9} introduced a dual-level architecture (EMNet and PixNet) that balances context and detail; however, it suffers from unstable color correction and insufficient pixel-level smoothness. Dong \cite{New10} utilized a dual-branch O-shaped structure and MSBMP to model spatial and channel-wise information; however, fine-grained cross-channel degradation remains underaddressed. An et al. \cite{New11} combined convolution and SSM using CML and MAFM modules, enhancing feature expressiveness but increasing oversmoothing risks. Overall, although SSM-based models improve global coupling degradation modeling, their ability to handle coupled spatially heterogeneous distortions and local texture recovery remains suboptimal.

	Existing methods primarily adopt the following strategies to address coupling degradation in complex scenes: adjusting datasets for one-to-many mappings; designing dedicated processing strategies, network branches, or modules for different types of degradation; or relying purely on data-driven models to directly learn the degradation mappings.  
	Although these approaches have achieved promising results in specific degradation scenarios, they generally overlook the intrinsic mechanisms of nonlinear coupling and struggle to model the joint effects of multiple degradations comprehensively.
	Only a few studies have attempted to construct a unified enhancement framework from a bottom-up perspective to capture the interactive characteristics among various degradations. 
	Consequently, achieving balanced enhancement under coupled degradations remains a significant challenge, and existing methods often suffer from limited adaptability and generalization in complex environments.

	\begin{figure}[!t]
		\centering
		\includegraphics[width= 3.5in]{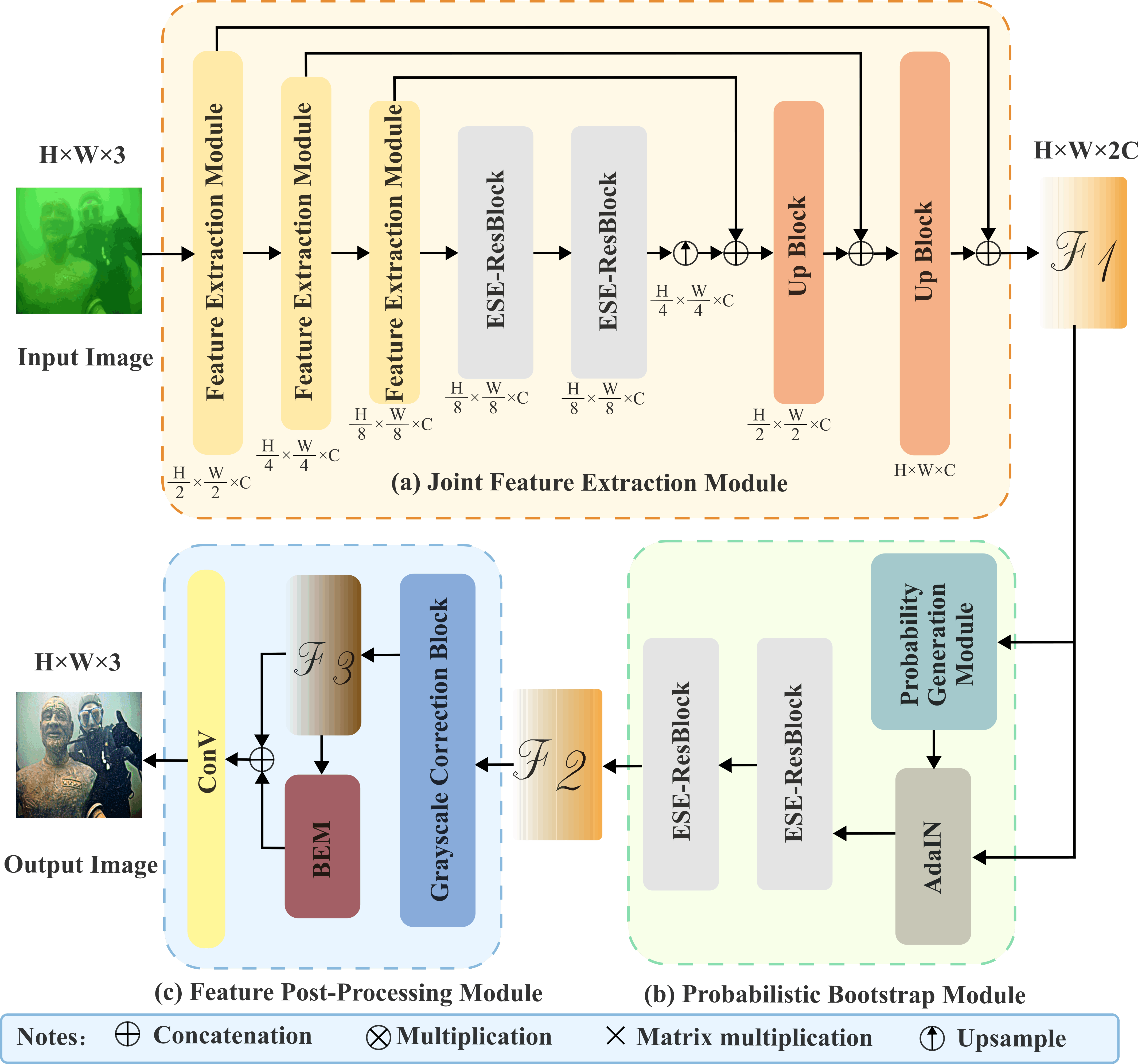}
		\caption{\textbf{Overall architecture of the JDPNet.} It comprises three main modules linked in sequence:  (a) JFE module, (b) PB module, and (c) FPP module.}
		\label{Overall architecture}
	\end{figure}

	\section{Proposed Method}
	In this section, we introduce the proposed method and explain how it addresses multiple coupled degradation scenarios to handle them effectively. We present a detailed description of the overall network architecture, components, and proposed loss function, encompassing the JFE module, PB module, FPP module, and the newly proposed loss function, AquaBalanceLoss.

	\subsection{Overall architecture}
	The overall network structure (Fig.~\ref{Overall architecture}) sequentially integrates the JFE module, PB module, and FPP modules. These components are designed to collaboratively mine, adjust, and optimize features in underwater images, ultimately enhancing the image quality. The JFE module functions as a feature-mining unit comprising an encoder--decoder architecture that extracts coupled latent feature maps ($F_1$) from the input images. These feature maps are processed using the PB module, which aligns the statistical properties of the content features with those of the training features, thereby preserving the original scene structure. The PB module efficiently generates images with the target features without requiring complex optimization. Finally, the FPP module refines the feature representation, addressing artifacts and boundary losses that may occur during network inference.

	\begin{figure}[!t]
		\centering
		\includegraphics[angle=0,width=3.5in]{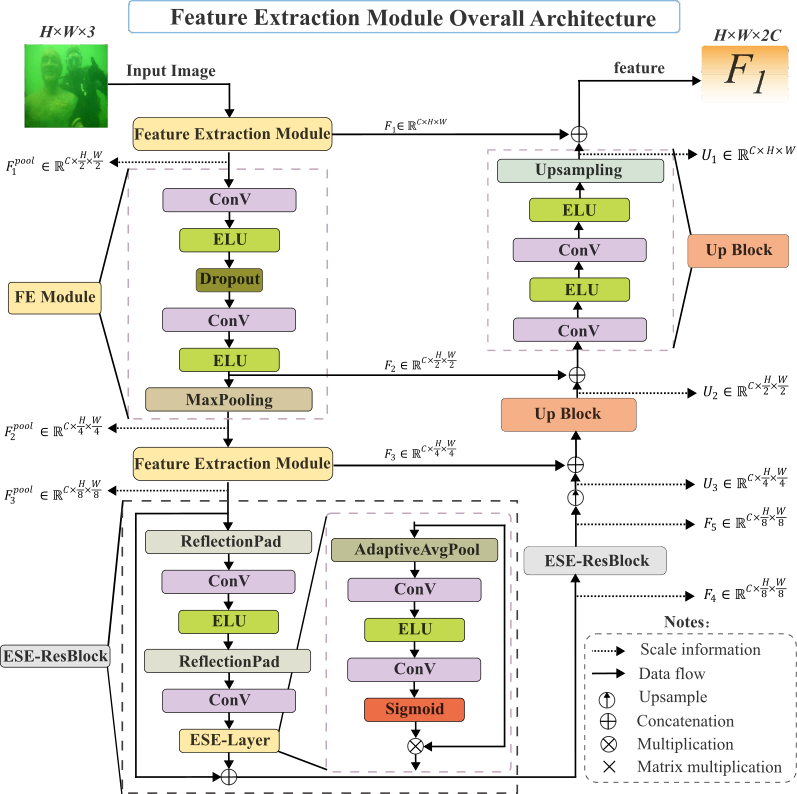}
		\caption{\textbf{Structure of JFE module.} The primary structure of the JFE module comprises: the feature extraction module (FE Module), ESE-ResBlock, and the Upsampling module, referred to as the Up Block. The input images, upon passing through the JFE module, result in the feature map $F_1$.}
		\label{JFE module}
	\end{figure}

	\subsection{Joint Feature Extraction Module}

	For CNNs, simply increasing the network depth, expanding the receptive field, or enlarging the solution space is not effective for capturing the necessary features, particularly in degraded underwater images, where capturing latent information significantly affects the final inference performance. To overcome this challenge, we proposed the JFE module. This component is structured as an encoder–-decoder comprising an FE module and ESE-ResBlock, both designed to extract latent features from input images (Fig.~\ref{JFE module}).
	
	In the proposed JFE module, the encoder is composed of three consecutive FE modules, each comprising six layers. Except for the first convolution, all convolutions had both the input and output channels set to 64. The spatial dimensions of the feature maps remain unchanged after convolutions and activations, and are only reduced by the MaxPooling operation. The output before pooling in the first FE module can be expressed as
	\begin{equation}
		\begin{split}
			F_1 &= \mathrm{ELU}\Big(\mathrm{Conv}\big(\mathrm{Dropout}(
			\mathrm{ELU}(\mathrm{Conv}(X, 3\times 3))) \big),\ 3\times 3 \Big), \\
			F_1 &\in \mathbb{R}^{C\times H\times W}
		\end{split}
	\end{equation}
	
	After MaxPooling:
	\begin{equation}
		F_1^{\mathrm{pool}} = \mathrm{MaxPool}(F_1),\quad F_1^{\mathrm{pool}} \in \mathbb{R}^{C\times \frac{H}{2} \times \frac{W}{2}}
	\end{equation}
	This process is repeated for the subsequent FE modules, yielding
	\begin{equation}
		F_2^{\mathrm{pool}} \in \mathbb{R}^{C\times \frac{H}{4} \times \frac{W}{4}},\quad
		F_3^{\mathrm{pool}} \in \mathbb{R}^{C\times \frac{H}{8} \times \frac{W}{8}}
	\end{equation}
	
	Given the complexity of underwater environments, enhancing the network's adaptive feature recalibration capability is crucial for extracting key coupled degradation features and suppressing irrelevant ones, thereby enabling a more efficient use of the model parameters. To address these challenges, we proposed the ESE-ResBlock (Fig.~\ref{JFE module}). 
	
	The processing procedure of ESE ResBlock can be defined as follows:

	\begin{align}
		F_4 =\ 
		&\mathrm{ESELayer}\Big( 
		\mathrm{Conv2d}\big( 
		\mathrm{ELU}(
		\mathrm{Conv2d}(F_3^{\mathrm{pool}},\, 3\!\times\!3)
		),\ 3\!\times\!3 
		\big) 
		\Big) \notag \\
		&+\, F_3^{\mathrm{pool}}
	\end{align}

	The ESELayer recalibrates channel-wise responses as
	\begin{equation}
		Y = X \odot \sigma \left(W_2 \varphi(W_1\, \mathrm{GAP}(X))\right)
	\end{equation}
	where $\mathrm{GAP}$ denotes global average pooling, $W_1$ and $W_2$ are $1 \times 1$ convolution weights, $\varphi$ is the ELU activation, $\sigma$ is the Sigmoid function, and $\odot$ denotes element-wise multiplication. ESE-ResBlock maintains the spatial size and enhances the nonlinear modeling capacity of the model, enabling the extraction and correction of complex coupled degradations.
	
	In the decoder, features are progressively upsampled and concatenated with encoder features at each scale, as follows:
	\begin{equation}
		U_3 = \mathrm{Up}(F_4), \quad U_3^{\mathrm{cat}} = \mathrm{Concat}[F_3, U_3]
	\end{equation}

	The subsequent upsampling and concatenation steps follow a similar structure, facilitating comprehensive multi-scale feature integration.
	
	The nonlinearity introduced by the ESE-ResBlock enhances the network's ability to model complex distortions and capture nonlinear features in underwater images, thereby improving its adaptability across diverse environmental conditions.  
	It facilitates dynamic weight allocation and smooth nonlinear transformations, enabling the network to better handle coupled degradations and correct nonlinear artifacts, ultimately enhancing generalization performance.
	
	Moreover, integrating ESE-ResBlock with the FE module ensures a smooth gradient flow, mitigates performance degradation due to vanishing gradients, and improves training stability and convergence speed. This is particularly critical for learning subtle features essential for underwater image restoration.

	\renewcommand{\algorithmicrequire}{\textbf{Input:}}
	\renewcommand{\algorithmicensure}{\textbf{Output:}}
	\begin{algorithm}
		\caption{PG module process}
		\label{PG algorithm}
		\begin{algorithmic}[1]
			\Require $ F_1 \in \mathbb{R}^{B \times 2C \times H \times W} $ \Comment{Input feature map}
			\Ensure $ \mu_{optimal}, \sigma_{optimal} $ \Comment{Final inputs for AdaIN}
			\State $\mu_f \gets \frac{1}{H \times W} \sum_{i=1}^{H} \sum_{j=1}^{W} f_{ij}$
			\State $\sigma_f \gets \sqrt{\frac{1}{H \times W} \sum_{i=1}^{H} \sum_{j=1}^{W} (f_{ij} - \mu_f)^2}$
			\State $a \gets Conv(\mu)$ \Comment{Using 1x1 convolution}
			\State $m \gets ConV(\mu)$ \Comment{1x1 Conv}
			\State $b \gets ConV(\sigma)$ \Comment{1x1 Conv}
			\State $n \gets ConV(\sigma)$ \Comment{1x1 Conv}
			\State $N_{mean} \gets \mathcal{N}(a, m^2)$ \Comment{Construct an n-dimensional Gaussian distribution for mean}
			\State $N_{std} \gets \mathcal{N}(b, n^2)$ \Comment{Construct an n-dimensional Gaussian distribution for std}
			\State $\mu_{optimal} \gets ExtractMean(N_{mean})$
			\State $\sigma_{optimal} \gets ExtractStd(N_{std})$
			
		\end{algorithmic}
	\end{algorithm}

	\subsection{Probabilistic Bootstrap Module}
	
	In the context of UIE, merely mining and decoupling the latent features of coupling degradation is insufficient for satisfactory enhancement.  
	To better model the global characteristics of coupled degradations and align the content and training features for a unified degradation adjustment, we drew inspiration from the effectiveness of AdaIN \cite{AdaIN} in style transfer and introduced a PB module.
	
	This module adaptively refines the image color, contrast, and structure through style modulation and feature recalibration while preserving the original scene layout (Fig.~\ref{Overall architecture}).  
	By integrating ESE-ResBlock with AdaIN \cite{AdaIN}, our approach enhances flexibility in handling complex degradations and maintains semantic consistency, ultimately yielding results that closely resemble in-air or shallow-water imaging.
	
	To further optimize the input features within the proposed PB module, we introduced the PG module to construct the required mean and standard deviation distributions. As shown in Algorithm\ref{PG algorithm}, we set the feature map output by the JFE module as the data matrix $F_1$, where $F_1 \in \mathbb{R}^{B \times 2C \times H \times W}$, and $B, C, H, W$ represent the batch size, number of channels, height, and width, respectively. First, for feature map $F_1$, we computed the mean and standard deviation of each channel of $f$:
	
	\begin{equation}
		\mu_f = \frac{1}{H \times W} \sum_{i=1}^{H} \sum_{j=1}^{W} f_{ij}
	\end{equation}

	\begin{equation}
		\sigma_f = \sqrt{\frac{1}{H \times W} \sum_{i=1}^{H} \sum_{j=1}^{W} (f_{ij} - \mu_f)^2}
	\end{equation}

	Furthermore, we extracted features from the mean vector using a 1×1 convolution kernel, resulting in 
	\begin{equation}
		a = ConV(\mu) , a \in \mathbb{R}^{B \times N \times 1 \times 1}
	\end{equation} 
	and standard deviation 
	\begin{equation}
		m = ConV(\mu) , m \in \mathbb{R}^{B \times N \times 1 \times 1}
	\end{equation}

	Similarly, we used a 1×1 convolution kernel to extract features from the standard deviation vector to obtain $b \in \mathbb{R}^{B \times N \times 1 \times 1}$ and standard deviation $n \in \mathbb{R}^{B \times N \times 1 \times 1}$.
	
	Following this, we extended the application of the Gaussian distribution to optimize and adjust the input for AdaIN \cite{AdaIN}. We used the extracted mean $a$ and standard deviation $m$ parameters to construct an n-dimensional Gaussian distribution, denoted by $N_{\text{mean}} = \mathcal{N}(a, m^2)$, to adjust the mean of the data. 
	Similarly, we used $b$ and $n$ to construct another n-dimensional Gaussian distribution, denoted by $N_{\text{std}} = \mathcal{N}(b, n^2)$, to adjust the standard deviation of the data.
	Finally, we extracted the means from both $N_{\text{mean}}$ and $N_{\text{std}}$ Gaussian distributions, reused them as the optimal unique mean and standard deviation, and formed the input for AdaIN \cite{AdaIN}:
	\begin{equation}
		\mu_{\text{optimal}} = ExtractMean(N_{\text{mean}})
	\end{equation} 
	and standard deviations. 
	\begin{equation}
		\sigma_{\text{optimal}} = ExtractStd(N_{\text{std}})
	\end{equation} 
	where $\text{ExtractMean}()$ and $\text{ExtractStd}()$ represent the operations for extracting the mean and standard deviations from a Gaussian distribution, respectively.
	
	After global adjustments, the integrated ESE-ResBlock continues to guide the network to focus on key features, enhancing the consistency of the AdaIN \cite{AdaIN} effect and improving the adaptability of the model to various underwater degradation conditions.

	\subsection{Feature Post-Processing Module}

	Although AdaIN \cite{AdaIN} enables flexible style modulation, it can also lead to excessive smoothing when the mean and variance adjustments are overapplied, resulting in the loss of fine details and color fidelity.  
	In our framework, this effect was further amplified after sequential processing by the JFE and PB modules; this lead to the occasional introduction of artifacts and boundary distortions in the final output.
	
	To mitigate these issues, we introduced an FPP module that refines feature representations based on existing feature maps.  
	It comprises a grayscale correction component and a BEM mechanism. The grayscale correction component compensates for over-adjustment by earlier layers, preserving tonal balance, while the BEM leverages Gaussian modulation to enhance structural detail and alleviates artifacts and boundary shifts introduced during inference.

	The core objective of the grayscale correction module is to achieve a more balanced and natural overall grayscale distribution in the output image by adaptively adjusting the brightness ratio of each channel. The module first calculates the global or local grayscale mean of the image and then scales each channel to bring its mean closer to the target grayscale value. The mathematical expression is as follows:
		
	\begin{equation}
		I_c^{\prime}(x,y) = I_c(x,y) \cdot \frac{\mu_g}{\mu_c}
	\end{equation}
	where $I_c(x,y)$ represents the brightness value of pixel $(x,y)$ in the $c$th channel, $\mu_c$ is the channel mean, and $\mu_g$ is the global target grayscale mean. The introduction of the grayscale correction module improved the subjective visual quality and color accuracy of the underwater images to a certain extent.
	
	Let the feature map obtained through the PB module be denoted as $F_2$, which, upon entering the FPP module and undergoing white balance processing, results in $F_3$. We apply a Gaussian filter to $F_3$ to extract its low-frequency components, which are denoted by $F_{low}$:
	\begin{equation}
		F_{\text{low}} = G(F_3, \omega)
	\end{equation}
	where \(G\) represents the Gaussian filter, and \(\omega\) is the balancing parameter for a scale of \(G\).
	
	The BEM \(F_{\text{BEM}}\) is then formed by
	\begin{equation}
		F_{\text{BEM}} = F_3 - F_{\text{low}} + \lambda
	\end{equation}
	where \(\lambda\) is used to maintain the BEM intensity.
	Finally, merging the BEM with $F_3$ results in an adjusted feature map.
	\begin{equation}
		{F_P} = \begin{cases}
			\frac{F_3F_{BEM}}{\lambda},& \text{if } F_{BEM} < \lambda  \\ 
			{1-\frac{(1-F_3)(1-F_{BEM})}{\lambda},}&{\text{otherwise.}} 
		\end{cases}
	\end{equation}
	
	After convolutional adjustments, the final module output is obtained, which serves as the final output image of the network, that is, an enhanced underwater image. Thus, the network inference process is complete.

	\subsection{AquaBalanceLoss}
	
	To achieve a better balance between color, sharpness, and contrast and to ensure that the network can effectively utilize coupled degradation information from the underlying mechanism, we propose a novel loss function, AquaBalanceLoss. This is a newly designed loss function tailored to the characteristics of underwater images. The function effectively constrains the network to proficiently learn the mapping between degraded features and ground-truth images with a focus on minimizing losses in luminance, contrast, and clarity. The basic expression for AquaBalanceLoss is as follows:
	\begin{equation}
		AquaBalanceLoss=|AbL_{out}-AbL_{in}+ \lambda_{imp} |^2
	\end{equation}
	where \( AbL_{\text{out}} \) represents the value of the output image after the relevant calculations and \( AbL_{\text{in}} \) denotes the value of the input image following similar computations. $\lambda_{imp}$ represents the bias amount.
	
	Next, we elucidate how the AbL is calculated. AbL comprises three components: the color index ($L_{coi}$), sharpness index ($L_{si}$), and contrast index ($L_{cti}$). The basic expression for AbL is as follows:
	\begin{equation}
		AbL = c_1 \times L_{coi} + c_2 \times L_{si} + c_3 \times L_{cti} \quad 
	\end{equation}
	where $c_1$, $c_2$, and $c_3$ denote the weights attributed to each property, with $c_1 = 0.029$, $c_2 = 0.295$, and $c_3 = 3.550$.

	We mapped different attributes onto a unified loss space, ensuring that each network parameter update during backpropagation is influenced by the aggregated gradients of all three attribute components. This effectively avoids solutions in which one attribute is optimized at the expense of others and enables a Pareto front approximation for multi-attribute optimization problems. Consequently, the output image achieves an optimal solution for each attribute that cannot be further improved without compromising other attributes.

	For $L_{coi}$, we used the opposing color spaces RG and YB because the colors in underwater images are mainly expressed as an imbalance between these two opposing color ranges. The formula is given by
	\begin{equation}
		L_{coi} = -0.027 \times l + 0.159 \times r
	\end{equation}
	
	The basic expressions for $l$ and $r$ are as follows: 
	\begin{equation}
		l = \sqrt{\mu_a(R_G)^2 + \mu_a(Y_B)^2}
	\end{equation}
	and
	\begin{equation}
		r = \sqrt{s_a(R_G)^2 + s_a(Y_B)^2}
	\end{equation}
	where \(\mu_a(\cdot)\) represents the alpha-trimmed mean of the pixel values, \(s_a(\cdot)\) represents the variance of the pixel values after applying alpha-trimming, which is the sum of squared deviations from the alpha-trimmed mean divided by the number of pixels.

	The $R_G$ and $Y_B$ color components can be obtained from
	\begin{equation}
		R_G = R - G, \quad Y_B = \frac{R + G}{2} - B \quad 
	\end{equation}
	
	For $L_{si}$, the sharpness of the image is evaluated by computing the weighted sum of the edge maps obtained after applying the Sobel edge detection to the image. The formula used is as follows:
	\begin{equation}
		L_{si} =\sum_{i=1}^{c} \lambda_i EME_i
	\end{equation}
	where \(EME_i\) denotes the result of applying Sobel edge detection to a channel and computing the intensity of the channel edge map (EME). The symbol $\lambda_i$ denotes the channel-weighting coefficient.

	$L_{cti}$ assesses the contrast of underwater images by computing the contrast within the image blocks. It is expressed as follows:
	\begin{equation}
		L_{cti} = w \times \sum(\alpha  \left( \frac{top}{bot} \right)^\alpha  \log\left(\frac{top}{bot}\right)) 
	\end{equation}
	where $top = max - min, \quad bot = max + min$. represent the maximum and minimum values, respectively, for each image block. Weight $\omega$ is defined as $-\frac{1}{(k1 \times k2)}$ and $\alpha$ is a parameter used to adjust the entropy scale. Increasing $\alpha$ increases the randomness of the assessment.

	\subsection{Discussion of Gradient angle on AquaBalanceLoss}
	
	Detailed analysis can be found in Appendix A.

	\subsection{Loss Functions}
	The loss function plays a critical role in guiding the network to improve the underwater image quality in terms of clarity, color fidelity, and visual detail.  
	Therefore, we propose a novel multi-objective loss function, AquaBalanceLoss, which integrates reconstruction loss and Kullback--Leibler (KL) divergence loss to jointly supervise the enhancement process.
	
	The reconstruction loss constrains the output to remain close to the ground truth, encouraging the network to learn effective mappings from degraded inputs to high-quality targets. 
	Meanwhile, the KL divergence loss aligns the learned feature distribution with that of the reference promoting the generation of perceptually natural and structurally coherent images.
	
	By jointly optimizing both objectives, AquaBalanceLoss enables the network to better model coupled degradation features, achieving a balanced enhancement across color, clarity, and contrast. The complete loss formula is defined as follows:
	
	\begin{equation}
		\begin{split}
			Loss = &\lambda_1 \cdot Loss_{vgg16} + \lambda_2 \cdot Loss_{KL} \\
			&+ \lambda_3 \cdot Loss_{Re} + \lambda_4 \cdot AquaBalanceLoss  
		\end{split}
	\end{equation}
	where $\lambda_1, \lambda_2, \lambda_3 \text{ and } \lambda_4 \text{ are } 0.025, 1, 0.1, \text{ and } 0.1, \text{ respectively.}$.

	This composite loss function effectively optimizes our model by capturing diverse image attributes and generating high-quality underwater images.

	\begin{figure*}[!t]
		\centering
		\includegraphics[width=\linewidth]{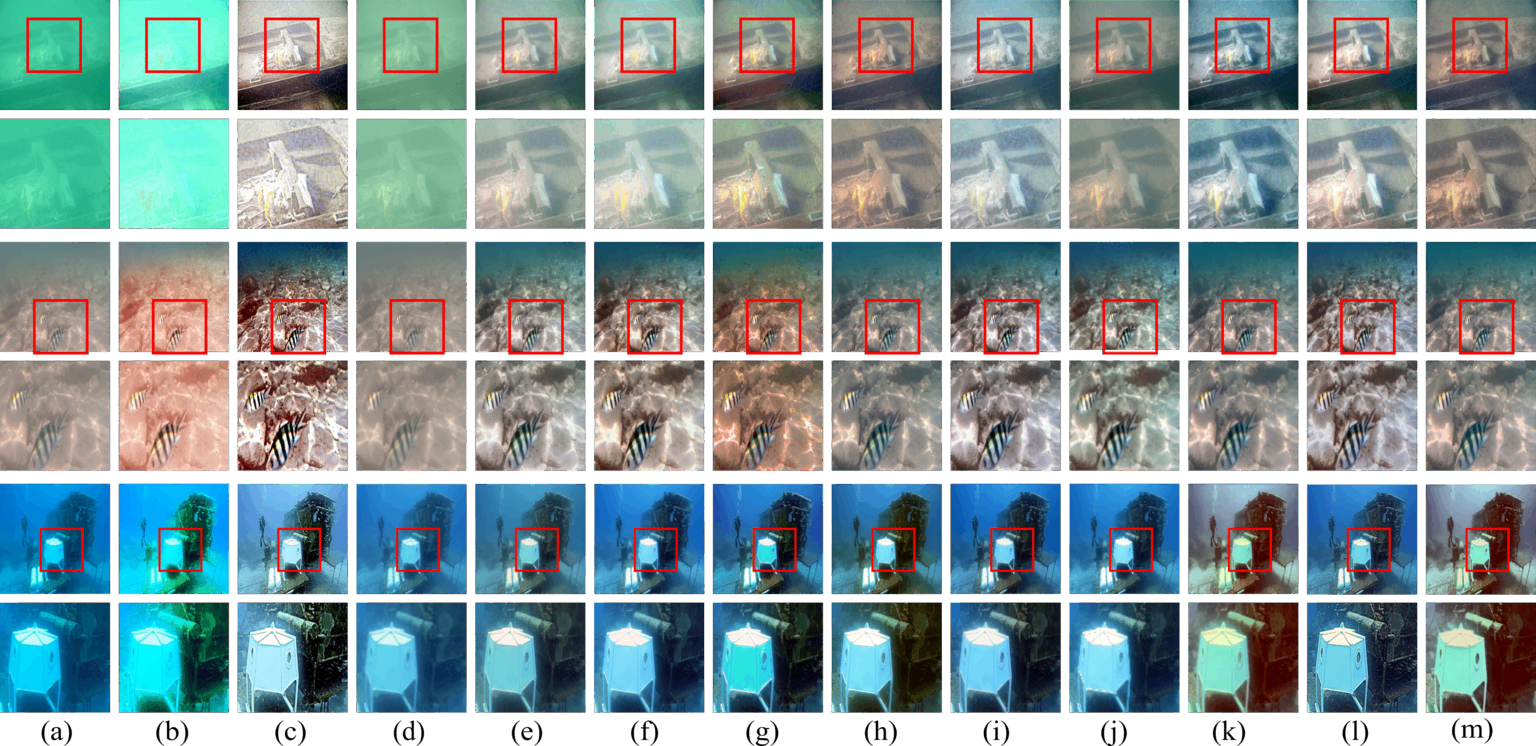}
		\caption{Visual comparison of different methods on UIEB dataset \cite{UIEB}. The visual comparison of different methods from left to right includes (a)Input, (b)GDCP\cite{ref40}, (c)HS2CM2A \cite{HS2CM2A},  (d)Shallow-UWnet \cite{ref64},  (e)PUIE-MC \cite{PUIE}, (f)FiveAPlus \cite{ref66}, (g)CWR \cite{CWR},  (h)Spectroformer \cite{spectroformer} ,(i)NU2Net\cite{NU2Net}, (j)X-caunet\cite{New1}, (k)WaterMamba\cite{New8}, (l)our method, and (m)ground truth.}
		\label{UIEB}
	\end{figure*}

	\begin{table*}[!t]
		\centering
		\small 
		
			\caption{Quantitative method comparisons across UIEB \cite{UIEB} datasets. The highest value is in bold, and the second-highest is underlined.}
			
			\label{UIEB_Data}
			\setlength{\tabcolsep}{8pt}
			\begin{tabularx}{\textwidth}{lYYYYYY}
				\toprule
				\textbf{Method} & \textbf{PSNR ↑} & \textbf{SSIM ↑} & \textbf{UIQM ↑} & \textbf{UCIQE ↑} & \textbf{CCF ↑} & \textbf{PCQI ↑} \\
				\midrule
				GDCP          & 14.200 & 0.739 & 2.203 & 0.602 & 33.270 & 0.939 \\
				UDCP          & 12.539 & 0.618 & 1.957 & 0.588 & 38.218 & 0.467 \\
				IBLA          & 17.036 & 0.751 & 2.840 & 0.602 & \uline{39.041} & 0.679 \\
				HS2CM2A       & 13.637 & 0.647 & 2.427 & 0.621 & \textbf{43.736} & 0.896 \\
				WaterNet      & 19.340 & 0.832 & 2.818 & 0.571 & 16.600 & 0.698 \\
				Shallow-uwnet & 17.648 & 0.654 & 2.997 & 0.537 & 19.311 & 0.731 \\
				UWCNN         & 15.400 & 0.775 & 2.126 & 0.506 & 15.073 & 0.773 \\
				Ucolor        & 20.710 & 0.841 & 2.928 & 0.579 & 21.553 & 0.825 \\
				PUIE-MC       & 22.180 & 0.888 & 2.947 & 0.589 & 19.465 & 0.926 \\
				CWR           & 21.223 & 0.853 & 2.817 & 0.608 & 16.893 & 0.714 \\
				FiveAPlus     & 22.142 & \uline{0.912} & 2.903 & 0.612 & 23.911 & \uline{0.964} \\
				Spectroformer & 23.004 & 0.907 & 2.924 & \uline{0.614} & 28.895 & 0.776 \\
				NU2Net        & 22.459 & 0.884 & \uline{3.005} & 0.592 & 20.693 & 0.962 \\
				X-caunet      & 21.408 & 0.789 & 2.807 & 0.602 & 22.772 & 0.740 \\
				WaterMamba    & \textbf{28.023} & 0.849 & 2.745 & 0.610 & 27.271 & 0.746 \\
				Ours          & \uline{23.007} & \textbf{0.915} & \textbf{3.045} & \textbf{0.626} & 29.908 & \textbf{0.979} \\
				\bottomrule
			\end{tabularx}
		
	\end{table*}

	\begin{table*}[!t]
		\centering
		\small
		
			\caption{Quantitative method comparisons across QDH and ROV datasets. The highest value is in bold, and the second-highest is underlined.}
			\label{qdh_rov_Data}
			
			\begin{tabularx}{\textwidth}{l|YYYY|YYYY}
				\toprule
				\multirow{2}{*}{Method} 
				& \multicolumn{4}{c|}{QDH} 
				& \multicolumn{4}{c}{ROV} \\
				& UIQM ↑ & UCIQE ↑ & CCF  ↑ & Running time/s
				& UIQM ↑ & UCIQE ↑ & CCF  ↑ & Running time/s \\
				\midrule
				GDCP         & 1.440 & 0.465 & 9.105   & 2.120   & 2.529 & 0.565 & 22.25  & 10.20  \\
				UDCP         & 0.895 & 0.483 & 27.43   & 37.530  & 1.101 & 0.536         & 26.03 & 146.67 \\
				IBLA         & 0.837 & 0.500 & \uline{30.50}  & 5.490   & 2.027 & 0.524         & \uline{27.54}  & 28.83  \\
				HS2CM2A      & 2.606 & \uline{0.619} & 29.57   & 9.430   & 2.707 & \uline{0.581} & 22.94  & 16.78  \\
				PUIE-MC      & 2.029 & 0.516 & 11.74   & 0.027  & 2.302 & 0.529         & 12.20  & 0.109  \\
				CWR          & \uline{2.875} & 0.496 & 12.43   & 1.150   & \textbf{3.205} & 0.478   & 13.58  & 0.812  \\
				FiveAPlus    & 2.105 & 0.536 & 12.65   & 0.058  & 2.368 & 0.541         & 11.80  & 0.219  \\
				Spectroformer& 2.607 & 0.500 & 15.75   & 0.061  & 2.920 & 0.541         & 22.04  & 0.067  \\
				X-caunet     & 1.927 & 0.527 & 11.41   & 0.108  & 1.668 & 0.521         & 9.09   & 0.129  \\
				WaterMamba   & 2.273 & 0.554 & 16.56   & 0.025  & 1.753 & 0.546   & 12.53  & 0.028  \\
				Ours         & \textbf{2.927} & \textbf{0.624} & \textbf{31.91} & 0.030
				& \uline{3.006} & \textbf{0.592} & \textbf{29.73}  & 0.091  \\
				\bottomrule
			\end{tabularx}
	\end{table*}

	\section{Experiments}
	This section presents a comprehensive evaluation of the proposed underwater image enhancement model. By comparing it with current state-of-the-art UIE methods, we undertook both quantitative and qualitative analyses from multiple dimensions to assess the effectiveness and superiority of our approach.
	
	\subsection{Datasets}
	In underwater image enhancement research, diverse and challenging datasets are essential for testing and validating enhancement algorithms. Our experiments covered multiple datasets to ensure a comprehensive evaluation and generalizability of our network performance. For a thorough comparative analysis, we used the real-world underwater image enhancement benchmark (UIEB \cite{UIEB}), which contains over 890 high-quality underwater images covering a wide range of underwater scenes, for training.
	
	To comprehensively test the network performance and generalizability in real-world scenarios, we utilized real underwater datasets C60, U45 \cite{U45}, UCCS \cite{UCCS},  EUVP \cite{EUVP} , and LSUI \cite{New5} for testing. U45 \cite{U45} was divided into three subsets based on the color biases from underwater degradation, low contrast, and blurriness: green, blue, and haze. The UCCS dataset \cite{UCCS} comprised 300 real underwater images encompassing blue, green, and blue-green environments and offers diverse marine biology and environmental conditions for analysis.
	C60 is a challenging real-world subset from UIEB containing 60 underwater images without reference images, used for evaluating generalization under difficult conditions.
	EUVP~\cite{EUVP} is the Enhancing Underwater Visual Perception dataset, including both paired and unpaired real underwater images of low/high perceptual quality, and is widely used for supervised and unsupervised UIE evaluation.
	LSUI~\cite{New5} is a large-scale dataset of about 5k real-world underwater image pairs covering diverse scenes and water types, with corresponding clear reference images for enhancement benchmarking.

	Furthermore, to assess the network performance in complex environments, we developed two new underwater image datasets: the ROV and QDH datasets. 
	The statistical information of the ROV and QDH datasets is presented in Appendix B.

\begin{figure}[h]
	\centering
	\includegraphics[width=\linewidth]{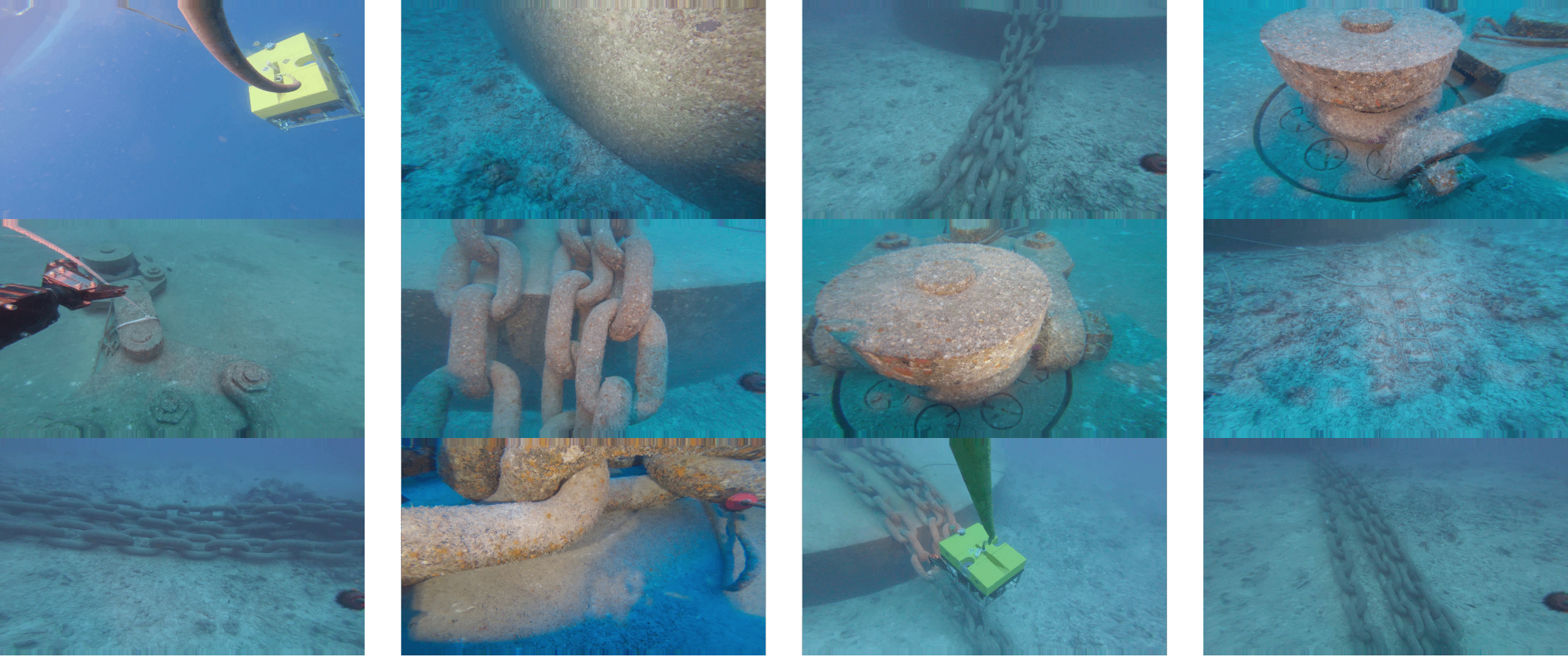}
	\caption{ROV dataset scene examples}
	\label{ROV-example}
\end{figure}

	The ROV dataset contains 7,066 images captured by remotely operated vehicles in various marine (saltwater) environments. The data cover different depths, illumination (natural light and sodium-lamp lighting), and turbidity levels, with water conditions ranging from clear to moderate and heavy haze. Typical scenes include seabeds, infrastructures, pipelines, chains, and other industrial structures, with main objects such as ROV equipment, marine facilities, and seafloor debris. All images are recorded at 4K resolution. Representative samples are shown in Fig. \ref{ROV-example}.
	
\begin{figure}[h]
	\centering
	\includegraphics[width=\linewidth]{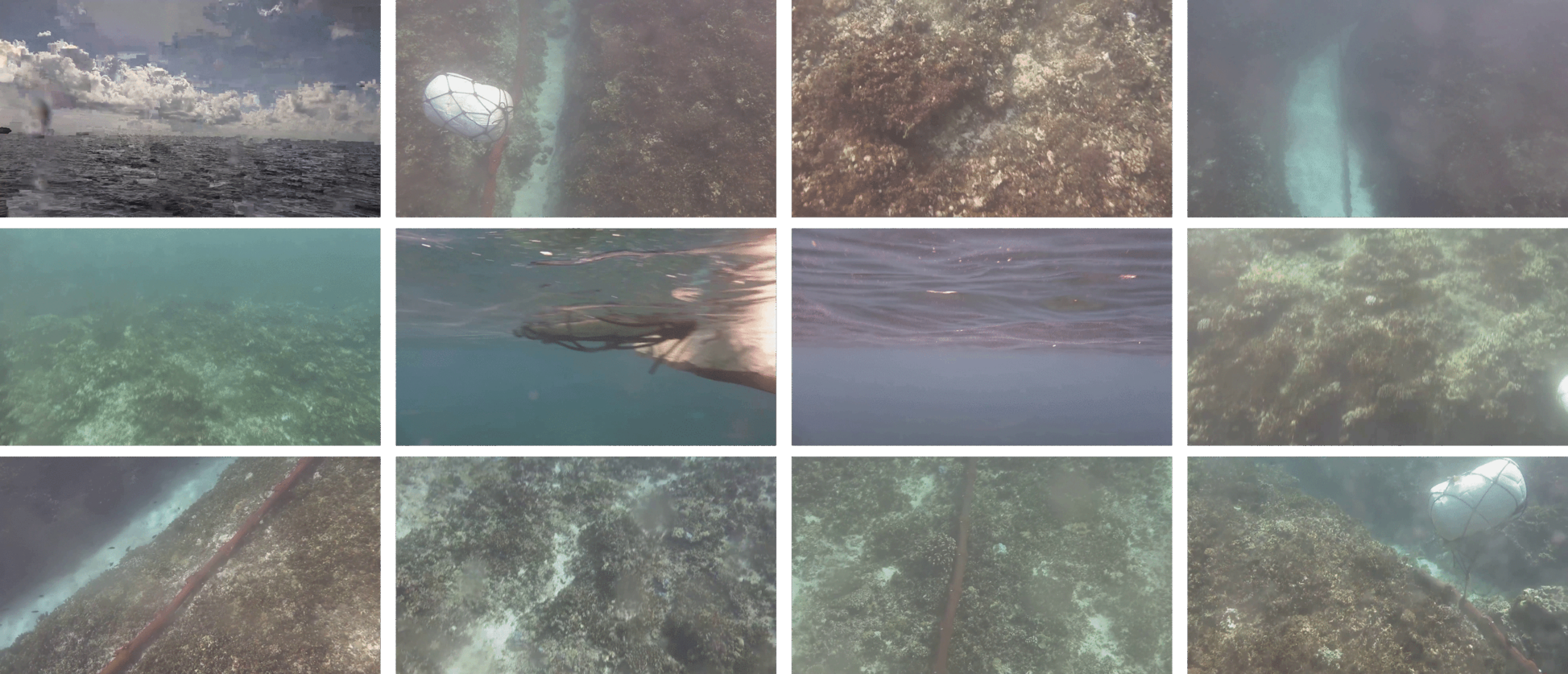}
	\caption{QDH dataset scene examples}
	\label{QDH-example}
\end{figure}

	The QDH dataset comprises 6,798 images collected by underwater robots in China’s Thousand Island Lake (freshwater). It exhibits diverse scenes and visibility conditions, with water ranging from clear to moderately and heavily turbid. Illumination is mainly normal to low, and several scenes are illuminated by sodium lamps, which further increase color inconsistency. The scenes cover natural lakebeds, submerged structures, aquaculture-related facilities, and other inland underwater environments, with main objects including man-made structures, underwater devices, and debris. All images are captured at 1080P resolution, as shown in Fig. \ref{QDH-example}.

	We conducted experiments on all the aforementioned datasets to thoroughly validate the effectiveness of the model in processing different water bodies and coupled degradation scenarios.

	\subsection{Training Details}
	Experiments were conducted on a system running Ubuntu 22.04, equipped with an Intel Core i9-13900K CPU and an RTX 4090 24GB GPU. The software environment included Python 3.10, PyTorch \cite{PyTorch} 2.1.0, and CUDA 11.8 support. Training was conducted for 300 epochs with a batch size of one using Adam as the optimization algorithm. The learning rate was initially set to $2 \times 10^{-4}$ and was adjusted to $1 \times 10^{-4}$ after 50 epochs. The data augmentation involved horizontal flipping and random rotations of 90°, 180°, and 270°. The input data were randomly cropped into 256×256 patches during the training.

	\subsection{Evaluation Indicators}
	For a comprehensive evaluation of the proposed underwater image enhancement method, we selected commonly used quantitative metrics to assess the image quality. The metrics include peak signal-to-noise ratio (PSNR \cite{PSNR}), structural similarity index (SSIM \cite{SSIM}), underwater image quality measure (UIQM \cite{UIQM}), underwater color image quality evaluation (UCIQE \cite{UCIQE}), colorfulness, contrast, and fog density (CCF) and patch-based contrast quality index (PCQI). In our experiments, these metrics were used to quantitatively analyze the image quality after enhancement using different methods. PSNR \cite{PSNR} and SSIM \cite{SSIM} assess the overall image quality and visual similarity, whereas UIQM \cite{UIQM} and UCIQE \cite{UCIQE} address unique quality issues in underwater images. A higher CCF indicates that the image is more colorful, has stronger contrast, and is clearer. A higher PCQI value indicates that the enhanced image is more similar to the reference high-quality image in terms of local contrast, structural details, and brightness consistency, reflecting superior enhancement performance.
	A comprehensive analysis of these metrics provided a thorough understanding of the performance of the proposed method.

	\subsection{Comparison with SOTA Methods}
	We compared our proposed method both qualitatively and quantitatively against current state-of-the-art underwater image enhancement techniques, including  UDCP\cite{UDCP}, GDCP\cite{ref40}, IBLA \cite{IBLA}, HS2CM2A \cite{HS2CM2A}, Water-Net \cite{ref60}, Shallow-UWnet \cite{ref64}, UWCNN \cite{ref61}, Ucolor \cite{ref65}, PUIE \cite{PUIE}, CWR \cite{CWR}, FiveAPlus \cite{ref66}, Spectroformer \cite{spectroformer}, NU2Net \cite{NU2Net}, X-caunet\cite{New1} and WaterMamba\cite{New8}. The qualitative results are shown in Fig.~\ref{UIEB} and Appendix B; the quantitative comparisons across multiple datasets are presented in Tables \ref{UIEB_Data} and Appendix B. 
	Our approach demonstrated competitive performance in maintaining color balance, visibility, and handling complex coupled degradation compared with other methods.

	\subsubsection{Qualitative comparison}
	We did not aim for complete consistency with the ground truth (GT) because of the subpar quality of some GT data. Instead, we focused on achieving more visually appealing natural properties in the images. In a few extremely low-visibility cases, slight residual color bias or detail smoothing may remain, which is mainly caused by the severe loss of local cues under strong scattering. All visual comparisons can be seen in Figure \ref{UIEB} and Appendix C.

\begin{figure*}[t]
	\centering
	\includegraphics[width=\linewidth]{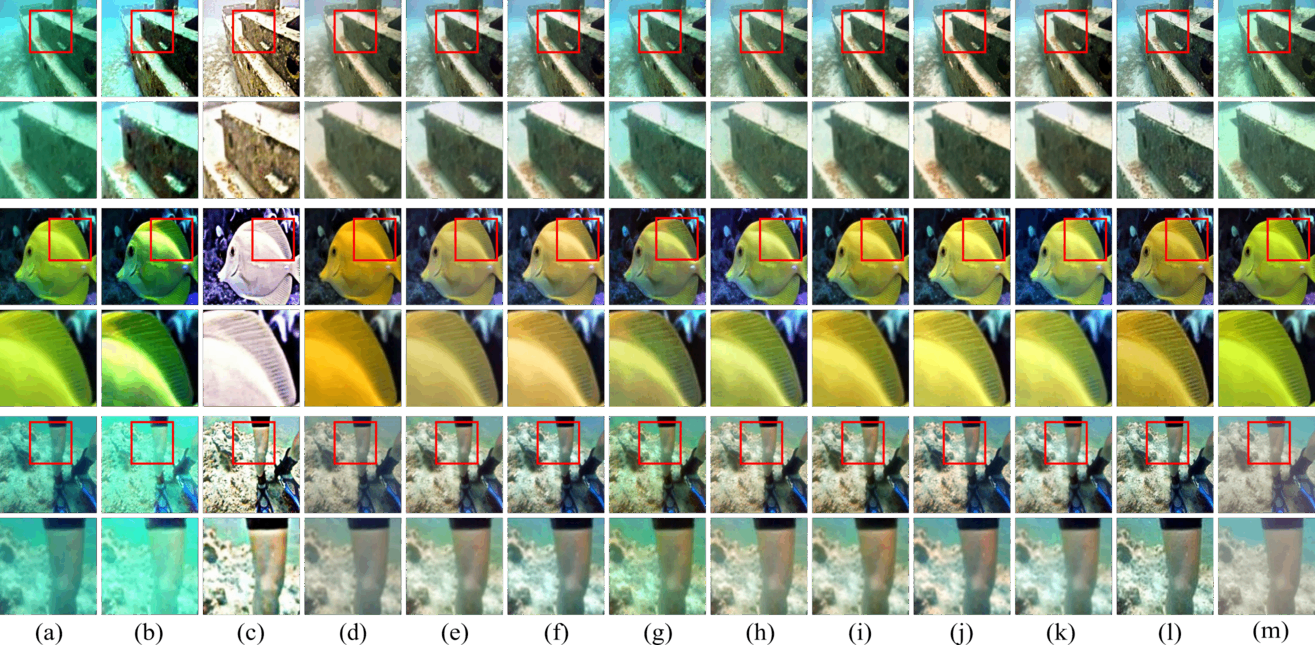}
	\caption{Visual comparison of different methods on the EUVP dataset \cite{EUVP}. The visual comparison of different methods from left to right includes (a)Input, (b)GDCP\cite{ref40}, (c)HS2CM2A \cite{HS2CM2A},  (d)Shallow-UWnet \cite{ref64},  (e)PUIE-MC \cite{PUIE}, (f)FiveAPlus \cite{ref66}, (g)CWR \cite{CWR},  (h)Spectroformer \cite{spectroformer} ,(i)NU2Net\cite{NU2Net}, (j)X-caunet\cite{New1}, (k)WaterMamba\cite{New8}, (l)our method, and (m)ground truth.}
	\label{EUVP}
\end{figure*}

\begin{figure*}[t]
	\centering
	\includegraphics[width=\linewidth]{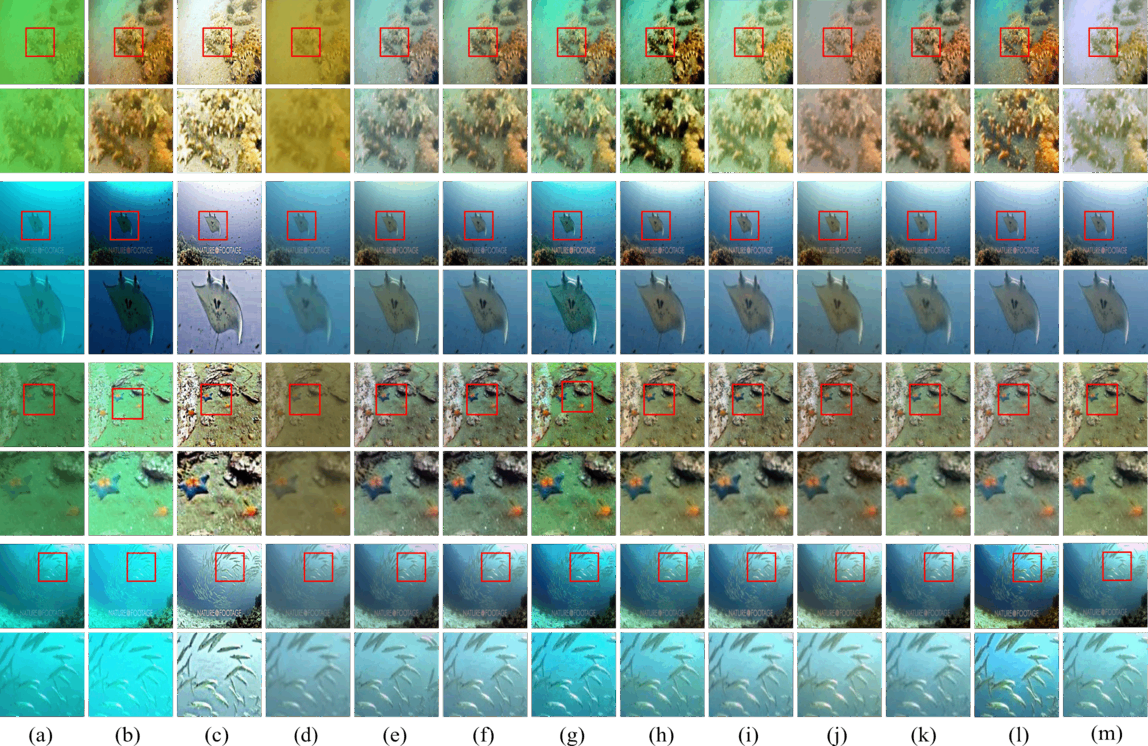}
	\caption{Visual comparison of different methods on the LSUI dataset\cite{New5}. The visual comparison of different methods from left to right includes (a)Input, (b)GDCP\cite{ref40}, (c)HS2CM2A \cite{HS2CM2A},  (d)Shallow-UWnet \cite{ref64},  (e)PUIE-MC \cite{PUIE}, (f)FiveAPlus \cite{ref66}, (g)CWR \cite{CWR},  (h)Spectroformer \cite{spectroformer} ,(i)NU2Net\cite{NU2Net}, (j)X-caunet\cite{New1}, (k)WaterMamba\cite{New8}, (l)our method, and (m)ground truth.}
	\label{LSUI}
\end{figure*}

\begin{figure*}[t]
	\centering
	\includegraphics[width=\linewidth]{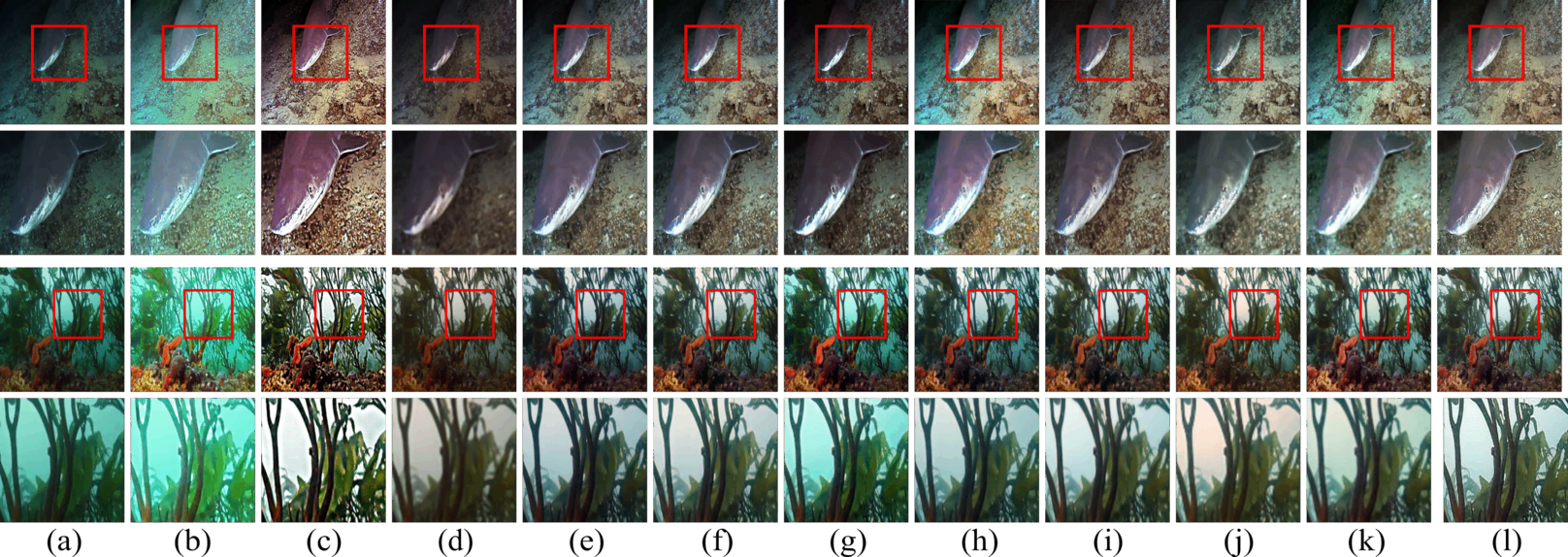}
	\caption{Visual comparison of different methods on the C60 dataset. The visual comparison of different methods from left to right includes (a)Input, (b)GDCP\cite{ref40}, (c)HS2CM2A \cite{HS2CM2A},  (d)Shallow-UWnet \cite{ref64},  (e)PUIE-MC \cite{PUIE}, (f)FiveAPlus \cite{ref66}, (g)CWR \cite{CWR},  (h)Spectroformer \cite{spectroformer} ,(i)NU2Net\cite{NU2Net}, (j)X-caunet\cite{New1}, (k)WaterMamba\cite{New8}, and (l)our method.}
	\label{C60}
\end{figure*}

\begin{figure*}[t]
	\centering
	\includegraphics[width=\linewidth]{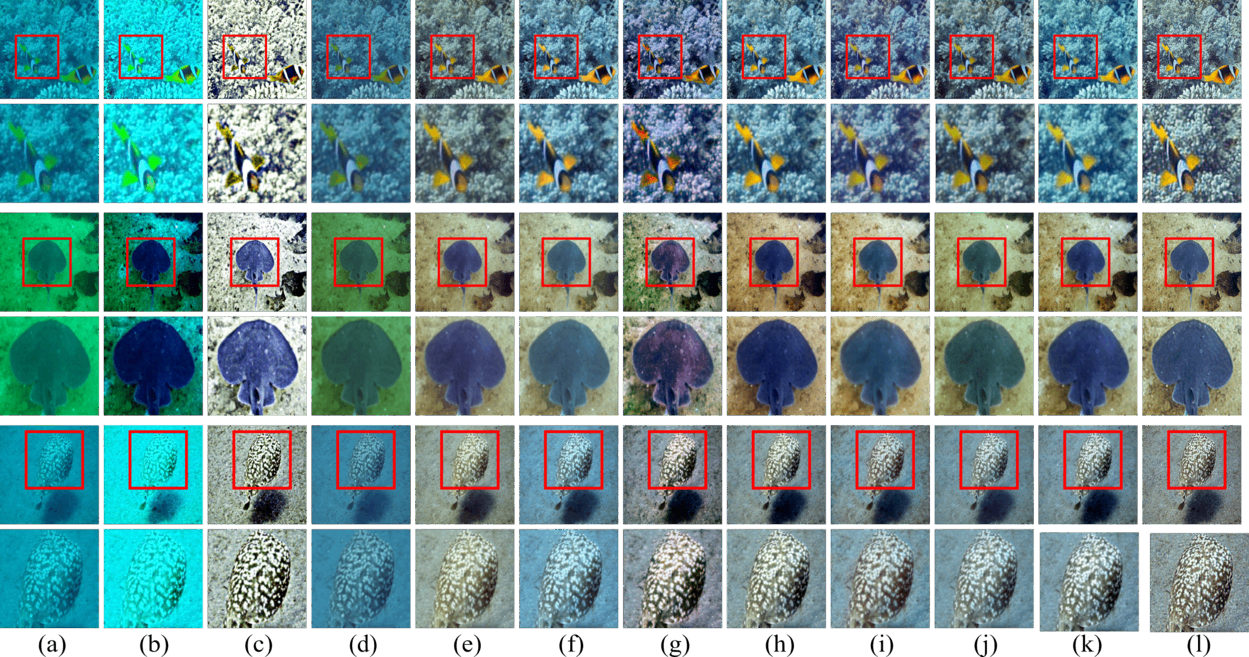}
	\caption{Visual comparison of different methods on the U45 dataset. The visual comparison of different methods from left to right includes (a)Input, (b)GDCP\cite{ref40}, (c)HS2CM2A \cite{HS2CM2A},  (d)Shallow-UWnet \cite{ref64},  (e)PUIE-MC \cite{PUIE}, (f)FiveAPlus \cite{ref66}, (g)CWR \cite{CWR},  (h)Spectroformer \cite{spectroformer} ,(i)NU2Net\cite{NU2Net}, (j)X-caunet\cite{New1}, (k)WaterMamba\cite{New8}, and (l)our method.}
	\label{U45}
\end{figure*}

		The proposed method demonstrates consistent superiority in addressing various types of coupled degradation across five representative underwater datasets (Fig.~\ref{UIEB} , Fig.\ref{EUVP} -- \ref{U45} ). For the UIEB~\cite{UIEB} dataset (Fig.~\ref{UIEB}), it exhibited excellent performance in the coupled degradation scenario of color casting and backscattering, achieving superior local contrast and detail preservation when previous methods failed. For the EUVP dataset (Fig.~\ref{EUVP}), our approach performed robustly in the coupled degradation scenario of color shift and scattering effects, delivering enhanced perceptual quality even in visually challenging regions. For the LSUI dataset (Fig. ~\ref{LSUI}), our framework showed a strong generalization in the coupled degradation scenario of color deviation and backscattering, attaining a balanced enhancement in color fidelity, contrast, and sharpness. In the C60 dataset (Fig. ~\ref{C60}), our model effectively handled the coupled degradation scenario of color distortion and lowered illumination, providing accurate brightness and color correction while preserving fine structural details. For the U45 dataset (Fig. ~\ref{U45}), the proposed method achieved a comprehensive enhancement in the coupled degradation scenario of color imbalance and texture degradation, maintaining clear advantages in terms of both local contrast and color fidelity under complex underwater conditions.		
		
\begin{figure*}[t]
	\centering
	\includegraphics[width=\linewidth]{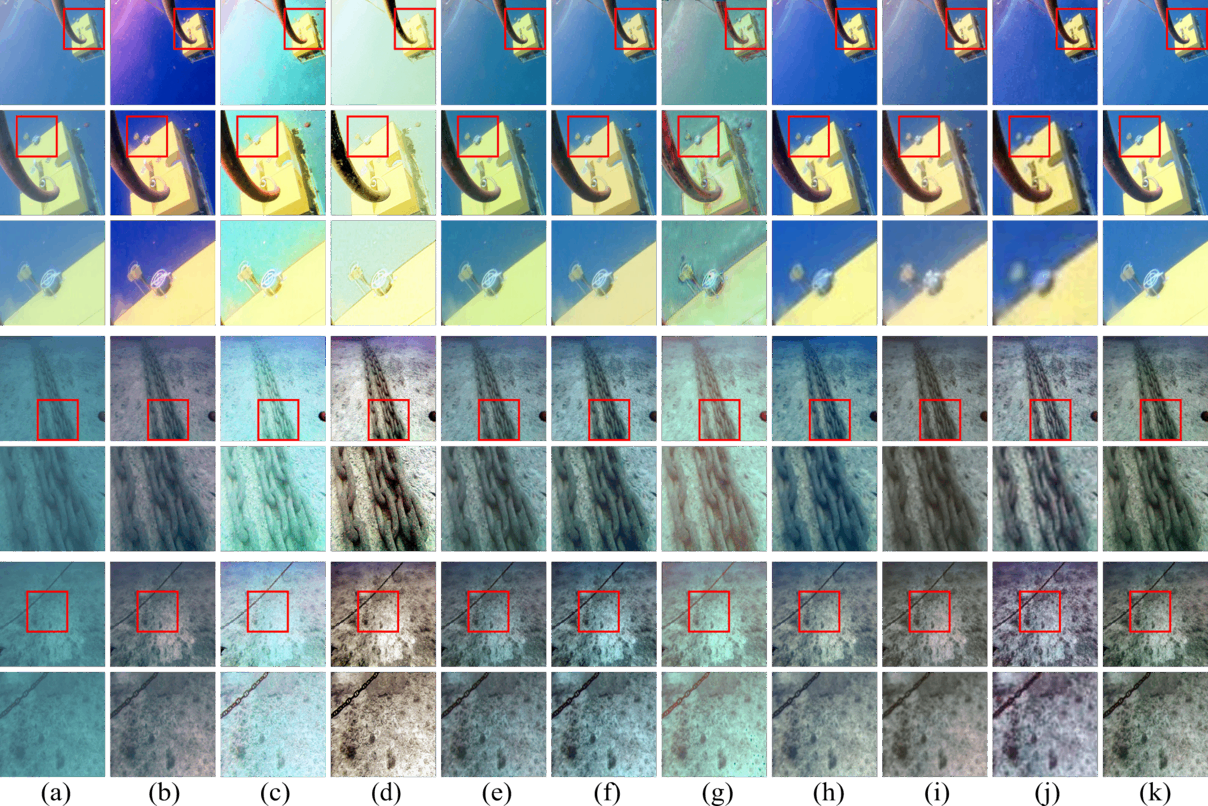}
	\caption{Visual comparison of different methods on the ROV dataset. The visual comparison of different methods from left to right includes (a)Input, (b)IBLA, (c)GDCP\cite{ref40}, (d)HS2CM2A \cite{HS2CM2A}, (e)PUIE-MC \cite{PUIE}, (f)FiveAPlus \cite{ref66}, (g)CWR \cite{CWR},  (h)Spectroformer \cite{spectroformer} ,(i)X-caunet\cite{New1}, (j)WaterMamba\cite{New8}, and (k)our method.}
	\label{ROV}
\end{figure*}

\begin{figure*}[t]
	\centering
	\includegraphics[width=\linewidth]{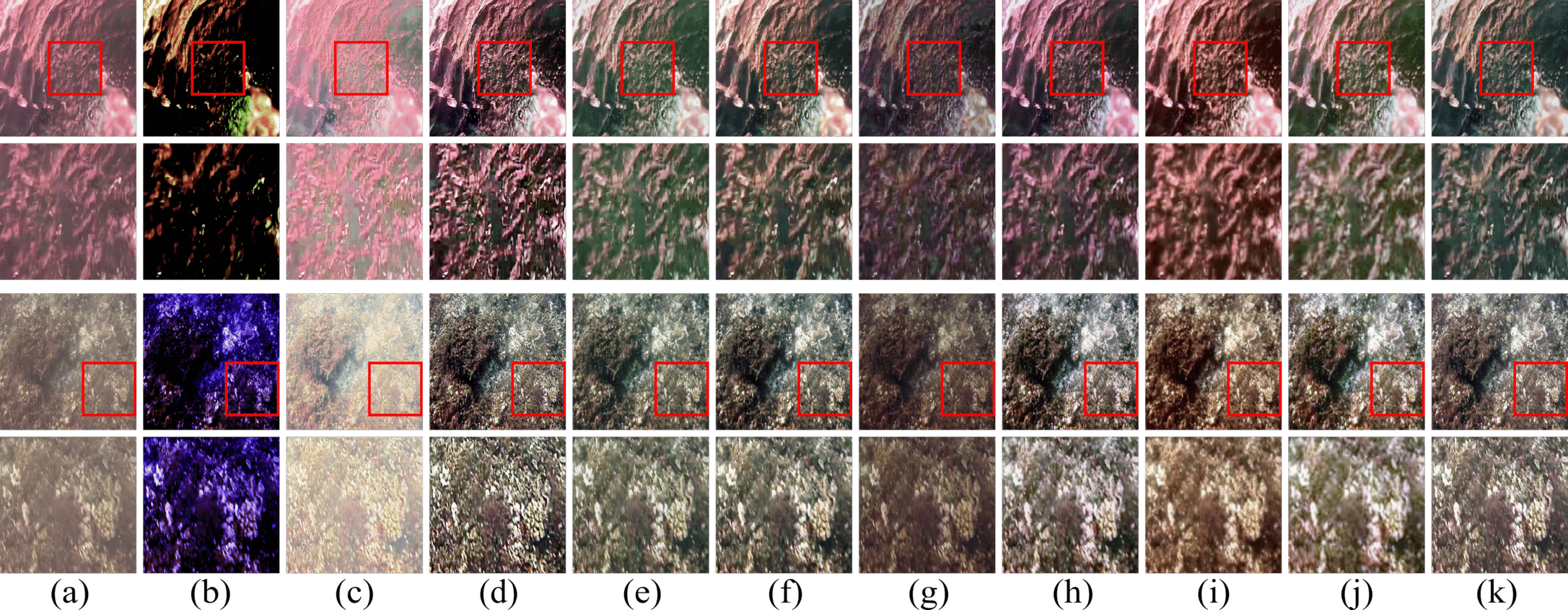}
	\caption{Visual comparison of different methods on the QDH dataset. The visual comparison of different methods from left to right includes (a)Input, (b)IBLA\cite{IBLA}, (c)GDCP\cite{ref40}, (d)HS2CM2A \cite{HS2CM2A}, (e)PUIE-MC \cite{PUIE}, (f)FiveAPlus \cite{ref66}, (g)CWR \cite{CWR},  (h)Spectroformer \cite{spectroformer} ,(i)X-caunet \cite{New1}, (j)WaterMamba \cite{New8}, and (k)our method.}
	\label{QDH-1}
\end{figure*}

\begin{figure*}[t]
	\centering
	\includegraphics[width=\linewidth]{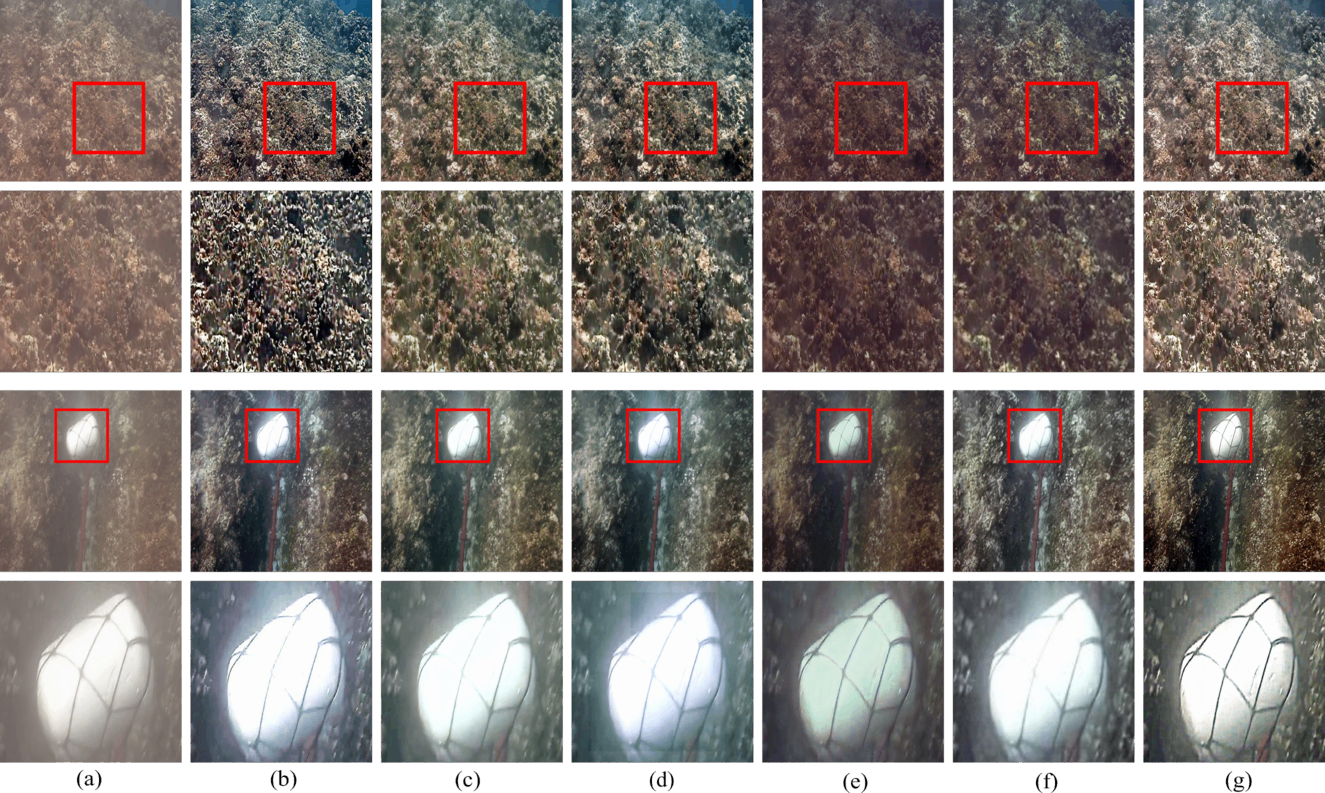}
	\caption{More visual comparisons of different methods on the QDH dataset. The visual comparison of different methods from left to right includes (a)Input, (b)HS2CM2A \cite{HS2CM2A},  (c)PUIE-MC \cite{PUIE}, (d)FiveAPlus \cite{ref66}, (e)CWR \cite{CWR},  (f)Spectroformer \cite{spectroformer} , and (g)our method.}
	\label{QDH-2}
\end{figure*}

		Moreover, our method delivered a robust enhancement across both marine and freshwater environments by effectively addressing coupled degradation (Fig.~\ref{ROV}--\ref{QDH-2}). For the 4K ROV dataset (Fig. ~\ref{ROV}), it performed well in scenarios involving coupled degradation of turbidity, low illumination, and color deviation, achieving superior detail preservation, contrast, and sharpness. For the QDH dataset (Fig.~\ref{QDH-1}, Fig.~\ref{QDH-2}), it exceled in handling the red-colored cast and fine-detail loss, demonstrating better adaptability than prior methods. Further comparisons show that our approach avoids overexposure and color loss, while achieving a balanced enhancement across color, contrast, and texture.
		
		Overall, these results validate the effectiveness and generalizability of our method in robustly handling multiple coupled degradations, consistently achieving a balanced enhancement of perceptual quality and physical image attributes across diverse underwater scenarios.

	\subsubsection{Quantitative comparison}

		Tables~\ref{UIEB_Data}  summarize the quantitative comparisons on UIEB dataset \cite{UIEB}. The quantitative comparison results of EUVP~\cite{EUVP}, LSUI~\cite{New5}, U45~\cite{U45}, C60, UCCS~\cite{UCCS}, QDH, and ROV can be found in Appendix B. The evaluation metrics included PSNR\cite{PSNR}, SSIM\cite{SSIM}, UIQM\cite{UIQM}, UCIQE\cite{UCIQE}, CCF, and PCQI, along with model complexity (Params, FLOPs), and runtime.
		
		For UIEB\cite{UIEB}, LSUI\cite{New5}, and EUVP\cite{EUVP} (Tables~\ref{UIEB_Data}, \ref{LSUI_Data}, \ref{EUVP_Data}), our method was consistently ranked first or near the top across all major metrics, demonstrating superior perceptual and structural quality under complex degradation. These results confirm its strong generalizability across diverse environments. For U45\cite{U45}, C60, and UCCS\cite{UCCS} (Table ~\ref{U45C60UCCS}), our approach achieved the best UIQM\cite{UIQM} scores while maintaining favorable efficiency in terms of parameter count and FLOPs, indicating an effective balance between performance and computational cost. For the high-resolution QDH and ROV datasets (Table~\ref{qdh_rov_Data}), our method performed competitively across all metrics, achieving the highest UCIQE and CCF scores on ROV and robust results on QDH. Despite not being the fastest, it was ranked among the top three in terms of speed, demonstrating its strong practicability in real-world settings.

		Notably, non-deep learning methods (e.g., GDCP~\cite{ref40}, UDCP~\cite{UDCP}, IBLA~\cite{IBLA}, and HS2CM2A~\cite{HS2CM2A}) often yield high CCF scores despite their inferior visual quality, exposing a key limitation of CCF in reflecting perceptual naturalness and color fidelity. As a statistical measure of colorfulness, contrast, and clarity, the CCF lacks sensitivity to over-enhancement and unnatural color distributions. Consequently, it may overestimate the quality of the visually distorted results. Thus, the CCF should be treated as an auxiliary indicator and used in conjunction with subjective assessment and other no-reference metrics to reliably evaluate the enhancement performance.
		
		Considering both qualitative and quantitative comparisons, the proposed method consistently surpassed existing state-of-the-art approaches in terms of perceptual quality, color fidelity, and detail preservation. These results indicate that our method achieved a favorable tradeoff between enhancement effectiveness and practical efficiency, demonstrating its robustness and generalization in diverse underwater image enhancement scenarios involving coupled degradations.

	\begin{figure}[t]
		\centering
		\includegraphics[angle=0,width=3in]{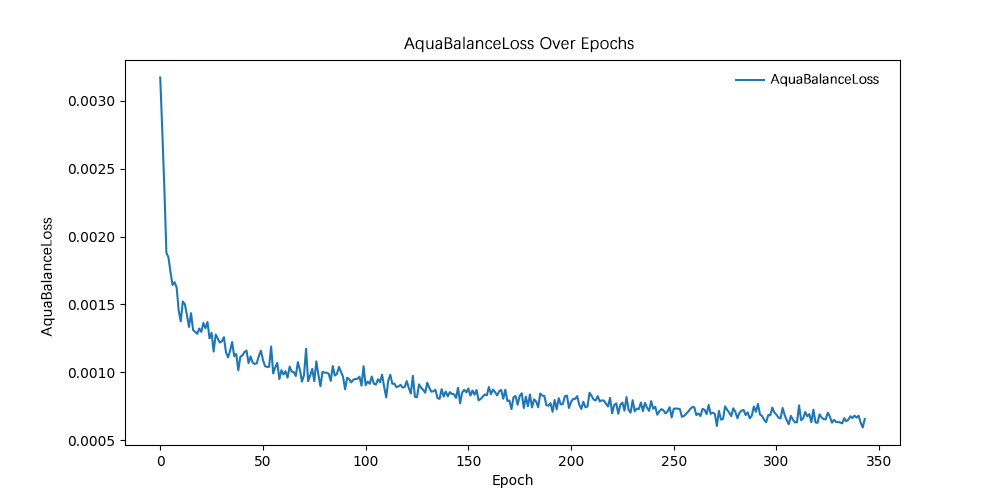}
		\caption{Training loss curve of AquaBalanceLoss on the UIEB dataset \cite{UIEB}, with the horizontal axis representing the epoch and the vertical axis representing the AquaBalanceLoss.}
		\label{fig_5}
	\end{figure}

	\subsection{AquaBalanceLoss}
\begin{figure}[t]
		\centering
		\includegraphics[angle=0,width=3in]{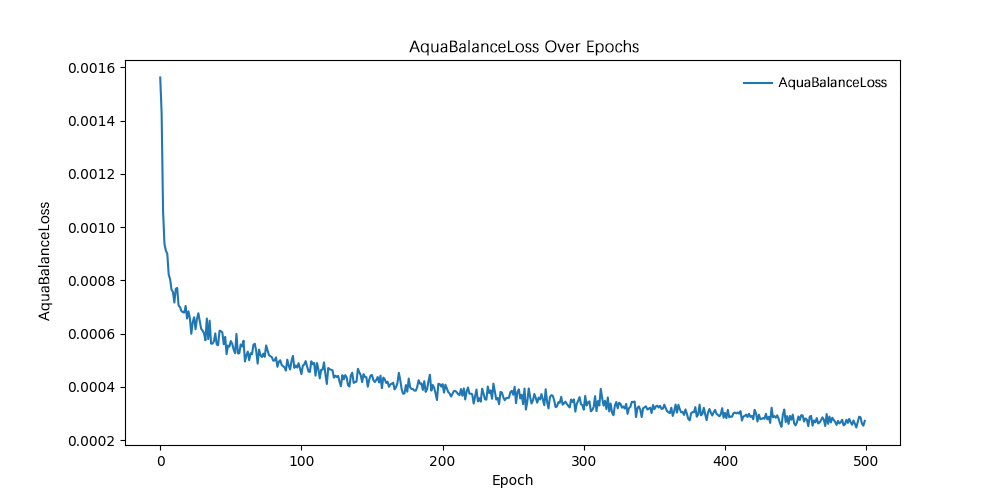}
		\caption{Training loss curve of AquaBalanceLoss on the EUVP dataset \cite{EUVP}, with the horizontal axis representing the epoch and the vertical axis representing the AquaBalanceLoss.}
		\label{fig_6}
	\end{figure}

	The loss stability and minimal performance variation indicate a high degree of model stability, which is crucial for underwater image enhancement tasks. Here, we analyze the convergence characteristics of the proposed AquaBalanceLoss. The loss function value gradually decreases and stabilizes as the training progresses (Fig.~\ref{fig_5} and ~\ref{fig_6}).
	
	To further evaluate the relationship between the convergence process and the model performance, we evaluated the model performance at various training stages (100, 200, 300, 400, and 500 epochs) on the UIEB \cite{UIEB} dataset. A comparison of the loss curves and performance metrics revealed the dynamic process of the model optimization.

\begin{figure}[h]
		\centering
		\includegraphics[angle=0,width=3in]{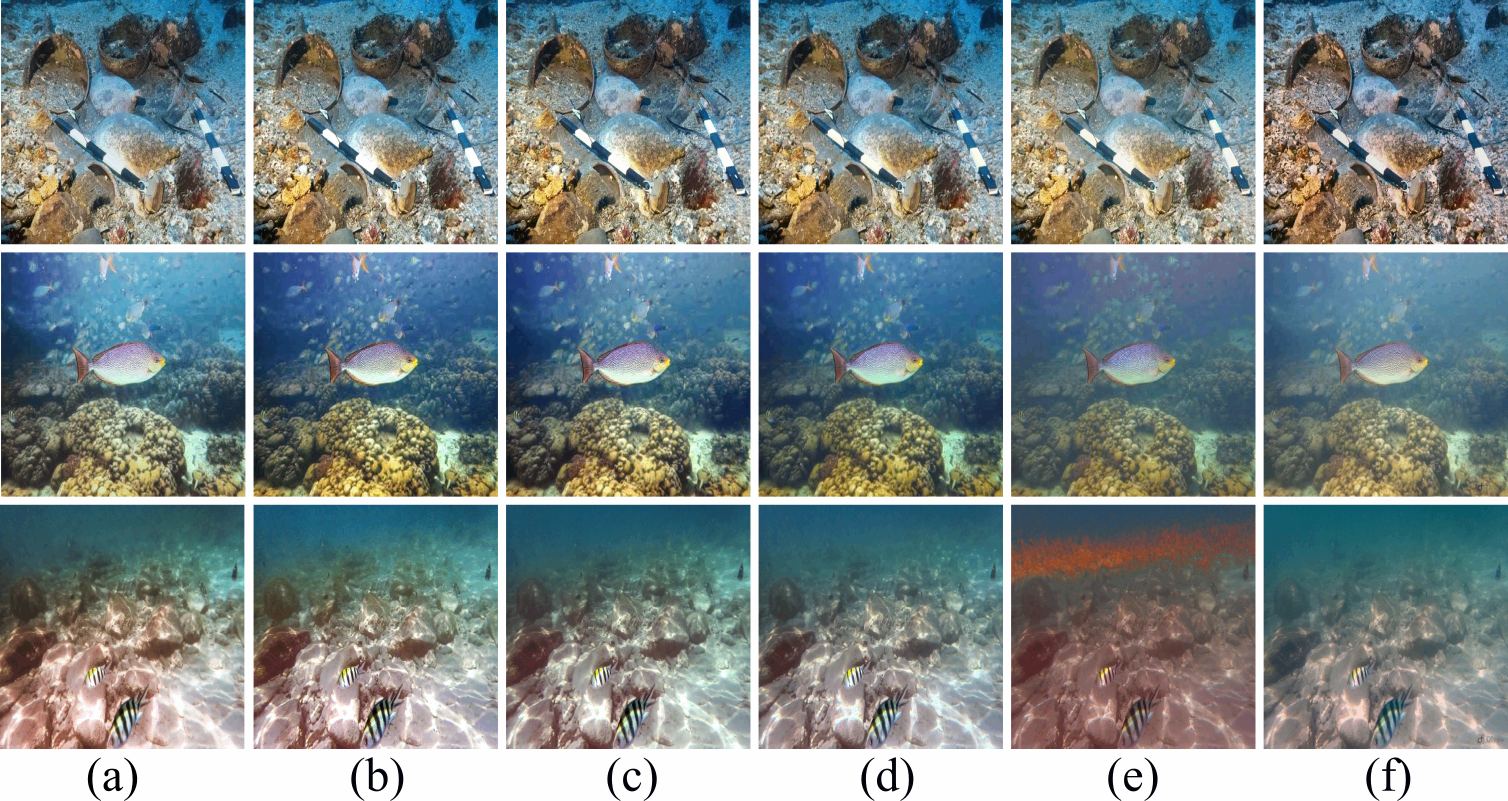}
		\caption{Visual comparison of models at different training epochs on the UIEB dataset \cite{UIEB}; from left to right, are the inference results for training epochs of 100(a), 200(b), 300(c), 400(d), 500(e), and the ground truth image(f).}
		\label{fig_7}
	\end{figure}

	The qualitative results demonstrate a gradual improvement in image quality during the initial 300 epochs (Fig.~\ref{fig_7}). We observed a continuous increase in image brightness throughout the training, although this resulted in a gradual loss of image detail. The model achieved optimal visual performance at 300 epochs. Beyond this, the visual quality gradually deteriorated, with inference errors appearing at 500 epochs (third row of Fig~\ref{fig_7}). This suggests that higher training epochs do not necessarily improve the performance, with overfitting becoming apparent after 500 epochs.
	
	The quantitative results in Table \ref{epoch12345} also suggest that higher training epochs are not always beneficial, as all metrics, except UIQM \cite{UIQM}, initially increase and then decrease. The UIQM \cite{UIQM} metric shows a continuous increase because AquaBalanceLoss dominates in the later training stages, leading to network bias toward this loss function and a sustained UIQM increase.
	
	To prevent overfitting, and based on quantitative and qualitative comparisons, we set the number of training epochs to 300. This decision was based on both the loss curve trend and generalizability of the model on independent test sets.
	
	We evaluated the performance stability of AquaBalanceLoss across different datasets. To understand this variability, we tracked the loss function changes during the same training process on the UIEB \cite{UIEB} and EUVP \cite{EUVP} datasets, as shown in Fig.~\ref{fig_5} and ~\ref{fig_6}. The results show that despite visual and statistical differences between datasets, AquaBalanceLoss exhibited low variability. The low fluctuations and consistent losses further demonstrate the robustness of the loss function.

	\begin{table}[h]
		\centering
		\caption{Quantitative data comparison of the impact of different training epochs on model performance on the UIEB dataset \cite{UIEB}. The highest value is represented in bold, and the second-highest value is represented with an underline.}
		\label{epoch12345}
		\begin{tabular}{cccccc}
			\toprule
			& 100 & 200 & 300 & 400 & 500 \\
			\midrule
			PSNR & 21.705 & 22.330 & {\bfseries 23.007} & \uline{22.746} & 22.416 \\
			SSIM & 0.857 & 0.877 & {\bfseries 0.915} & \uline{0.882} & 0.880 \\
			UIQM & 3.036 & 3.009 & 3.045 & \uline{3.055} & {\bfseries 3.067} \\
			UCIQE & 0.603 & \uline{0.615} & {\bfseries 0.626} & \uline{0.615} & 0.606 \\
			PCQI & 0.947 & 0.969 & \uline{0.979} & 0.948 & \textbf{0.991} \\
			CCF & 22.809 & 25.746 & \textbf{29.908} & \uline{25.989} & 25.540 \\
			\bottomrule
		\end{tabular}

	\end{table}

	\section{Ablation Study}
	To demonstrate the effectiveness of the proposed components and loss functions and to determine the optimal weight for AquaBalanceLoss, we conducted subsequent ablation studies on the UIEB \cite{UIEB}, U45 \cite{U45}, C60 and ROV datasets.

	\subsection{Effectiveness of Modules}
	
	In this section, we evaluate the effectiveness of each module in addressing coupled degradation scenarios as well as the collaborative performance among different modules.

	\begin{figure}[t]
		\centering
		\includegraphics[angle=0,width=3in]{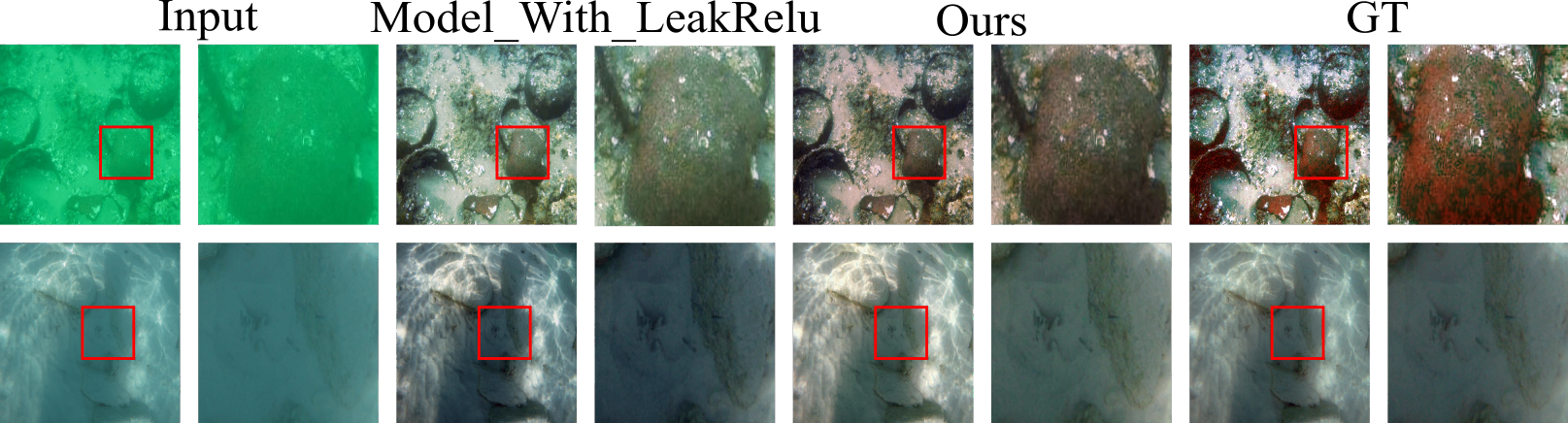}
		\caption{Ablation experiment of the JFE module considering nonlinear factors on the UIEB dataset: (a) Raw input image, (b) network with replaced layers and modules, and (c) complete model with all proposed components. }
		\label{ELU-LeakRelu-UIEB}
	\end{figure}

	\begin{figure}[t]
		\centering
		\includegraphics[angle=0,width=3in]{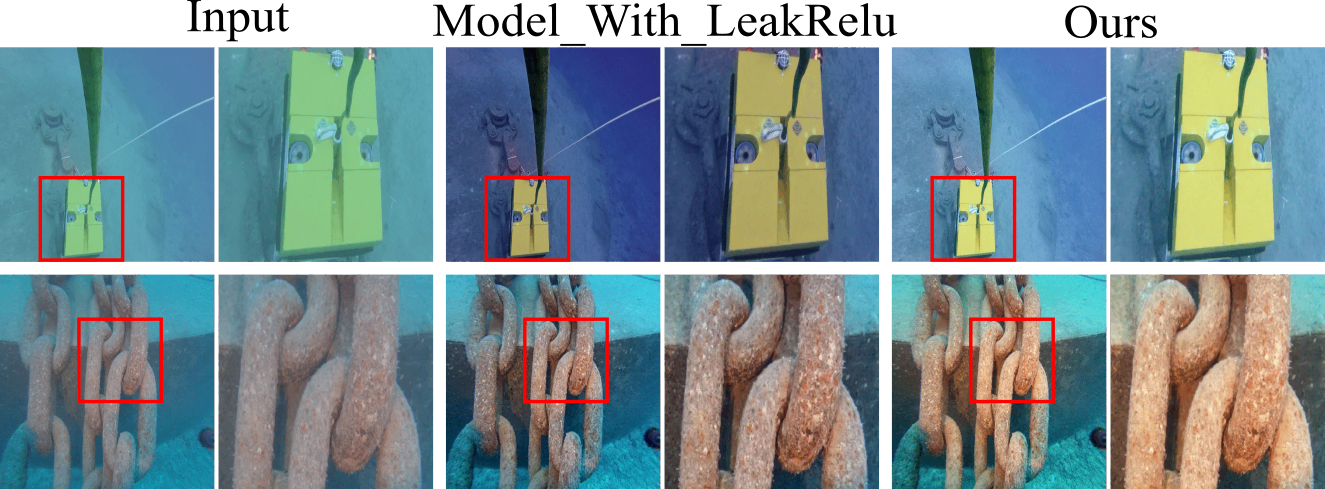}
		\caption{Ablation experiment of the JFE module considering nonlinear factors on the ROV dataset: (a) raw input image, (b) network with replaced layers and modules, and (c) complete model with all proposed components.}
		\label{ELU-LeakRelu-ROV}
	\end{figure}

	\subsubsection{Effectiveness of the JFE Module}
	To rigorously evaluate the effectiveness of the proposed JFE module in addressing various forms of coupled degradation and to quantitatively assess its contribution to the enhancement capability of the network, we conducted comprehensive ablation experiments from two perspectives: nonlinearity and module substitutability.

\begin{figure}[t]
		\centering
		\includegraphics[angle=0,width=3in]{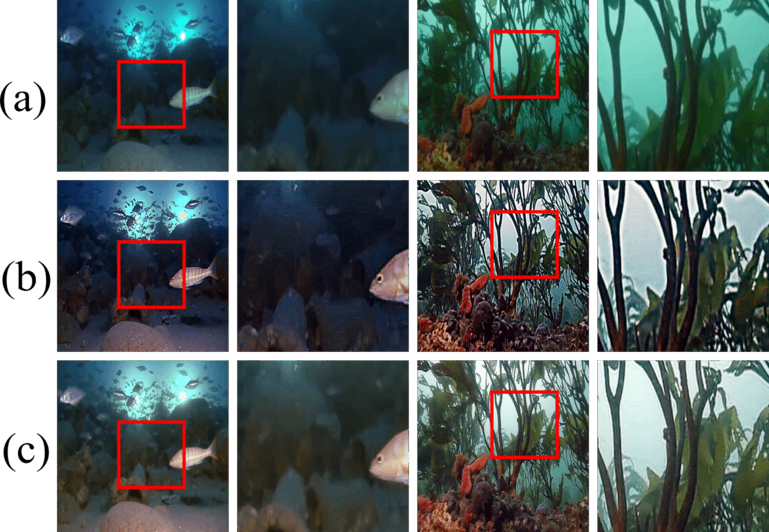}
		\caption{Visual comparison of the ablation experiment for the JFE module on the C60 dataset. From left to right: (a)original image, (b)inference result of the model after replacing the JFE module, and (c)inference result of the complete model.}
		\label{JFEAblationStudy}
	\end{figure}

		For the module substitutability analysis, we designed two network variants: one incorporating the complete JFE module, and the other in which the ESE-ResBlock was replaced by a standard SE-ResNet block; the FE module was substituted by conventional convolution and pooling layers. The qualitative results are shown in Fig.~\ref{JFEAblationStudy}. Local zoom-ins are provided for detailed comparison. Notably, in the second row, variant (b) failed to enhance the brightness after color correction, resulting in suboptimal performance under a coupled color cast and low illumination. The final row (b) demonstrates insufficient detail enhancement and brightness improvement, with the overall color rendering clearly inferior to that of the complete model.

		For nonlinearity analysis, we compared the performances of two additional variants: one incorporating the complete JFE module and the other in which the ELU activation function was replaced by LeakyReLU. The ELU function provided a substantially greater nonlinear representational capacity than LeakyReLU. This setup enabled a focused evaluation of the influence of nonlinearity on the overall expressive power of the network in the context of underwater image enhancement (UIE). The qualitative results are shown in Fig.\ref{ELU-LeakRelu-UIEB} and\ref{ELU-LeakRelu-ROV}. As shown in the first rows of Fig.~\ref{ELU-LeakRelu-UIEB}, replacing the ELU with LeakyReLU leads to diminished color correction completeness, reduced backscatter suppression, and lower local contrast. Similarly, in the second row, the LeakyReLU variant exhibits inferior performance in terms of both brightness and local color contrast compared with the original model. As shown in the first row of Fig.~\ref{ELU-LeakRelu-ROV}, the complete model demonstrated a clear superiority over the LeakyReLU-based variant in terms of overall brightness and color expressiveness. Similarly, the second row indicates that the complete model outperformed the variant in terms of color correction capability. These findings demonstrate that the impact of nonlinearity extends beyond simple color, contrast, or brightness adjustments, which are crucial for comprehensively handling strongly coupled and nonlinear degradations. This observation substantiates our initial hypothesis that nonlinearity plays a pivotal role in the unified modeling and representation of complex degradation effects.
	
	In summary, the comparative results of these ablation studies clearly reinforce the effectiveness of the proposed JFE module. Our design strengthens the network's ability to focus on a variety of coupled degradation features and demonstrates that the explicit incorporation of nonlinearity substantially enhances the network's expressive power. Consequently, these improvements translated into significant advancements in the quality of underwater images, particularly in challenging and highly degraded scenarios.

	\subsubsection{Effectiveness of the FPP Module}
\begin{figure}[t]
			\centering
			\includegraphics[angle=0,width=3in]{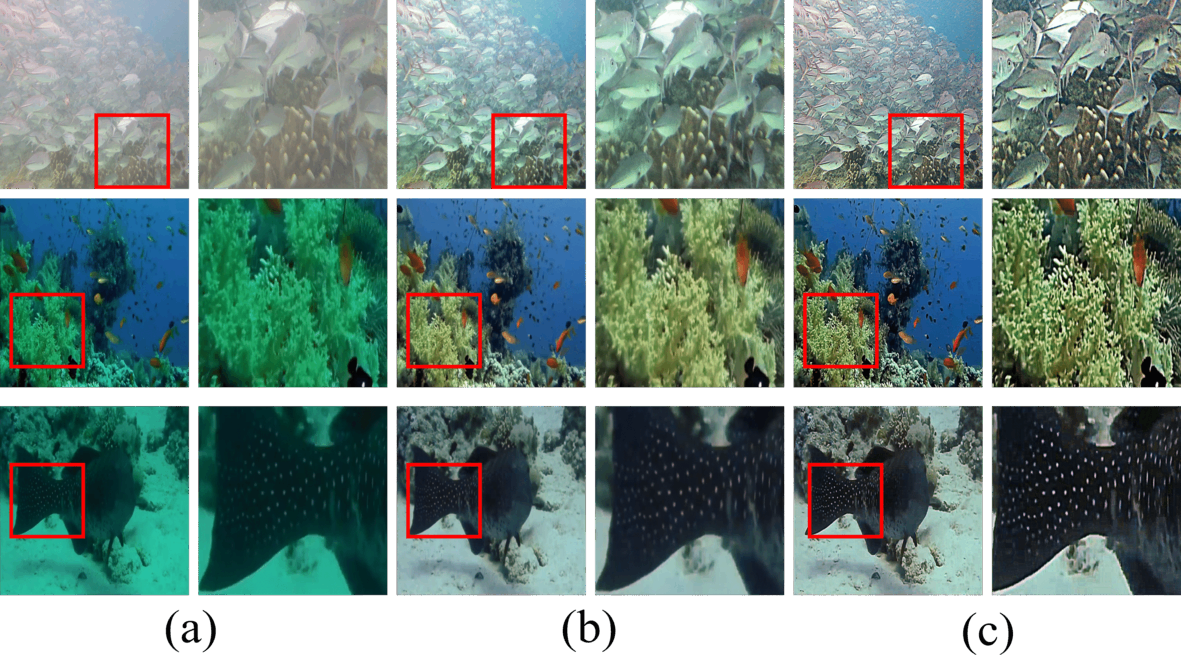}
			\caption{Visual comparison of the ablation experiment for the FPP module on the C60 dataset. From left to right: (a) original image, (b) inference result of the model after deleting the FPP module, and (c) inference result of the complete model.}
			\label{FPPAblationStudy}
		\end{figure}

	We explored the impact of the FPP module on network performance. By adding or removing the FPP module and comparing its performance, we further validated its effectiveness. The visual comparisons in Fig.~\ref{FPPAblationStudy} show significant improvements in image details and clarity after processing with the FPP module, confirming its intended design.

	\subsubsection{Effectiveness of Collaborative Performance among Different Modules}

\begin{figure}[t]
		\centering
		\includegraphics[angle=0,width=3in]{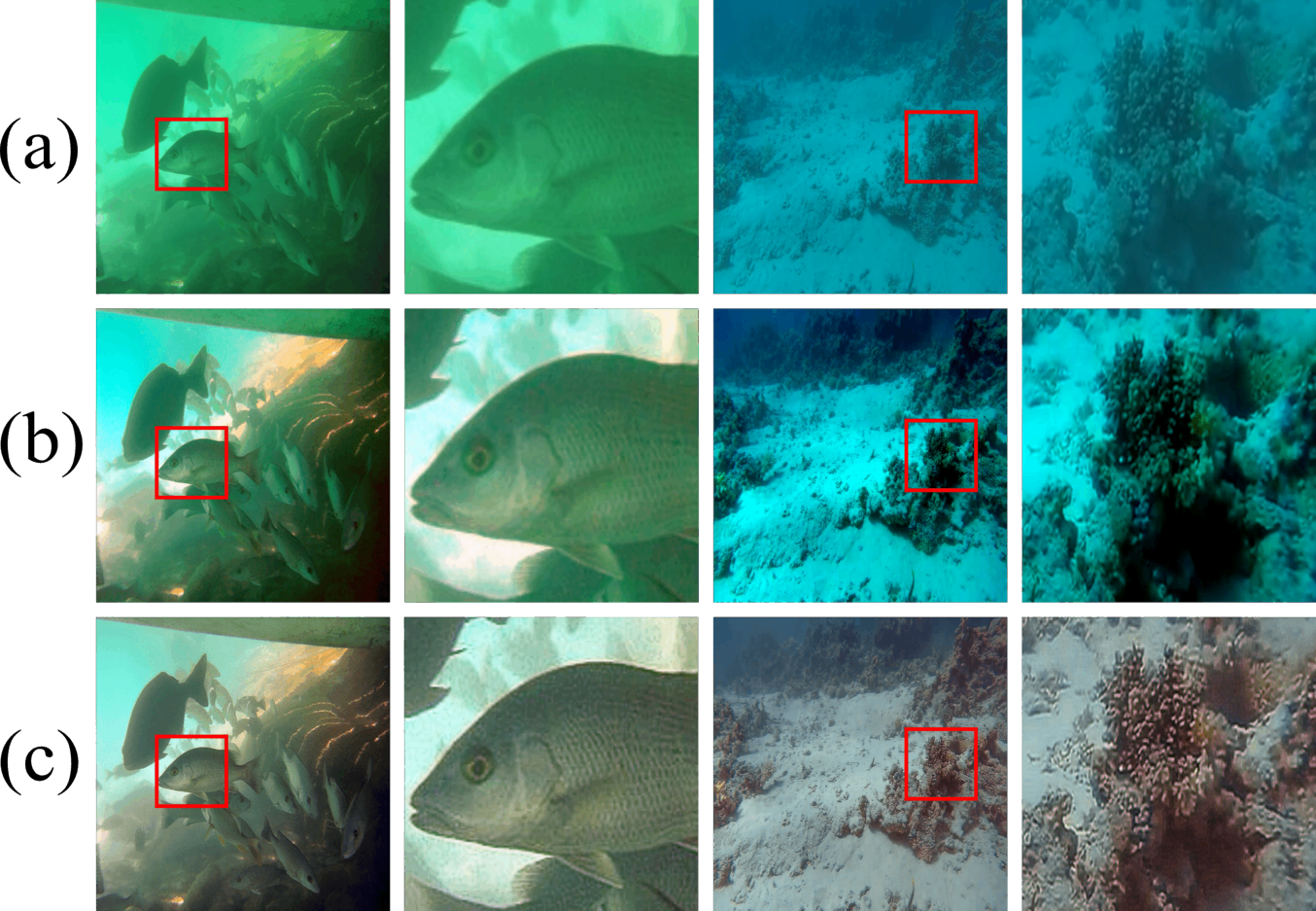}
		\caption{Visual comparison of the ablation experiment for the PB and FPP modules on the C60 dataset: (a) original image, (b) image only using JFE module, and  (c)inference result of the complete model.}
		\label{woPBFPP}
	\end{figure}

		To further assess the contributions of the PB and FPP modules, ablation experiments were performed in which only the JFE module was retained. The qualitative results are shown in Fig.~\ref{woPBFPP}. The results clearly indicate that using the JFE module alone is inadequate for the unified handling of coupled degradations and fails to resolve any single degradation, such as color cast, low illumination, or backscattering when they occur together. Compared to the complete model, the JFE-only variant is unable to perform a thorough and integrated processing of coupled degradations. This limitation is mainly caused by the absence of a PB module, which plays a critical role in adjusting the global consistency and leveraging latent information under coupled degradation conditions. Without PB, the enhancement remains incomplete, even with JFE's latent feature extraction. The introduction of PB enables a more comprehensive use of the information extracted by JFE, significantly improving global consistency and demonstrating the effective synergy between JFE and PB for the unified and robust handling of coupled degradations.

		Moreover, the ablation result using only the JFE module is noticeably inferior to the complete model in terms of detail preservation (Fig. \ref{woPBFPP}). This was caused by the lack of the FPP module, which is responsible for further refining texture information. Notably, no visual artifacts were introduced, further confirming the complementary roles of the FPP and PB modules in artifact removal and detail enhancement. When both PB and FPP modules were employed, the model consistently produced artifact-free images with enhanced edge sharpness and improved local details.

		The quantitative results in Table~\ref{woModuleData} demonstrate that the complete model achieved the best overall performance across all evaluation metrics (UIQM, UCIQE, and CCF), attaining the highest values in each category under coupled degradation scenarios. Notably, removing any individual module resulted in a significant decline in quantitative performance when addressing coupled degradations. These findings underscore the necessity of each module---joint feature extraction, global consistency adjustment, and fine-grained detail refinement---for comprehensively and effectively handling the complex interactions inherent in coupled underwater degradations, thereby enabling superior enhancement outcomes.

		\begin{table}[t]
		\centering
			\caption{Quantitative data comparison of ablation experiments for various modules on the C60 dataset. The highest value is represented in bold, and the second-highest value is represented with an underline.}
			\label{woModuleData}
			\begin{tabular}{ccccccc}
				\toprule
				JFE-ELU & JFE-Mod & PB & FPP & UIQM & UCIQE & CCF \\
				\midrule
				& \checkmark & \checkmark & \checkmark & 2.298 & 0.558 & 16.405 \\
				\checkmark &           & \checkmark & \checkmark & \uline{2.465} & \uline{0.571} & \uline{25.559} \\
				\checkmark & \checkmark & \checkmark &           & 2.183 & 0.558 & 21.395 \\
				\checkmark & \checkmark &           &           & 2.003 & 0.550 & 15.099 \\
				\checkmark &           & \checkmark &           & 2.132 & 0.540 & 14.835 \\
				\checkmark & \checkmark & \checkmark & \checkmark & \textbf{2.627} & \textbf{0.589} & \textbf{28.924} \\
				\bottomrule
			\end{tabular}
		
	\end{table}

	\subsection{Effectiveness of Loss Function}
	To validate the effectiveness of the different loss functions in underwater image restoration tasks, we conducted a series of comparative experiments. This study evaluated the contribution of loss functions, such as the KL loss, reconstruction loss, and AquaBalanceLoss in enhancing the model performance.
	
	In the experimental setup, we used a consistent model architecture and training strategy to ensure a fair comparison between different loss functions. All models were trained on the UIEB dataset \cite{UIEB} and evaluated on the U45 \cite{U45} and challenging C60 datasets using multiple performance metrics such as PSNR \cite{PSNR}, SSIM \cite{SSIM}, UIQM \cite{UIQM}, and UCIQE \cite{UCIQE}.

\begin{figure}[t]
	\centering
	\includegraphics[angle=0,width=3in]{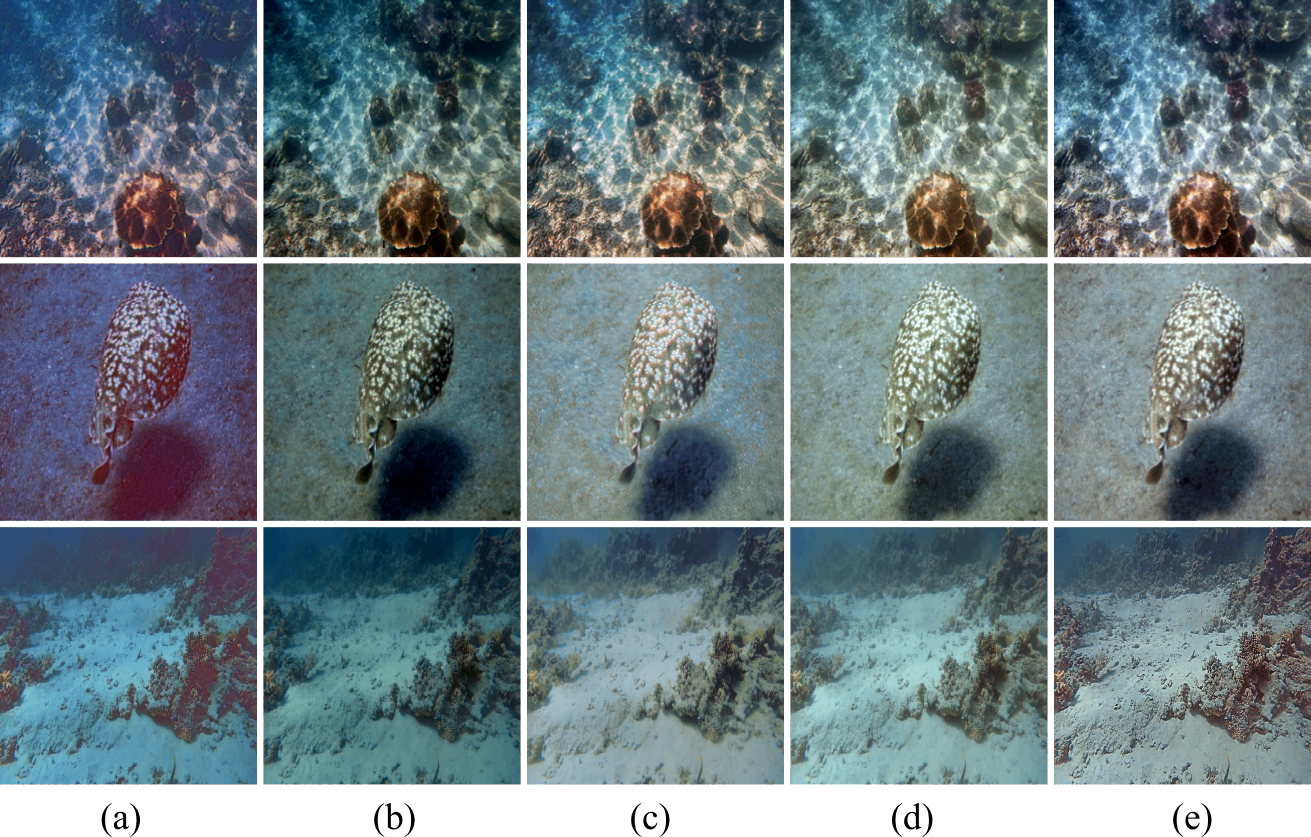}
	\caption{
		Visual comparison of the ablation experiment for various loss functions on the U45 \cite{U45} and C60 datasets. From left to right, the images correspond to the inference results of the model with (a)$ loss_{KL} $removed, (b)$loss_{Re}$ removed, (c)$loss_{vgg16} $ removed, (d)AquaBalanceLoss removed, and (e)complete model with all loss functions included.}
	\label{fig_10}
\end{figure}

	Fig.~\ref{fig_10} reveals that the absence of $loss_{KL}$ and $loss_{Re}$, compared to (e) where all loss functions are applied, significantly affected the results, leading to severe low-brightness issues. The absence of $loss_{vgg16}$ results in serious artifacts and a notable loss of image detail. Notably, removing only AquaBalanceLoss, as shown in (d), led to a noticeable gap in color, contrast, and detail performance compared to (e), highlighting the effectiveness of AquaBalanceLoss in enhancing color, sharpness, and contrast.

	In Table \ref{tab_4} and considering the visual aspects, we focused particularly on the performance without AquaBalanceLoss and $loss_{vgg16}$. Table \ref{tab_4} shows that the full model achieved optimal performance on the UIEB dataset \cite{UIEB} with the inclusion of AquaBalanceLoss, leading to significant metric improvements on the U45 \cite{U45} and C60 datasets. All analyses support our hypothesis that AquaBalanceLoss fosters a better balance between color, clarity, and contrast by guiding the network’s learning of multiple coupled degradation losses.

	\begin{table*}[t]
		\centering
		\caption{Quantitative data comparison of ablation experiments for various loss functions on the UIEB \cite{UIEB}, U45 \cite{U45}, and C60 datasets. The highest value is represented in bold, and the second-highest value is underlined.}
		\label{tab_4}
		\setlength{\tabcolsep}{0.15in}
		
	\begin{tabular}{l|ccccc|cc|cc}
		\toprule
		& \multicolumn{5}{c|}{UIEB} & \multicolumn{2}{c|}{U45} & \multicolumn{2}{c}{C60} \\
		& PSNR & SSIM & UIQM & UCIQE & PCQI & UIQM & UCIQE & UIQM & UCIQE \\
		\midrule
		w/o KL\_loss        & 17.69  & 0.828 & 2.98  & 0.594 & 0.7145   & 3.38  & 0.588 & \underline{2.668} & 0.576 \\
		w/o Re\_loss        & 17.497 & 0.795 & 2.316 & 0.607 & 0.6577   & 2.64  & \underline{0.596} & 2.193 & \underline{0.577} \\
		w/o vgg16\_loss     & 21.48  & 0.869 & 2.94  & 0.594 & \underline{0.8658} & \underline{3.332} & 0.591 & 2.615 & 0.566 \\
		w/o AquaBalanceLoss & \underline{22.66}  & \underline{0.898} & \underline{2.958} & 0.59  & 0.9404   & 3.37  & 0.58  & 2.464 & 0.57 \\
		Full                & \textbf{23.007} & \textbf{0.915} & \textbf{3.045} & \textbf{0.626} & \textbf{0.979} & \textbf{3.425} & \textbf{0.611} & \textbf{2.627} & \textbf{0.589} \\
		\bottomrule
	\end{tabular}

\end{table*}

\subsection{Effectiveness of Optimal Weight of AquaBalanceLoss}
To ensure that the proposed loss function achieved the best balance in adjusting the color, sharpness, and brightness of underwater images, we carefully adjusted and optimized the AquaBalanceLoss weights. In this study, we explored the impact of these parameters on the model performance by gradually increasing the weights from zero.

The experimental results are listed in Table \ref{tab_5}. As the AquaBalanceLoss weight increased incrementally, the model's performance first improved and then declined. PSNR \cite{PSNR} and SSIM \cite{SSIM} increased and then decreased, whereas UIQM \cite{UIQM} fluctuated but followed a similar pattern, possibly because of the overemphasis on a specific metric at the expense of the others. UCIQE \cite{UCIQE} exhibited a trend of first decreasing and then increasing.

Considering the comprehensive balance with visual performance and in line with the original intent behind designing this loss function, we ultimately set the weight of AquaBalanceLoss to 0.1. This adjustment aims to achieve a better equilibrium between brightness, contrast, and color, thereby providing a network with a more effective performance constraint. This careful calibration of AquaBalanceLoss weights guarantees that the network optimizes image enhancement in a manner that respects the multifaceted nature of underwater imagery, thereby ensuring improved visual quality across diverse underwater imaging conditions.

\begin{table}[t]
	\centering
	\caption{Quantitative data comparison of the impact of different weights of AquaBalanceLoss on network inference performance on the UIEB dataset \cite{UIEB}. The highest value is represented in bold, and the second-highest value is underlined.}
	\label{tab_5}
	\setlength{\tabcolsep}{0.15 in}
	
	\begin{tabular}{c|cccc}
	
		\toprule
		Setting & PSNR & SSIM & UIQM & UCIQE \\
		\midrule
		0.05 & \uline{22.4707} & \uline{0.8966} & 2.9355 & 0.6296 \\
		0.1 & {\bfseries 23.0066} & {\bfseries 0.9149} & \uline{3.0451} & 0.6258 \\
		0.2 & 21.8956 & 0.8789 & {\bfseries 3.0491} & 0.6237 \\
		0.3 & 21.9880 & 0.8804 & 2.8898 & 0.6304 \\
		0.4 & 21.9110 & 0.8564 & 2.9837 & \uline{0.6324} \\
		0.5 & 21.8511 & 0.8744 & 2.9414 & {\bfseries 0.6351} \\
		\bottomrule
	\end{tabular}
	
\end{table}

\subsection{Effectiveness and Generalization of AquaBalanceLoss Across Network Architectures}

\begin{figure}[t]
	\centering
	\includegraphics[angle=0,width=3.5in]{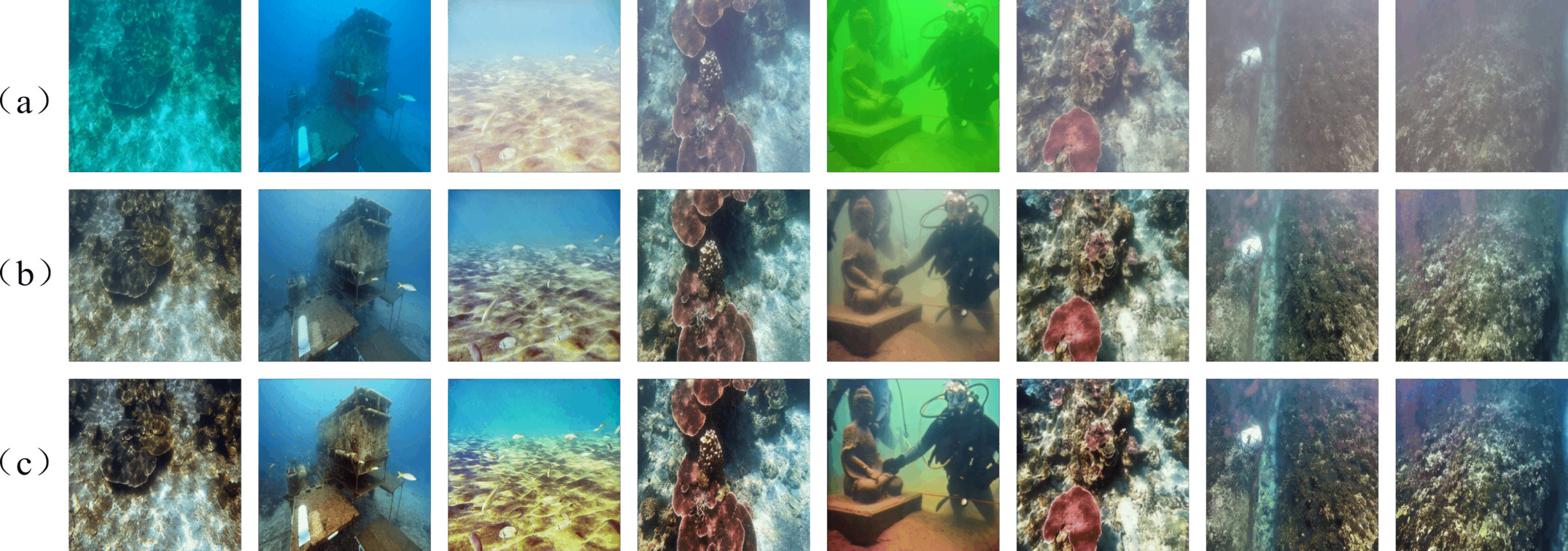}
	\caption{Qualitative visual comparison of PUIE-Net (CNN-based) before and after adding AquaBalanceLoss: (a) original input image; (b) inference result of the baseline model; (c) inference result of the model with AquaBalanceLoss added.}
	\label{PUIE_UIQM}
\end{figure}

\begin{figure}[t]
	\centering
	\includegraphics[angle=0,width=3.5in]{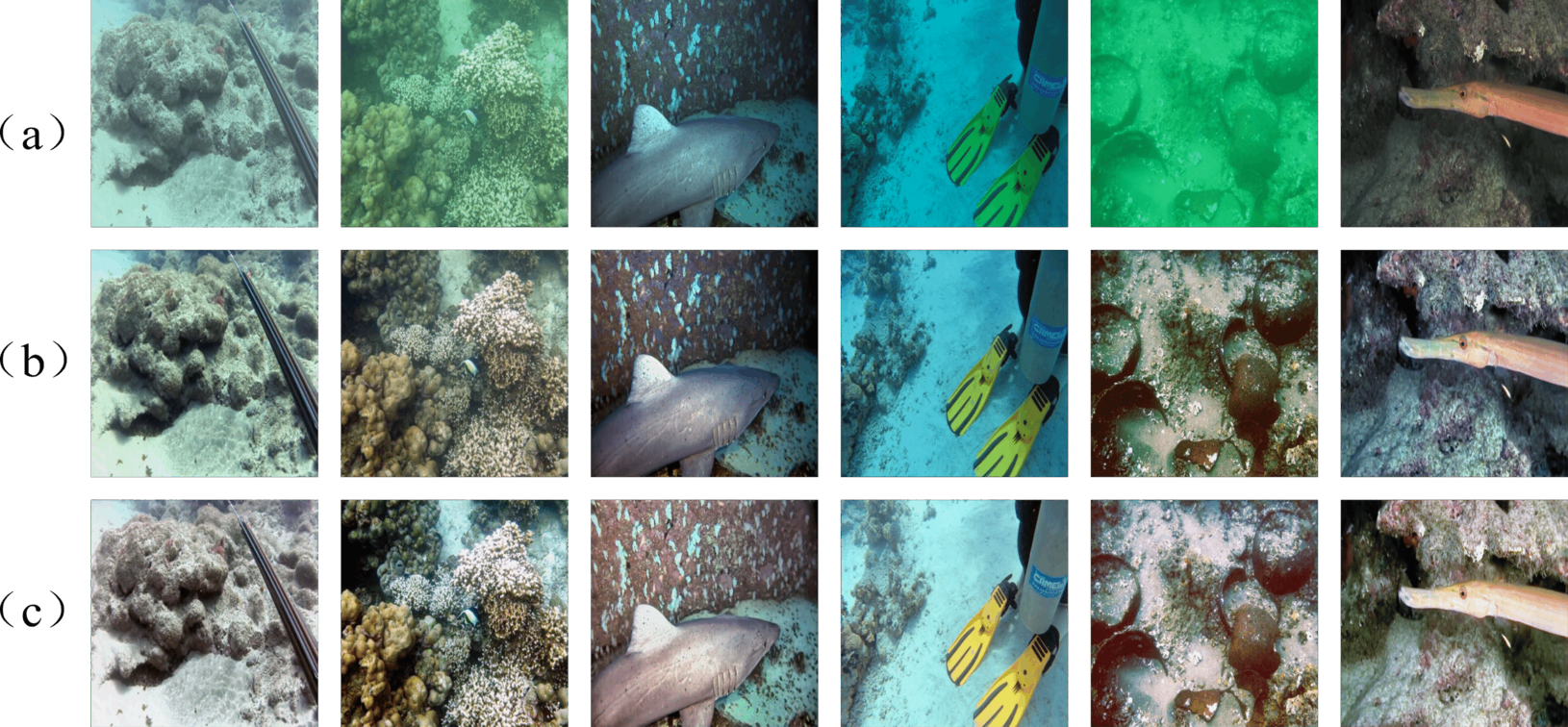}
	\caption{Qualitative visual comparison of Spectroformer (Transformer-based) before and after adding AquaBalanceLoss: (a) original input image; (b) inference result of the baseline model; (c) inference result of the model with AquaBalanceLoss added.}
	\label{Spec_UIQM}
\end{figure}

\begin{figure}[t]
	\centering
	\includegraphics[angle=0,width=3.5in]{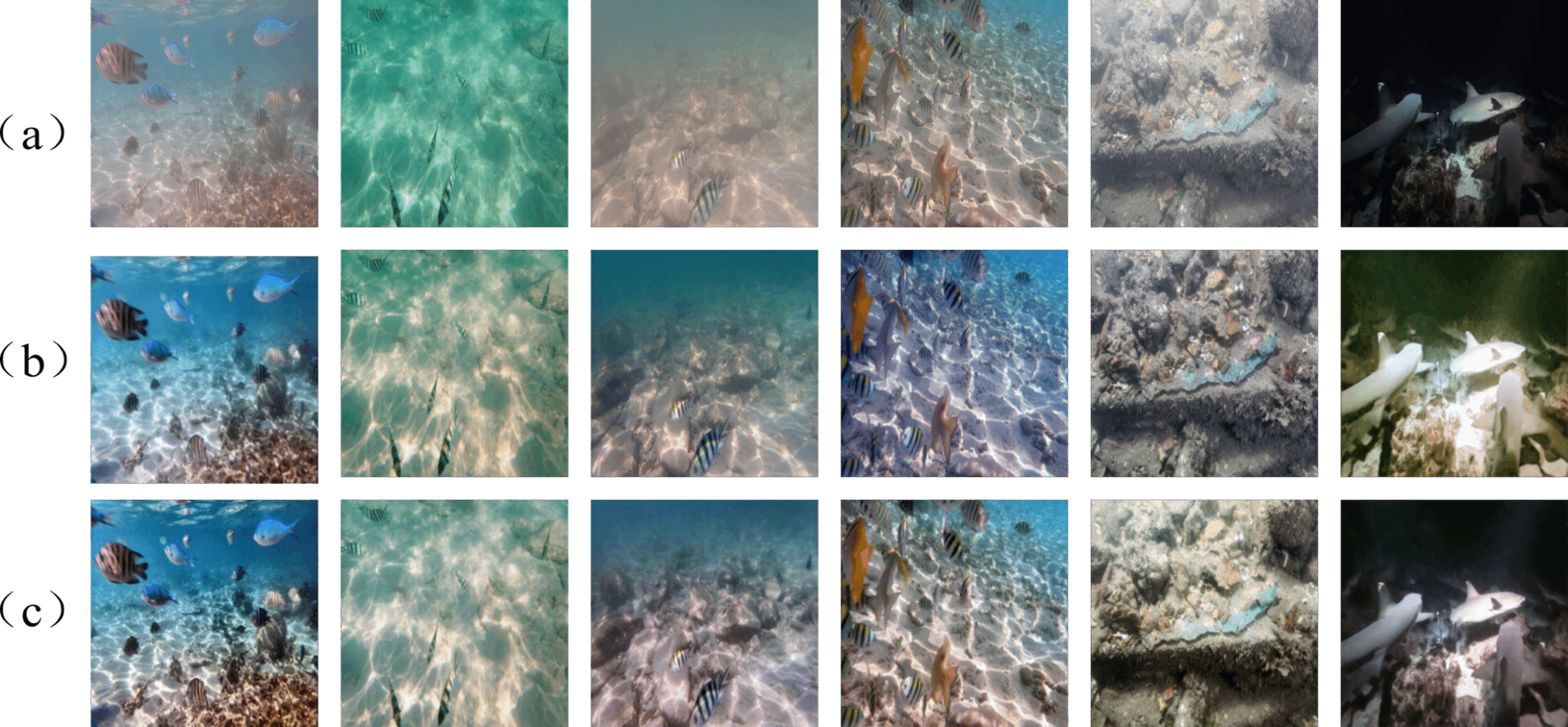}
	\caption{Qualitative visual comparison of WaterMamba (Mamba-based) before and after adding AquaBalanceLoss: (a) original input image; (b) inference result of the baseline model; (c) inference result of the model with AquaBalanceLoss added.}
	\label{WaterMamba_UIQM}
\end{figure}

	We further conducted additional experiments on CNN-based, Transformer-based, and Mamba-based models. By incorporating the proposed AquaBalanceLoss, we performed qualitative and quantitative evaluations and comparisons of their inference performance to further demonstrate the generalization ability of the proposed loss. The qualitative comparisons are shown in Figs. \ref{PUIE_UIQM}, \ref{Spec_UIQM}, and \ref{WaterMamba_UIQM}, and the quantitative comparisons are presented in Table \ref{CNN-Trans-Mamba-ABL}.
	
	Combining the above cross-model qualitative experimental results, AquaBalanceLoss shows consistent and significant enhancement benefits on CNN-, Transformer-, and Mamba-based architectures: it not only effectively suppresses the typical bluish-green color cast in underwater images and improves color restoration accuracy, but also further strengthens scattering removal ability, local contrast, and detail sharpness. After incorporating this loss, different models all exhibit more natural tones, clearer textures, and better brightness levels, indicating that the supervision signal provided by AquaBalanceLoss is largely structure-agnostic and has good generalization ability, and can guide the model to mine nonlinear degradation information and achieve stable, reliable, high-quality underwater image enhancement.

\subsection{Effectiveness of UIE}

	Finally, we evaluate the effectiveness of the proposed method on several downstream underwater vision tasks, including underwater depth estimation, semantic segmentation, edge detection, saliency detection, and object detection. Specifically, we adopt MiDaS \cite{MiDas} for underwater depth estimation, RCF \cite{RCF} for edge detection, SUIM-Net \cite{SUIM} for image segmentation, YOLOv5 for object detection, and U2Net \cite{U2Net} for saliency detection, all of which are highly representative models in the underwater vision field.
	
	The qualitative visual comparisons for all downstream tasks are provided in Figs. \ref{RUODOD}, \ref{URPC2020OD}, \ref{MiDaS-UIE}, \ref{RCF-UIE}, \ref{SUIMNet-UIE}, and \ref{U2Net-UIE} in Appendix~E, while quantitative object-detection results are summarized in Table~\ref{yolo}. Taken together, these experiments show that the proposed JDPNet-based UIE method yields stable and consistent gains in object detection, depth estimation, semantic segmentation, edge detection, and saliency detection. The improvements in color consistency, local contrast, and structural details of the enhanced images provide more reliable input representations for high-level vision models, thereby significantly boosting their perceptual performance. Overall, both the qualitative and quantitative results demonstrate the practical value and good generalization ability of our method as a UIE approach for downstream underwater perception and understanding tasks.

\begin{table}[t]
	\centering
	
		\caption{Quantitative comparison of detection performance of YOLOv5 on the URPC2020 and RUOD dataset before and after UIE using the JDPNet model.}
		\label{yolo}
		\begin{tabular}{lccccc}
			\toprule
			Dataset & Data & P & R & mAP50 & mAP50-95 \\
			\midrule
			\multirow{2}{*}{RUOD} 
			& original & 0.854 & 0.674 & 0.759 & 0.504 \\
			& UIE      & \textbf{0.871} & \textbf{0.789} & \textbf{0.850} & \textbf{0.586} \\
			\multirow{2}{*}{URPC2020} 
			& original & 0.731 & 0.485 & 0.570 & 0.299 \\
			& UIE      & \textbf{0.783} & \textbf{0.654} & \textbf{0.708} & \textbf{0.393} \\
			\bottomrule
		\end{tabular}
	
\end{table}

\section{Conclusion}

To address the challenges of effectively managing multiple coupled degradations in complex underwater environments, we propose a novel model called JDPNet for underwater image enhancement. We introduced several targeted architectural designs to achieve more effective management of multiple coupled degradation scenarios. These include the JFE module for extracting coupled degradation information, the PB module for adjusting content features and aligning training characteristics, and the FPP module for feature postprocessing to refine feature representations and correct edge artifacts. Furthermore, a new loss function, AquaBalanceLoss, specifically designed for underwater characteristics, was introduced to constrain JDPNet for effective balancing of color, clarity, and contrast. Our ablation experiments demonstrate the effectiveness of our proposed components, with our model achieving state-of-the-art enhancement in quality and visual appeal across various underwater datasets. These advancements significantly enhanced the visual quality of images in complex underwater scenes and established a robust foundation for the visual perception capabilities of underwater robotic operations. Future studies will integrate physically based-models of scattering and absorption into network design, expand training data diversity to cover a broader range of underwater degradations, develop degradation-aware loss functions, and further explore self-supervised learning and domain adaptation to enhance robustness and generalization under complex real-world conditions.

\appendices

\section{Discussion of Gradient angle on AquaBalanceLoss}

We established the core formulation of AquaBalanceLoss based on color, sharpness, and contrast, incorporating “color fidelity,” “detail preservation,” and “contrast enhancement” as independent objectives within a unified multi-objective loss framework. This design imposes perceptual constraints on the network, forcing it to jointly optimize color, contrast, and sharpness, rather than converging to trivial or attribute-biased solutions. During backpropagation, the network is driven toward perceptual equilibrium, which substantially enhances generalization and robustness, ensuring that, even under previously unseen or coupled degradations, the enhanced images remain balanced and free from overfitting to specific attributes.

Traditional losses often suffer from implicit redundancy. However, the three components of AquaBalanceLoss exhibit near-orthogonality in the gradient space, constraining the network output from complementary dimensions: the color term regulates global chromaticity, the sharpness term supervises hierarchical detail, and the contrast term governs the local luminance structure. This parallel, multi-dimensional optimization enables the network to approximate the Pareto front of the multi-objective problem, achieving solutions where no attribute’s improvement is achieved at the expense of another.

For the color loss ($L_{coi}$), gradients are predominantly derived from variations in the mean and variance of RG/YB channels, rendering the network highly sensitive to global color distribution and robust against color casts. This loss promotes both the elimination of chromatic aberration and the maximization of color-information entropy, preventing mode collapse and encouraging outputs that align with natural image statistics.

For the sharpness loss ($L_{si}$), we adopted a block-based local contrast metric using the logarithmic ratio of block-wise maximum and minimum pixel values. This formulation is mathematically sensitive to low-contrast details and amplifies gradients where subtle structures are present. It mitigates false sharpening and prevents edge attenuation. Summing across blocks ensures robustness to global luminance shifts and concentrates the gradient flow on local structures.

The contrast loss ($L_{cti}$) is designed as a local extremum-based logarithmic entropy metric, possessing properties of information entropy: the greater the diversity of local contrasts, the higher is the overall score. Adaptive scaling by power transformation prevents gradient explosion or vanishing, ensuring stable optimization and maximizing spatial entropy and information density in the reconstructed image.

Collectively, AquaBalanceLoss leverages robust statistical measures, gradient orthogonality, and global information flow to maximize multi-attribute information entropy while minimizing redundancy. This results in efficient gradient aggregation, improved training stability, and enhanced perceptual consistency and discriminability of the final output.

\section{Statistical information of QDH and ROV datasets}

\begin{figure*}[t]
	\centering
	\includegraphics[angle=0,width=7in]{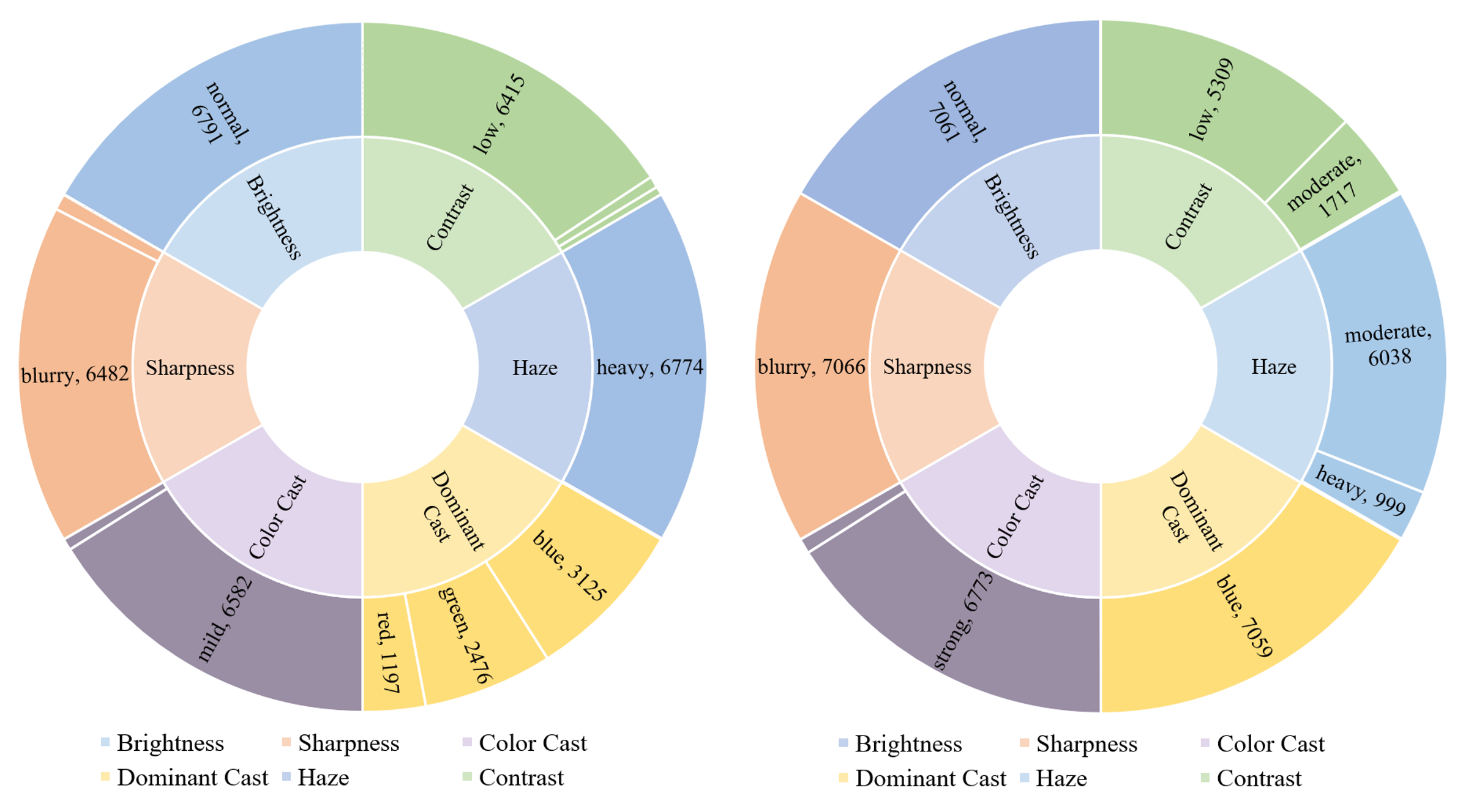}
	\caption{Sunburst visualizations of the QDH (left) and ROV (right) dataset statistics. For QDH, most images are affected by heavy haze, blur, and low contrast under normal illumination with mainly blue/green dominant color casts, indicating that QDH is dominated by complex, highly degraded underwater scenes. For ROV, most images also have normal illumination but suffer from blur, low or moderate contrast, moderate-to-heavy haze, and strong blue color cast, further demonstrating complex and severely degraded underwater conditions.}
\label{QDH+ROV}
\end{figure*}

\section{More comparisons of representative methods on public datasets and newly created datasets}

\begin{table*}[h]
	\centering
	\small
	
		\caption{Quantitative method comparisons across EUVP \cite{EUVP} datasets. The highest value is in bold, and the second-highest is underlined.}
		\label{EUVP_Data}
		\begin{tabularx}{\textwidth}{lYYYYYY}
			\toprule
			\textbf{Method} & \textbf{PSNR ↑} & \textbf{SSIM ↑} & \textbf{UIQM ↑} & \textbf{UCIQE ↑} & \textbf{CCF ↑} & \textbf{PCQI ↑} \\
			\midrule
			GDCP          & 13.345 & 0.663 & 2.179 & 0.601 & 48.856 & 0.833 \\
			UDCP          & 16.581 & 0.669 & 2.190 & 0.603 & 41.612 & 0.567 \\
			IBLA          & 17.032 & 0.781 & 2.317 & 0.604 & \uline{50.835} & 0.723 \\
			HS2CM2A       & 11.209 & 0.419 & 2.140 & 0.611 & \textbf{61.568} & 0.818 \\
			WaterNet      & 18.260 & 0.767 & 2.765 & 0.579 & 30.089 & 0.820 \\
			Shallow-uwnet & 18.580 & 0.773 & 2.988 & 0.368 & 25.972 & 0.807 \\
			UWCNN         & 15.518 & \textbf{0.844} & 2.842 & 0.543 & 20.107 & 0.467 \\
			Ucolor        & 19.318 & 0.774 & 2.815 & 0.563 & 28.254 & 0.826 \\
			PUIE-MC       & 19.104 & 0.795 & \uline{3.217} & 0.598 & 26.270 & \uline{0.837} \\
			CWR           & \uline{20.012} & 0.789 & 2.966 & 0.609 & 22.909 & 0.591 \\
			FiveAPlus     & 19.241 & 0.790 & 3.085 & 0.609 & 33.557 & 0.834 \\
			Spectroformer & 19.196 & 0.785 & 2.852 & 0.610 & 29.801 & 0.619 \\
			NU2Net        & 19.001 & 0.751 & 3.156 & 0.593 & 29.219 & 0.817 \\
			X-caunet      & 19.903 & 0.805 & 3.093 & 0.615 & 32.263 & 0.827 \\
			WaterMamba    & 18.981 & 0.790 & 3.077 & \uline{0.616} & 36.090 & 0.833 \\
			Ours          & \textbf{20.203} & \uline{0.831} & \textbf{3.225} & \textbf{0.618} & 39.789 & \textbf{0.844} \\
			\bottomrule
		\end{tabularx}

\end{table*}

\begin{table*}[t]
	\centering
	\small
	
		\caption{Quantitative method comparisons across LSUI datasets\cite{New5}. The highest value is in bold, and the second-highest is underlined.}
		
		\label{LSUI_Data}
		\begin{tabularx}{\textwidth}{lYYYYYY}
			\toprule
			\textbf{Method} & \textbf{PSNR ↑} & \textbf{SSIM ↑} & \textbf{UIQM ↑} & \textbf{UCIQE ↑} & \textbf{CCF ↑} & \textbf{PCQI ↑} \\
			\midrule
			GDCP          & 13.287 & 0.695 & 2.261 & 0.603 & 37.884 & 0.782 \\
			UDCP          & 14.619 & 0.648 & 2.238 & 0.577 & 34.535 & 0.506 \\
			IBLA          & 17.392 & 0.720 & 2.281 & 0.595 & \uline{40.499} & 0.703 \\
			HS2CM2A       & 12.497 & 0.559 & 2.316 & \uline{0.608} & \textbf{53.140} & 0.888 \\
			WaterNet      & 20.563 & 0.820 & 2.888 & 0.590 & 23.814 & 0.876 \\
			Shallow-uwnet & 19.869 & 0.707 & 2.921 & 0.554 & 20.641 & 0.710 \\
			UWCNN         & 10.723 & 0.377 & 2.174 & 0.431 & 11.377 & 0.846 \\
			Ucolor        & 20.780 & 0.834 & 2.980 & 0.590 & 23.459 & 0.805 \\
			PUIE-MC       & 20.592 & 0.832 & 3.162 & 0.574 & 22.488 & 0.872 \\
			CWR           & 20.344 & 0.808 & 2.644 & 0.602 & 17.477 & 0.689 \\
			FiveAPlus     & 21.299 & \uline{0.839} & 3.066 & 0.600 & 26.607 & 0.877 \\
			Spectroformer & 20.928 & 0.817 & 2.891 & 0.604 & 27.974 & 0.729 \\
			NU2Net        & \uline{21.729} & 0.838 & \uline{3.186} & 0.592 & 24.757 & 0.877 \\
			X-caunet      & 20.831 & 0.806 & 3.009 & 0.596 & 26.252 & 0.871 \\
			WaterMamba    & 20.748 & 0.766 & 3.015 & 0.603 & 29.622 & \uline{0.890} \\
			Ours          & \textbf{21.975} & \textbf{0.843} & \textbf{3.210} & \textbf{0.610} & 30.658 & \textbf{0.916} \\
			\bottomrule
		\end{tabularx}
	
\end{table*}

\begin{figure}[t]
	\centering
	\includegraphics[angle=0,width=3.5in]{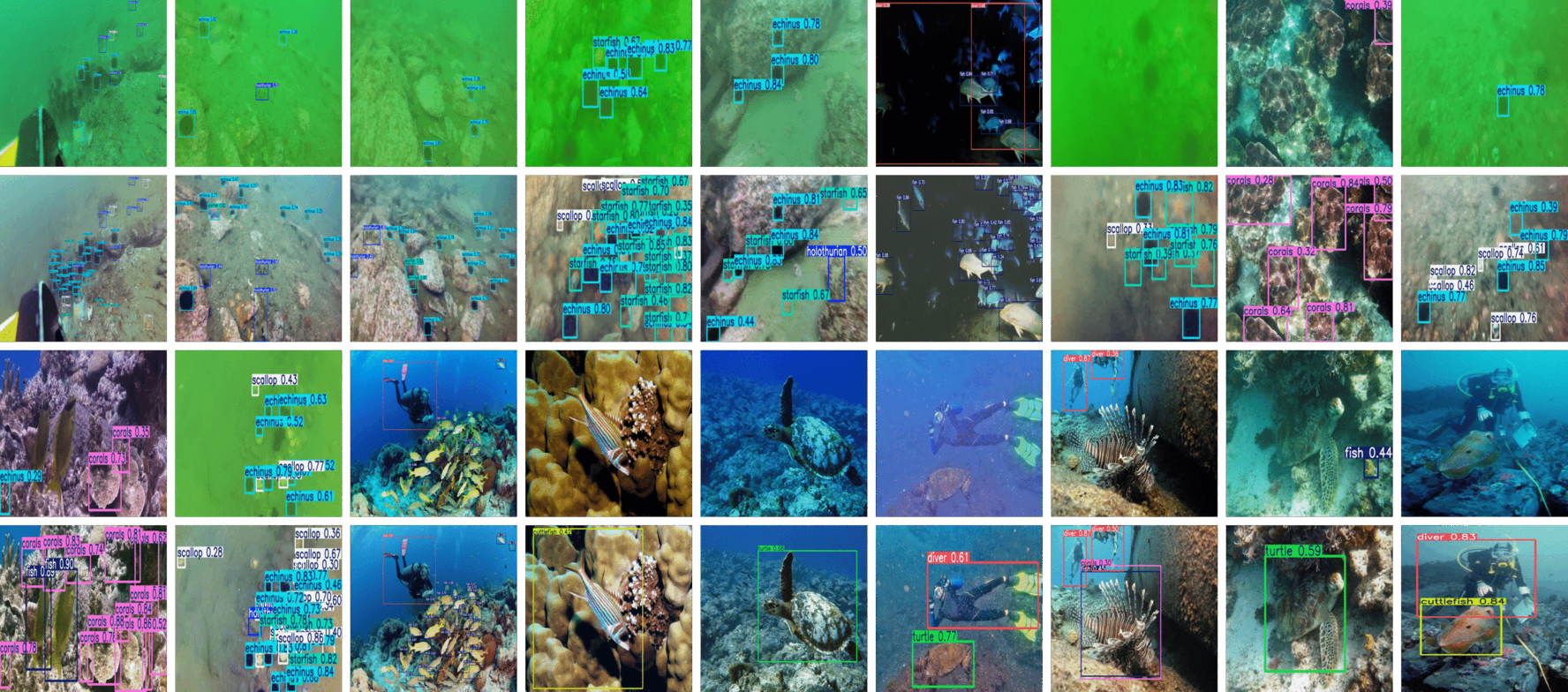}
	\caption{Comparison of detection performance of YOLOv5 on the RUOD dataset before and after UIE using the JDPNet model. The first, third, and fifth rows show the original images, while the second, fourth, and sixth rows show the corresponding UIE images.}
	\label{RUODOD}
\end{figure}

\begin{figure}[t]
	\centering
	\includegraphics[angle=0,width=3.5in]{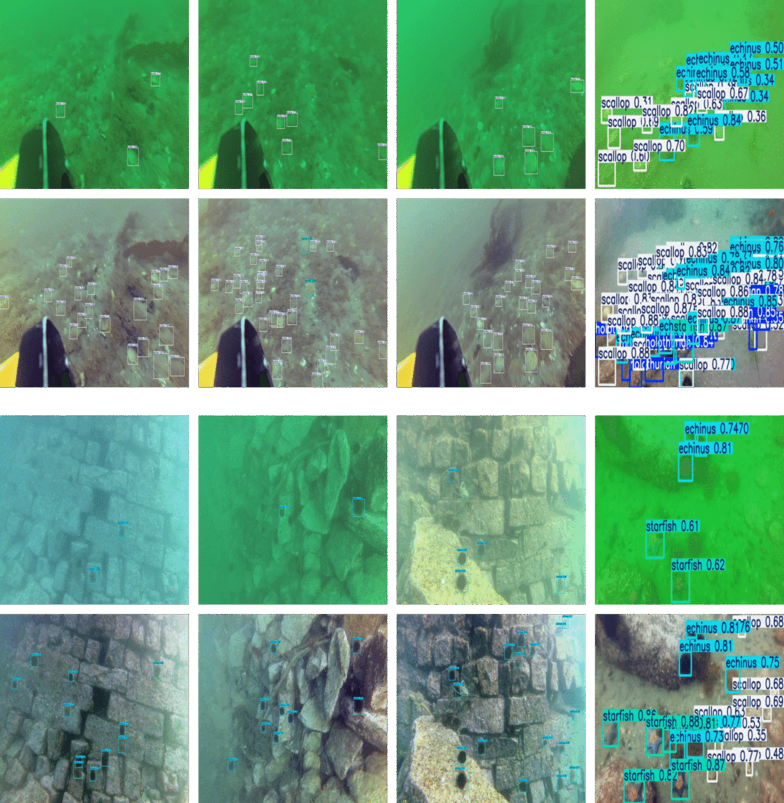}
	\caption{Comparison of detection performance of YOLOv5 on the URPC2020 dataset before and after UIE using the JDPNet model. The first and third rows show the original images, while the second and fourth rows show the corresponding UIE images.
		.}
	\label{URPC2020OD}
\end{figure}

\begin{table*}[!t]
	\centering
	\caption{Quantitative method comparisons across U45\cite{U45}, C60, and UCCS\cite{UCCS} datasets. The highest value is in bold, and the second-highest is underlined.}
	\label{U45C60UCCS}
	\small
	\begin{tabularx}{\textwidth}{l|YYY|YYY|YYY|YY}
		\toprule
		\textbf{Method} 
		& \multicolumn{3}{c|}{\textbf{U45}} 
		& \multicolumn{3}{c|}{\textbf{C60}} 
		& \multicolumn{3}{c|}{\textbf{UCCS}} 
		& \textbf{Params (M)} & \textbf{FLOPs (G)} \\
		& UIQM ↑ & UCIQE ↑ & CCF  ↑
		& UIQM ↑ & UCIQE ↑ & CCF  ↑
		& UIQM ↑ & UCIQE ↑ & CCF  ↑
		&  &  \\
		\midrule
		GDCP          & 2.389 & 0.593 & 35.868 & 1.810 & 0.565 & 25.799 & 2.386 & 0.555 & 19.435 & /     & /      \\
		UDCP          & 2.502 & 0.595 & 39.109 & 1.245 & 0.529 & 31.546 & 2.363 & 0.534 & 26.511 & /     & /      \\
		IBLA          & 2.387 & 0.565 & \uline{46.341} & 1.662 & 0.584 & \uline{32.473} & 2.115 & 0.525 & 25.195 & /     & /      \\
		HS2CM2A       & 2.716 & 0.602 & \textbf{53.333} & 2.294 & \textbf{0.615} & \textbf{36.204} & 2.888 & \textbf{0.588} & \textbf{38.435} & /     & /      \\
		WaterNet      & 2.957 & 0.576 & 21.625 & 2.113 & 0.550 & 21.013 & 3.057 & 0.569 & 18.453 & 24.81  & 142.90 \\
		Shallow-uwnet & 3.101 & 0.514 & 19.539 & 2.212 & 0.512 & 14.534 & 2.705 & 0.480 & 12.193 & 0.219  & 20.14  \\
		UWCNN         & 3.079 & 0.546 & 16.322 & 2.384 & 0.506 & 12.282 & 3.025 & 0.498 & 12.595 & 0.040  & 28.47  \\
		Ucolor        & 3.159 & 0.573 & 19.558 & 2.482 & 0.553 & 18.345 & 3.019 & 0.550 & 18.976 & 157.40 & 34.68  \\
		PUIE-MC       & 3.216 & 0.576 & 22.414 & 2.187 & 0.561 & 13.696 & 3.128 & 0.544 & 17.876 & 1.41   & 30.09  \\
		CWR           & 3.210 & 0.599 & 21.742 & 2.472 & \uline{0.593} & 14.839 & 3.109 & 0.567 & 17.970 & 11.4   & 42.43  \\
		FiveAPlus     & \uline{3.394} & 0.600 & 27.457 & 2.110 & 0.580 & 16.659 & 3.106 & 0.557 & 20.444 & 0.009  & 0.59   \\
		Spectroformer & 3.258 & \uline{0.610} & 34.279 & \uline{2.518} & 0.584 & 23.739 & 2.948 & 0.546 & 17.854 & 2.41   & 15.76  \\
		NU2Net        & 3.195 & 0.590 & 26.052 & 2.223 & 0.556 & 15.330 & \uline{3.192} & 0.555 & 20.141 & 3.15   & 29.95  \\
		X-caunet      & 3.336 & 0.597 & 31.018 & 2.095 & 0.559 & 14.518 & 2.993 & 0.540 & 17.139 & 31.78  & 261.5  \\
		WaterMamba    & 3.263 & 0.604 & 34.624 & 2.330 & 0.582 & 18.718 & 3.144 & 0.550 & 20.085 & 3.69   & 7.53   \\
		Ours          & \textbf{3.425} & \textbf{0.611} & 35.727 & \textbf{2.627} & 0.589 & 28.924 & \textbf{3.345} & \uline{0.574} & \uline{27.370} & 1.401  & 28.40  \\
		\bottomrule
	\end{tabularx}
\end{table*}

\section{More qualitative and quantitative comparison results on Effectiveness and Generalization of AquaBalanceLoss Across Network Architectures}

\begin{table*}[t]
	\centering
	\caption{Performance comparison of PUIE (CNN-based), Spectroformer (Transformer-based), and WaterMamba (Mamba-based) with and without ABL on different datasets.}
	\label{CNN-Trans-Mamba-ABL}
	\resizebox{\textwidth}{!}{%
		\begin{tabular}{llccc ccc ccc}
			\toprule
			\multirow{2}{*}{Set} & \multirow{2}{*}{Metric} &
			\multicolumn{3}{c}{PUIE (CNN-based)} &
			\multicolumn{3}{c}{Spectroformer (Transformer-based)} &
			\multicolumn{3}{c}{WaterMamba (Mamba-based)} \\
			\cmidrule(lr){3-5}\cmidrule(lr){6-8}\cmidrule(lr){9-11}
			& & base & with ABL & D-value & base & with ABL & D-value & base & with ABL & D-value \\
			\midrule
			\multirow{6}{*}{UIEB}
			& PSNR  & 22.180 & 22.729 &  \textbf{0.549} & 23.004 & 25.026 &  \textbf{2.022}  & 28.023 & 22.144 & -5.879 \\
			& SSIM  &  0.888 &  0.889 &  \textbf{0.001} &  0.907 &  0.899 & -0.008  &  0.849 &  0.761 & -0.088 \\
			& UIQM  &  2.947 &  3.041 &  \textbf{0.094} &  2.924 &  2.943 &  \textbf{0.019}  &  2.745 &  2.793 &  \textbf{0.048} \\
			& UCIQE &  0.589 &  0.606 &  \textbf{0.017} &  0.614 &  0.611 & -0.003  &  0.610 &  0.615 &  0.005 \\
			& PCQI  &  0.926 &  0.961 &  \textbf{0.035} &  0.776 &  0.851 &  \textbf{0.075}  &  0.746 &  0.762 &  \textbf{0.017} \\
			& CCF   & 19.465 & 24.248 &  \textbf{4.783} & 28.895 & 29.599 &  \textbf{0.704}  & 27.271 & 28.963 &  \textbf{1.692} \\
			\midrule
			\multirow{6}{*}{EUVP}
			& PSNR  & 19.104 & 18.768 & -0.336 & 19.196 & 18.642 & -0.5543 & 18.981 & 19.982 &  \textbf{1.001} \\
			& SSIM  &  0.795 &  0.796 &  \textbf{0.001} &  0.785 &  0.799 &  \textbf{0.014}  &  0.790 &  0.797 &  \textbf{0.007} \\
			& UIQM  &  3.217 &  3.218 &  \textbf{0.001 }&  2.852 &  2.970 &  \textbf{0.118}  &  3.077 &  3.080 &  \textbf{0.003} \\
			& UCIQE &  0.598 &  0.615 &  \textbf{0.017} &  0.610 &  0.619 &  \textbf{0.009}  &  0.616 &  0.604 & -0.012 \\
			& PCQI  &  0.837 &  0.843 &  \textbf{0.006} &  0.619 &  0.673 &  \textbf{0.054}  &  0.833 &  0.845 &  \textbf{0.012} \\
			& CCF   & 26.270 & 32.713 &  \textbf{6.443} & 29.801 & 34.414 &  \textbf{4.613}  & 36.090 & 37.560 &  \textbf{1.470} \\
			\midrule
			\multirow{6}{*}{LSUI}
			& PSNR  & 20.592 & 20.554 & -0.038 & 20.928 & 20.988 &  \textbf{0.060}  & 20.748 & 20.840 &  0.092 \\
			& SSIM  &  0.832 &  0.839 &  \textbf{0.007} &  0.817 &  0.817 &  \textbf{0.000}  &  0.766 &  0.767 &  \textbf{0.001} \\
			& UIQM  &  3.162 &  3.188 &  \textbf{0.026} &  2.891 &  2.893 &  \textbf{0.002}  &  3.015 &  3.133 &  \textbf{0.118} \\
			& UCIQE &  0.574 &  0.603 &  \textbf{0.029} &  0.604 &  0.606 &  \textbf{0.002}  &  0.603 &  0.606 &  \textbf{0.003} \\
			& PCQI  &  0.872 &  0.883 &  \textbf{0.011} &  0.729 &  0.789 &  \textbf{0.060}  &  0.890 &  0.891 &  \textbf{0.001} \\
			& CCF   & 22.488 & 26.876 &  \textbf{4.388} & 27.974 & 28.347 &  \textbf{0.373}  & 29.622 & 28.600 & -1.022 \\
			\midrule
			\multirow{3}{*}{U45}
			& UIQM  &  3.216 &  3.412 &  \textbf{0.196} &  3.258 &  2.978 & -0.280  &  3.263 &  3.289 &  \textbf{0.026} \\
			& UCIQE &  0.576 &  0.600 &  \textbf{0.024} &  0.610 &  0.613 &  \textbf{0.003}  &  0.604 &  0.593 & -0.011 \\
			& CCF   & 22.414 & 28.889 &  \textbf{6.475} & 34.279 & 35.571 &  \textbf{1.292}  & 34.624 & 34.769 &  \textbf{0.145} \\
			\midrule
			\multirow{3}{*}{C60}
			& UIQM  &  2.187 &  2.382 & \textbf{ 0.195} &  2.518 &  2.581 &  \textbf{0.063}  &  2.330 &  2.950 &  \textbf{0.620 }\\
			& UCIQE &  0.561 &  0.584 &  \textbf{0.023} &  0.584 &  0.587 &  \textbf{0.003}  &  0.582 &  0.585 &  \textbf{0.003} \\
			& CCF   & 13.696 & 18.445 &  \textbf{4.749} & 23.739 & 24.323 &  \textbf{0.584}  & 18.718 & 22.640 &  \textbf{3.922} \\
			\midrule
			\multirow{3}{*}{QDH}
			& UIQM  &  2.029 &  2.378 & \textbf{ 0.349} &  2.607 &  2.908 &  \textbf{0.301}  &  2.273 &  3.301 &  \textbf{1.028} \\
			& UCIQE &  0.516 &  0.542 &  \textbf{0.026} &  0.500 &  0.529 &  \textbf{0.029}  &  0.554 &  0.585 &  \textbf{0.031} \\
			& CCF   & 11.740 & 13.294 &  \textbf{1.554} & 15.750 & 19.641 & \textbf{ 3.891}  & 16.560 & 22.119 &  \textbf{5.559} \\
			\midrule
			\multirow{3}{*}{ROV}
			& UIQM  &  2.302 &  2.921 &  \textbf{0.619} &  2.920 &  3.000 &  \textbf{0.080}  &  1.753 &  3.131 &  \textbf{1.378} \\
			& UCIQE &  0.529 &  0.552 &  \textbf{0.023} &  0.541 &  0.549 &  \textbf{0.008}  &  0.546 &  0.555 &  \textbf{0.009} \\
			& CCF   & 12.200 & 16.520 &  \textbf{4.320} & 22.040 & 20.614 & -1.426  & 12.530 & 19.751 &  \textbf{7.222} \\
			\midrule
			\multirow{3}{*}{UCCS}
			& UIQM  &  3.128 &  3.250 &  \textbf{0.122} &  2.948 &  3.132 &  \textbf{0.184}  &  3.144 &  3.244 &  \textbf{0.100} \\
			& UCIQE &  0.544 &  0.572 &  \textbf{0.028} &  0.546 &  0.553 &  \textbf{0.007}  &  0.550 &  0.550 &  \textbf{0.000} \\
			& CCF   & 17.876 & 22.181 &  \textbf{4.305} & 17.854 & 24.615 &  \textbf{6.761}  & 20.085 & 23.867 &  \textbf{3.782} \\
			\bottomrule
	\end{tabular}}
\end{table*}

\section{More qualitative comparative results on the effectiveness of UIE}

\begin{figure}[t]
	\centering
	\includegraphics[angle=0,width=3.5in]{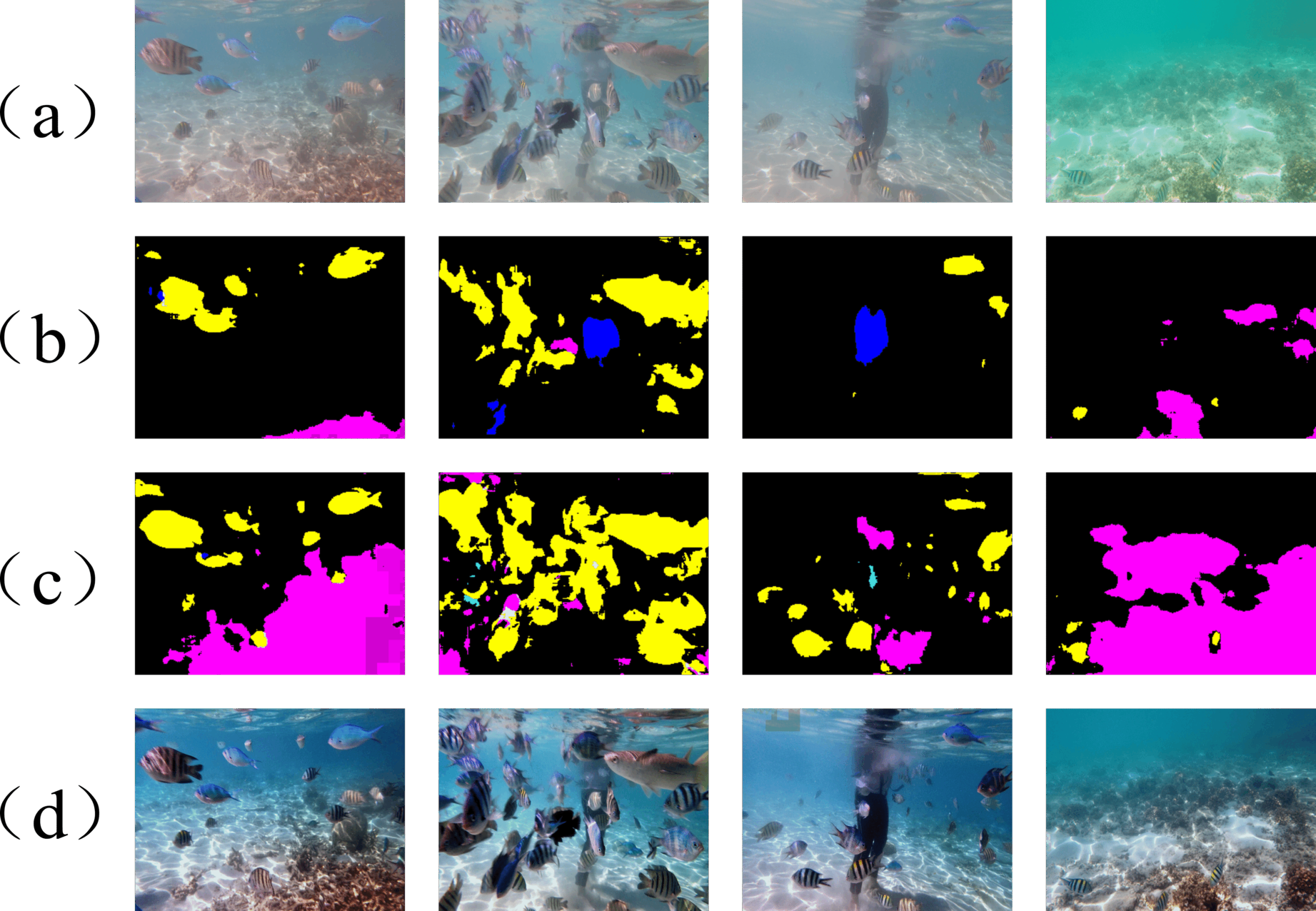}
	\caption{Qualitative visual comparison of SUIM-Net before and after JDPNet processing: (a) original input image; (b) segmentation result on the original image; (c) segmentation result after JDPNet processing; (d) JDPNet-enhanced image.}
	\label{SUIMNet-UIE}
\end{figure}

\begin{figure}[t]
	\centering
	\includegraphics[angle=0,width=3.5in]{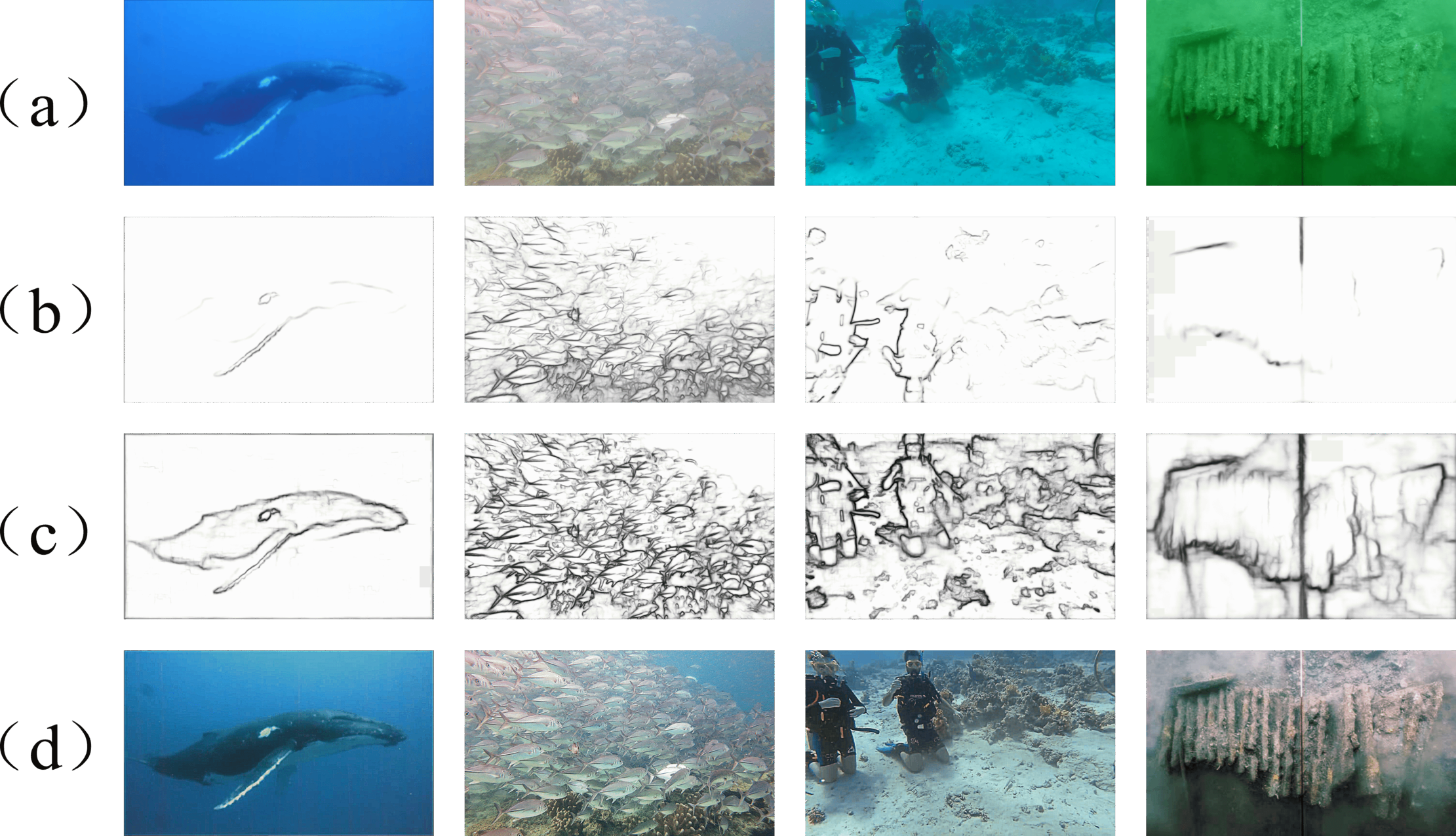}
	\caption{Qualitative visual comparison of RCF before and after JDPNet processing: (a) original input image; (b) edge detection result on the original image; (c) edge detection result after JDPNet processing; (d) JDPNet-enhanced image.}
	\label{RCF-UIE}
\end{figure}

\begin{figure}[t]
	\centering
	\includegraphics[angle=0,width=3.5in]{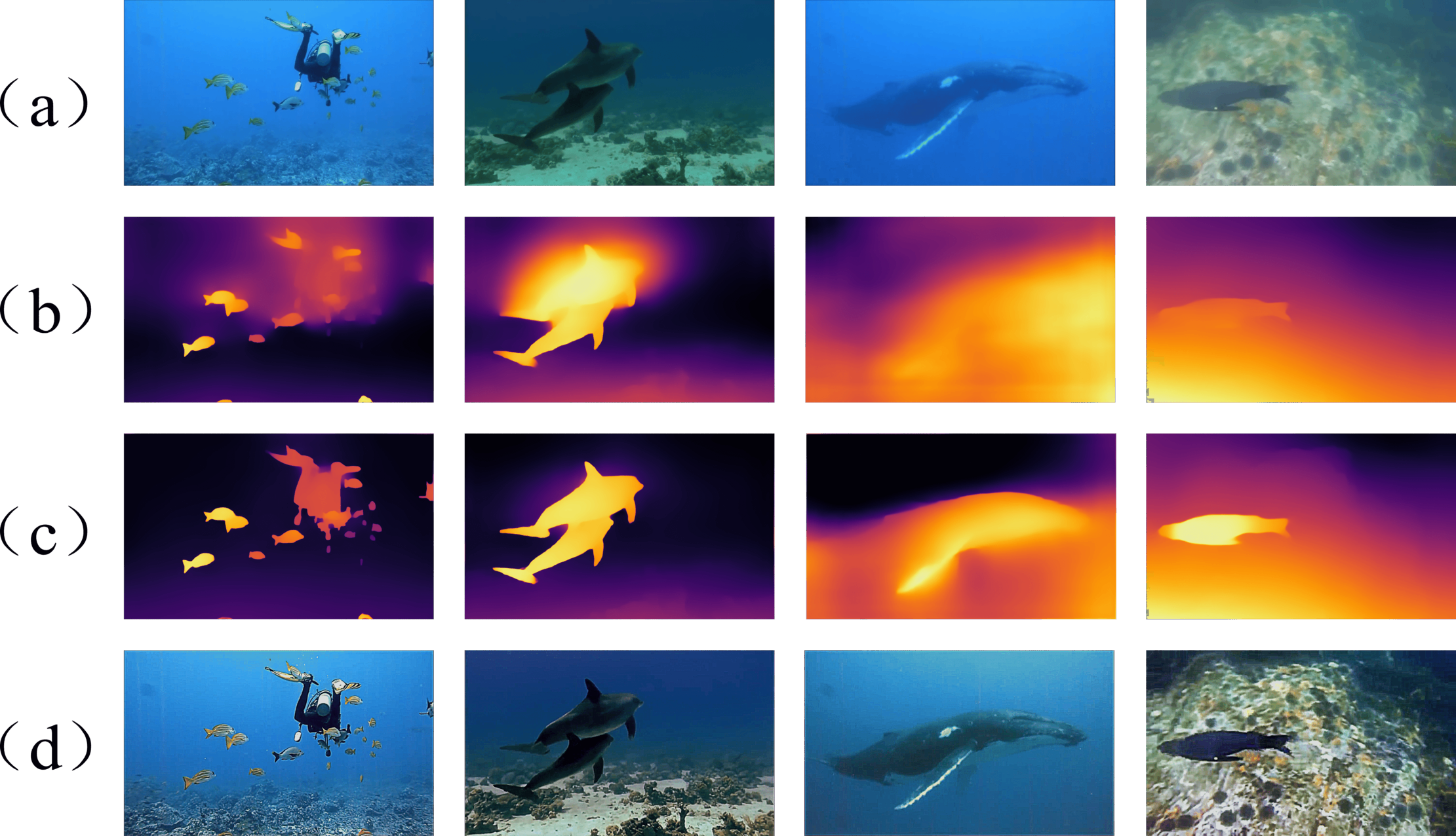}
	\caption{Qualitative visual comparison of MiDaS before and after JDPNet processing: (a) original input image; (b) depth estimation result on the original image; (c) depth estimation result after JDPNet processing; (d) JDPNet-enhanced image.}
	\label{MiDaS-UIE}
\end{figure}

\begin{figure}[t]
	\centering
	\includegraphics[angle=0,width=3.5in]{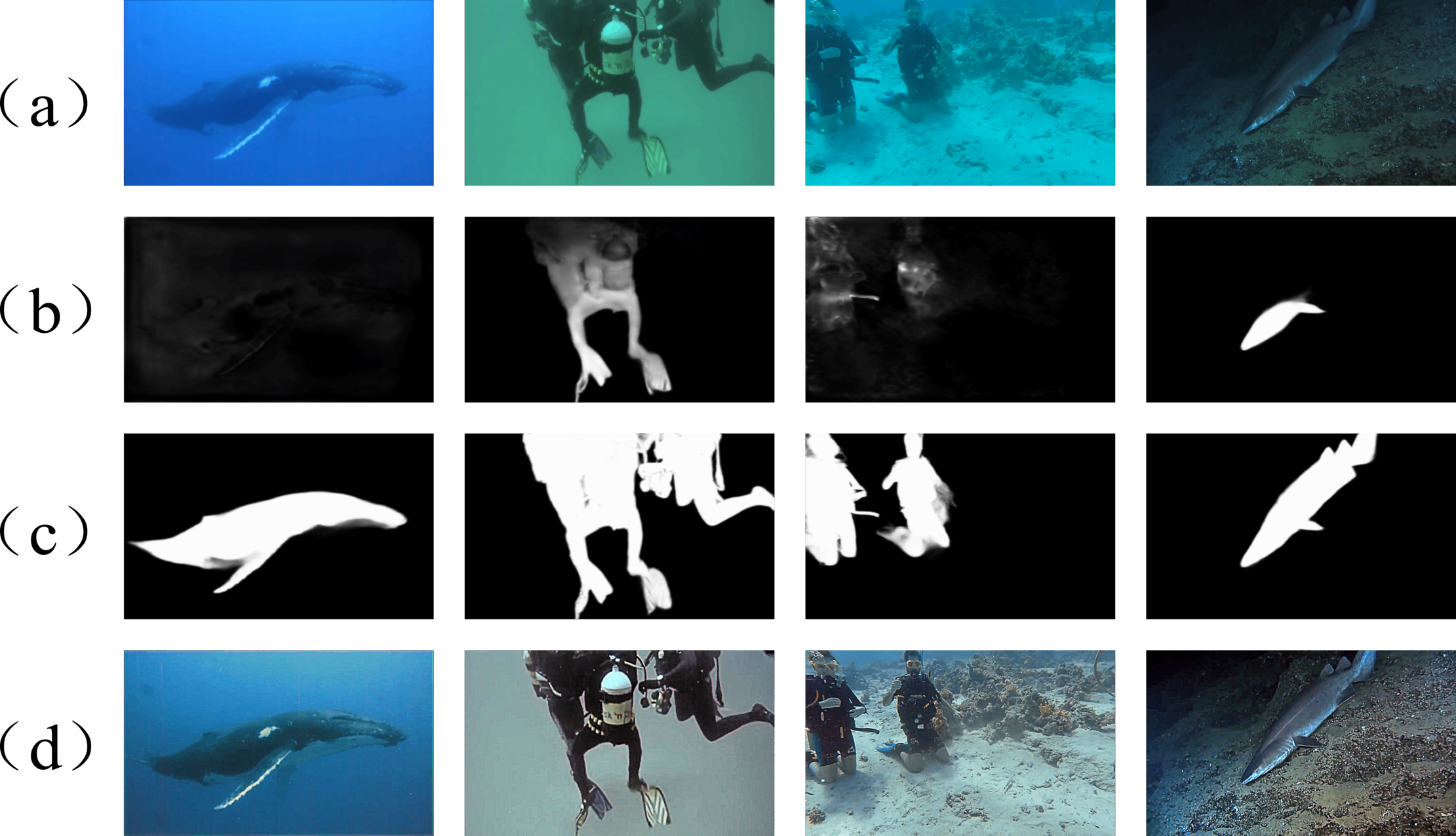}
	\caption{Qualitative visual comparison of U2-Net before and after JDPNet processing: (a) original input image; (b) saliency detection result on the original image; (c) saliency detection result after JDPNet processing; (d) JDPNet-enhanced image.}
	\label{U2Net-UIE}
\end{figure}

\newpage
\newpage
\newpage

\begin{IEEEbiography}[{\includegraphics[width=1in,height=1.25in,clip,keepaspectratio]{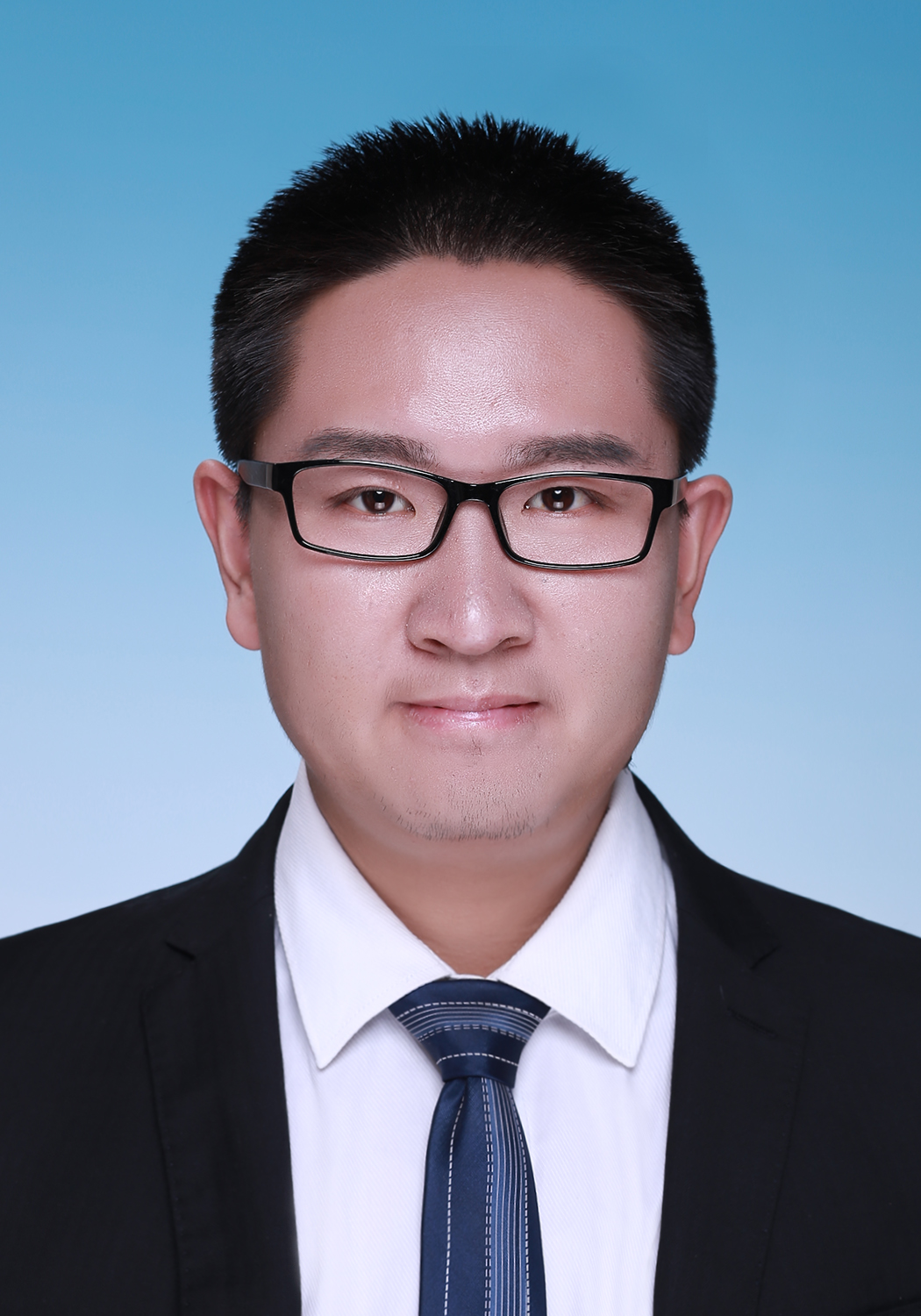}}]
	{Tao Ye}
	received his B.S. degree in measurement and control technology and instrumentation from the China University of Mining and Technology, Xuzhou, China, in 2009, M.S. degree in mechanical and electronic engineering from the China University of Mining and Technology, Beijing, China, in 2012, and Ph.D. degree in measurement technology and instruments from the Key Laboratory of Precision Opto-Mechatronics Technology of Ministry of Education, Beihang University, Beijing, in 2016. From March 2016 to March 2019, he worked as an engineer with the Beijing Institute of Remote Sensing and Equipment.
	He is currently an Associate Professor with the School of Mechanical and Electrical Engineering, China University of Mining and Technology.	
	His current research interests include deep learning and traffic detection.
\end{IEEEbiography}

\begin{IEEEbiography}[{\includegraphics[width=1in,height=1.25in,clip,keepaspectratio]{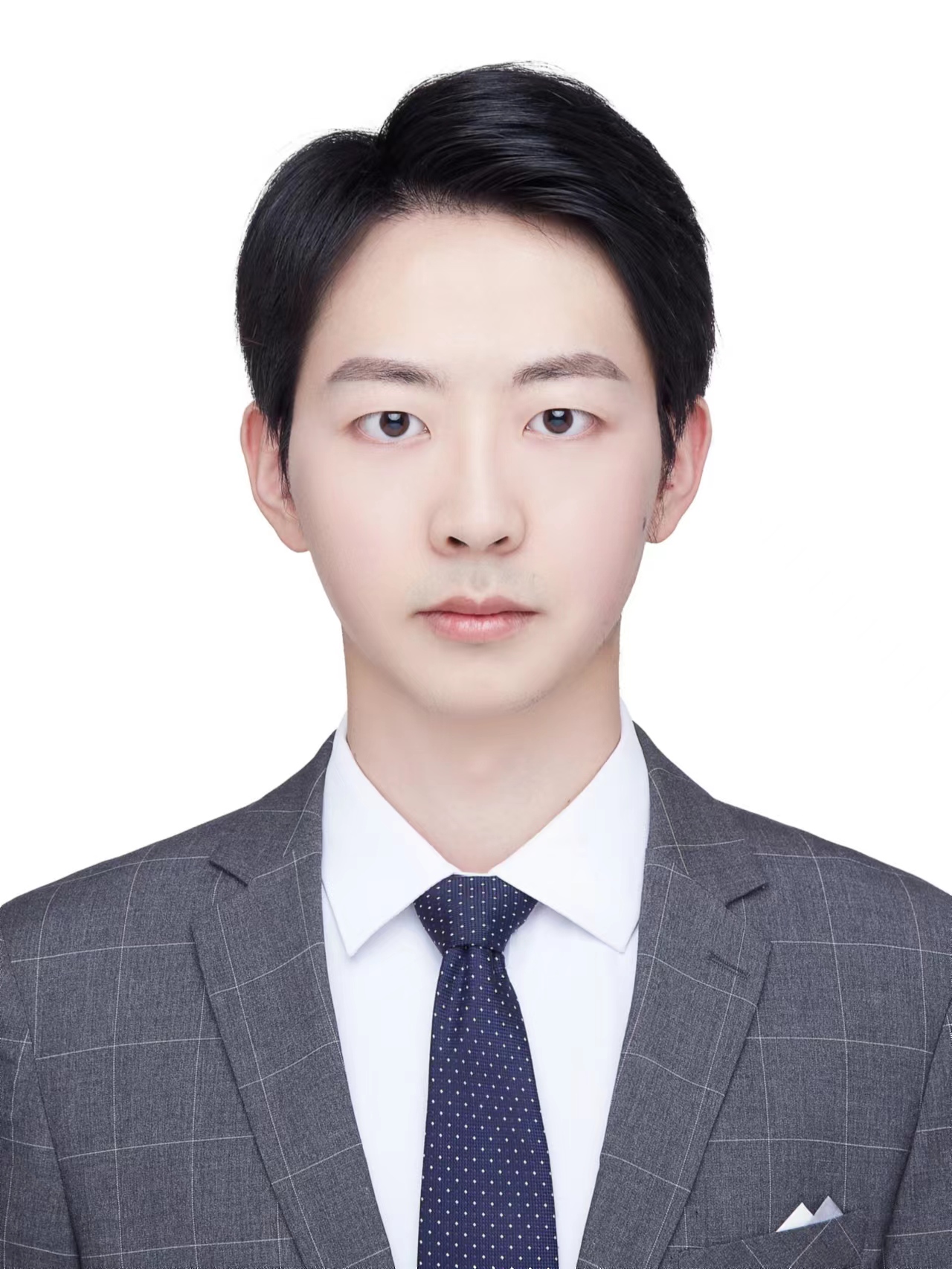}}]
	{Hongbin Ren}
	received his Master of Science degree in Mechanical Engineering from China University of Mining and Technology (Beijing) in 2025, and is currently a doctoral candidate in the School of Mechanical and Electrical Engineering at China University of Mining and Technology (Beijing). His current research interests include deep learning, computer vision, and image restoration.
\end{IEEEbiography}

\begin{IEEEbiography}[{\includegraphics[width=1in,height=1.25in,clip,keepaspectratio]{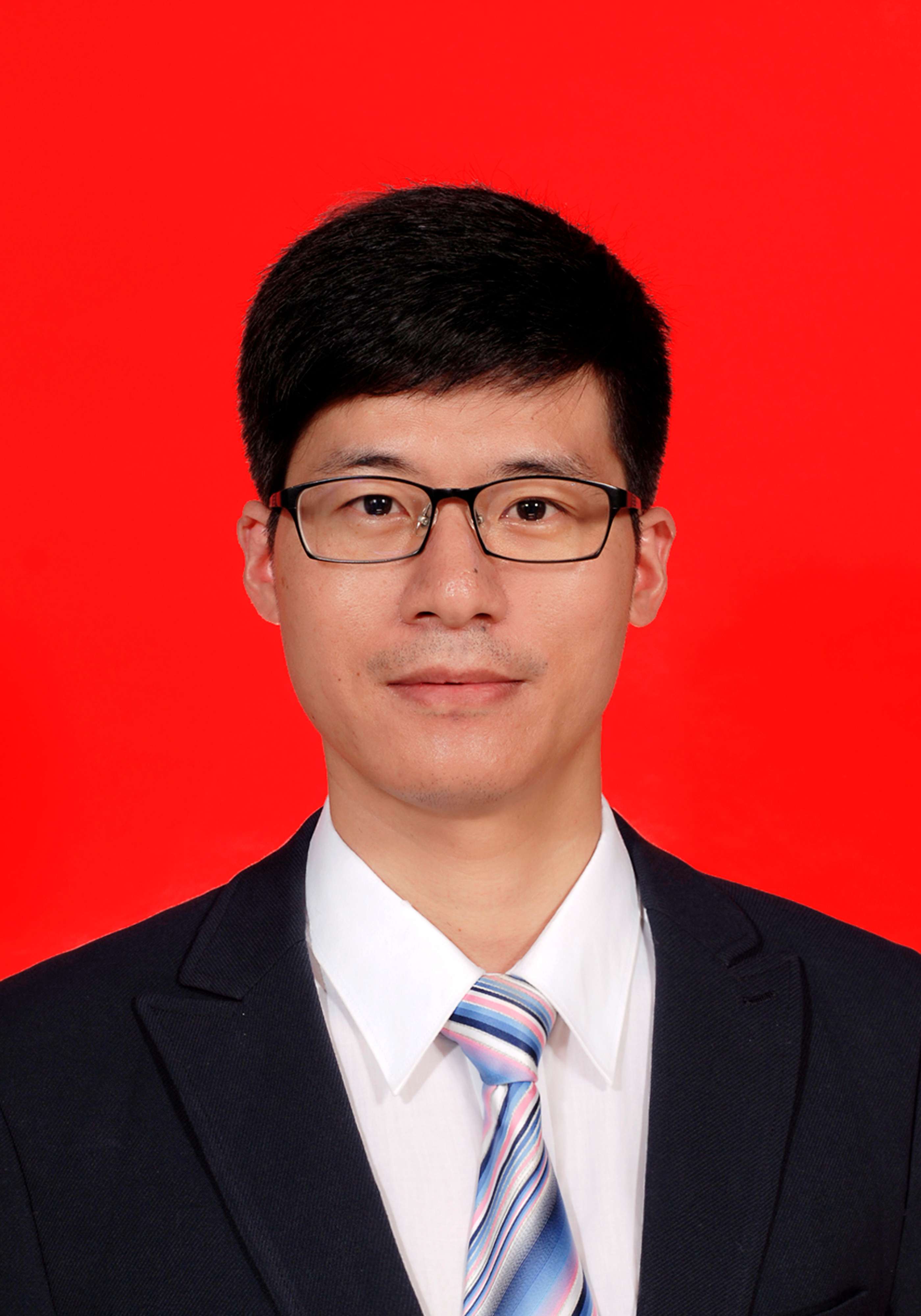}}]
	{Chongbing Zhang} received his master's degree in circuits and systems from the Chinese Academy of Sciences Shanghai Institute of Technical Physics, Shanghai, China, in 2017. He is currently an engineer at the China Shipbuilding Science Research Center, mainly focusing on the research and design of underwater intelligent perception systems.
\end{IEEEbiography}

\begin{IEEEbiography}[{\includegraphics[width=1in,height=1.25in,clip,keepaspectratio]{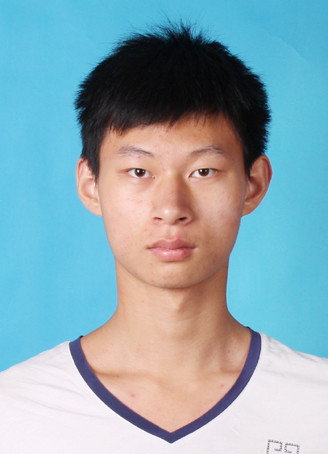}}]
	{Haoran Chen}
	received his B.S. degree in mechanical engineering from the China University of Mining and Technology, Beijing, China, in 2022. He is currently a graduate student of mechanical engineering with the College of Mechanical and Electrical Engineering, China University of Mining and Technology. His current research interests include deep learning, computer vision, and image segmentation.
\end{IEEEbiography}

\begin{IEEEbiography}[{\includegraphics[width=1in,height=1.25in,clip,keepaspectratio]{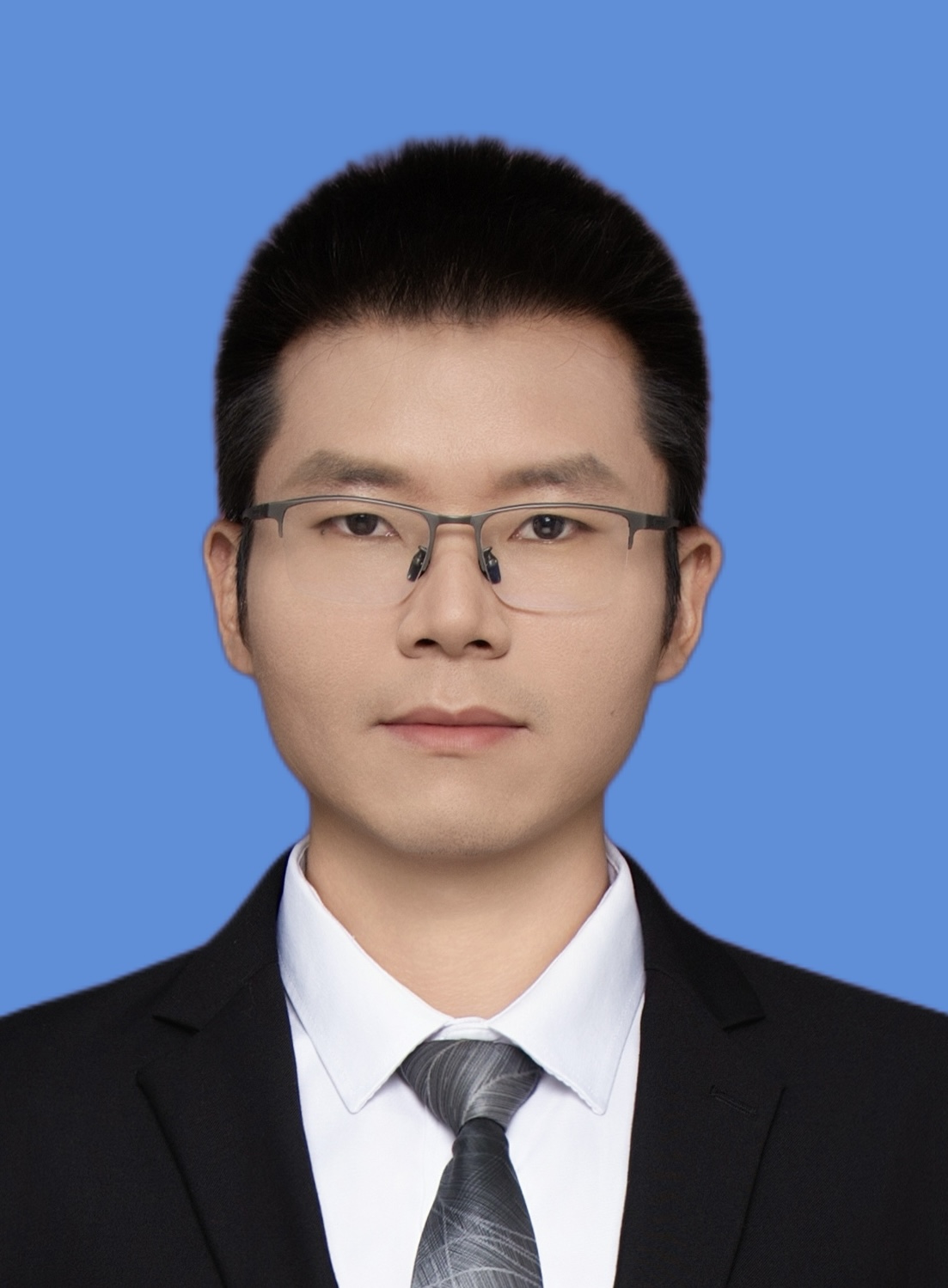}}]
	{Xiaosong Li} 
	received his Ph.D. in measurement technology and instruments from Beihang University in 2021. He is currently an associate professor with the School of Physics and Optoelectronic Engineering, Foshan University, Foshan, China; His research interests include image processing, pattern recognition, and optical metrology.
\end{IEEEbiography}

\vfill

\end{document}